\documentclass[nohyperref]{article}

\usepackage[usenames,dvipsnames]{xcolor}

\usepackage{microtype}
\usepackage{graphicx}
\usepackage{subfigure}
\usepackage{booktabs} %

\usepackage{hyperref}

\usepackage[arxiv]{icml2023}

\usepackage{amsmath}
\usepackage{amssymb}
\usepackage{mathtools}
\usepackage{amsthm}

\usepackage[capitalize,noabbrev]{cleveref}

\theoremstyle{plain}

\theoremstyle{definition}

\theoremstyle{remark}

\usepackage[textsize=tiny]{todonotes}

\usepackage{color}
\usepackage{xspace}
\usepackage{soul}
\usepackage{listings}
\usepackage{caption}
\usepackage{multirow, makecell}
\usepackage{multicol}

\usepackage{tabularx}
\newcolumntype{L}[1]{>{\raggedright\let\newline\\\arraybackslash\hspace{0pt}}m{#1}}
\newcolumntype{C}[1]{>{\centering\let\newline\\\arraybackslash\hspace{0pt}}m{#1}}
\newcolumntype{R}[1]{>{\raggedleft\let\newline\\\arraybackslash\hspace{0pt}}m{#1}}
\newcolumntype{Y}{>{\centering\arraybackslash}X}

\definecolor{codegreen}{rgb}{0,0.6,0}
\definecolor{codegray}{rgb}{0.5,0.5,0.5}
\definecolor{codepink}{RGB}{252, 142, 172}
\definecolor{codepurple}{rgb}{0.58,0,0.82}
\definecolor{backcolour}{RGB}{245,245,245}
\lstdefinestyle{mystyle}{
    backgroundcolor=\color{backcolour},   
    commentstyle=\color{magenta},
    keywordstyle=\color{blue},
    numberstyle=\tiny\color{codegray},
    stringstyle=\color{codepurple},
    basicstyle=\fontfamily{\ttdefault}\footnotesize,
    breakatwhitespace=false,         
    breaklines=true,                 
    captionpos=b,                    
    keepspaces=true,    
    frame=single,
    numbersep=5pt,                  
    showspaces=false,                
    showstringspaces=false,
    showtabs=false,                  
    tabsize=2,
    classoffset=1, %
    otherkeywords={range},
    keywordstyle=\color{violet},
    classoffset=0,
}

\definecolor{snsgray}{RGB}{179, 179, 179}
\definecolor{snsorange}{RGB}{252, 141, 98}
\definecolor{snsblue}{RGB}{141, 160, 203}

\definecolor{coolgrey}{RGB}{157,157,157}
\definecolor{lightgrey}{RGB}{235,238,238}
\definecolor{lightteal}{RGB}{198,211,222}
\definecolor{cyan}{RGB}{136, 204, 238}
\definecolor{teal}{RGB}{68, 170, 153}
\definecolor{sand}{RGB}{221, 204, 119}
\definecolor{rose}{RGB}{204, 102, 119}
\definecolor{red}{RGB}{250, 94, 91}
\definecolor{orange}{RGB}{255, 200, 63}
\definecolor{yellow}{RGB}{254, 239, 109}

\definecolor{darkgreen}{rgb}{0.09, 0.45, 0.27}

\newcommand{\vima}[0]{\mbox{VIMA}\xspace}
\newcommand{\vimabench}[0]{\textsc{VIMA-Bench}\xspace}
\newcommand{\webpage}[0]{\href{https://vimalabs.github.io/}{\texttt{vimalabs.github.io}}}
\newcommand{\bestscore}[1]{\textcolor{darkgreen}{\mathbf{#1}}}

\newcommand{\para}[1]{\paragraph{#1}\looseness=-1}

\usepackage{soul}

\icmltitlerunning{\vima: General Robot Manipulation with Multimodal Prompts}

\begin{document}

\twocolumn[
\icmltitle{\vima: General Robot Manipulation with Multimodal Prompts}

\icmlsetsymbol{equal_contribution}{$\dagger$}
\icmlsetsymbol{equal_advising}{$\ddagger$}

\begin{icmlauthorlist}
\icmlauthor{Yunfan Jiang}{Stanford}
\icmlauthor{Agrim Gupta}{Stanford,equal_contribution}
\icmlauthor{Zichen Zhang}{Charles_aff,equal_contribution}
\icmlauthor{Guanzhi Wang}{NVIDIA,Caltech,equal_contribution}
\icmlauthor{Yongqiang Dou}{Tsinghua}
\icmlauthor{Yanjun Chen}{Stanford}
\icmlauthor{}{}
\icmlauthor{Li Fei-Fei}{Stanford}
\icmlauthor{Anima Anandkumar}{NVIDIA,Caltech}
\icmlauthor{Yuke Zhu}{NVIDIA,Austin,equal_advising}
\icmlauthor{Linxi Fan}{NVIDIA,equal_advising}
\end{icmlauthorlist}

\icmlaffiliation{Stanford}{Stanford University;}
\icmlaffiliation{Charles_aff}{Macalester College, now at Allen Institute for AI;}
\icmlaffiliation{NVIDIA}{NVIDIA;}
\icmlaffiliation{Caltech}{Caltech;}
\icmlaffiliation{Tsinghua}{Tsinghua;}
\icmlaffiliation{Austin}{UT Austin. Work done during the first author's internship at NVIDIA. $\dagger$: Equal contribution. $\ddagger$: Equal advising}

\icmlkeywords{Robot Learning, Foundation Model, Transformer, Language Model, Multi-Task Learning}

\vskip 0.3in
]

\printAffiliationsAndNotice{}  %

\begin{abstract}

Prompt-based learning has emerged as a successful paradigm in natural language processing, where a single general-purpose language model can be instructed to perform any task specified by input prompts. Yet task specification in robotics comes in various forms, such as imitating one-shot demonstrations, following language instructions, and reaching visual goals. They are often considered different tasks and tackled by specialized models. We show that a wide spectrum of robot manipulation tasks can be expressed with \textit{multimodal prompts}, interleaving textual and visual tokens.
Accordingly, we develop a new simulation benchmark that consists of thousands of procedurally-generated tabletop tasks with multimodal prompts, 600K+ expert trajectories for imitation learning, and a four-level evaluation protocol for systematic generalization.
We design a transformer-based robot agent, \vima, that processes these prompts and outputs motor actions autoregressively. \vima features a recipe that achieves strong model scalability and data efficiency. It outperforms alternative designs in the hardest zero-shot generalization setting by up to $2.9\times$ task success rate given the same training data. With $10\times$ less training data, \vima still performs $2.7\times$ better than the best competing variant.
Code and video demos are available at \webpage.
\end{abstract}

\section{Introduction}
\label{sec:introduction}

Transformer models~\citep{vaswani2017attention} have given rise to remarkable multi-task consolidation across many AI domains. For example, users can describe a task using natural language prompt to GPT-3~\citep{brown2020gpt3}, allowing the same model to perform question answering, machine translation, text summarization, etc. Prompt-based learning provides an accessible and flexible interface to communicate a natural language understanding task to a general-purpose model.

We envision that a generalist robot should have a similarly intuitive and expressive interface for task specification. What does such an interface for robot learning look like? As a motivating example, consider a personal robot tasked with household activities. We can ask the robot to bring us a cup of water by a simple natural language instruction. If we require more specificity, we can instead instruct the robot to ``bring me \texttt{<image of the cup>}". For tasks requiring new skills, the robot should be able to adapt, preferably from a few video demonstrations \citep{duan2017oneshot}. Tasks that need interaction with unfamiliar objects can be easily explained via a few image examples for \textit{novel concept grounding} \citep{hermann2017grounded}. Finally, to ensure safe deployment, we can further specify visual constraints like ``do not enter \texttt{<image>} room". 

To enable a single agent with all these capabilities, we make three key contributions in this work:
1) a novel \textbf{multimodal prompting formulation} that converts a wide spectrum of robot manipulation tasks into one sequence modeling problem;
2) a \textbf{large-scale benchmark} with diverse tasks to systematically evaluate an agent's scalability and generalization; and
3) a \textbf{multimodal-prompted robot agent} capable of multi-task and zero-shot generalization.
\begin{figure*}[t]
    \centering
    \makebox[\textwidth][c]{\includegraphics[width=1.0\textwidth]{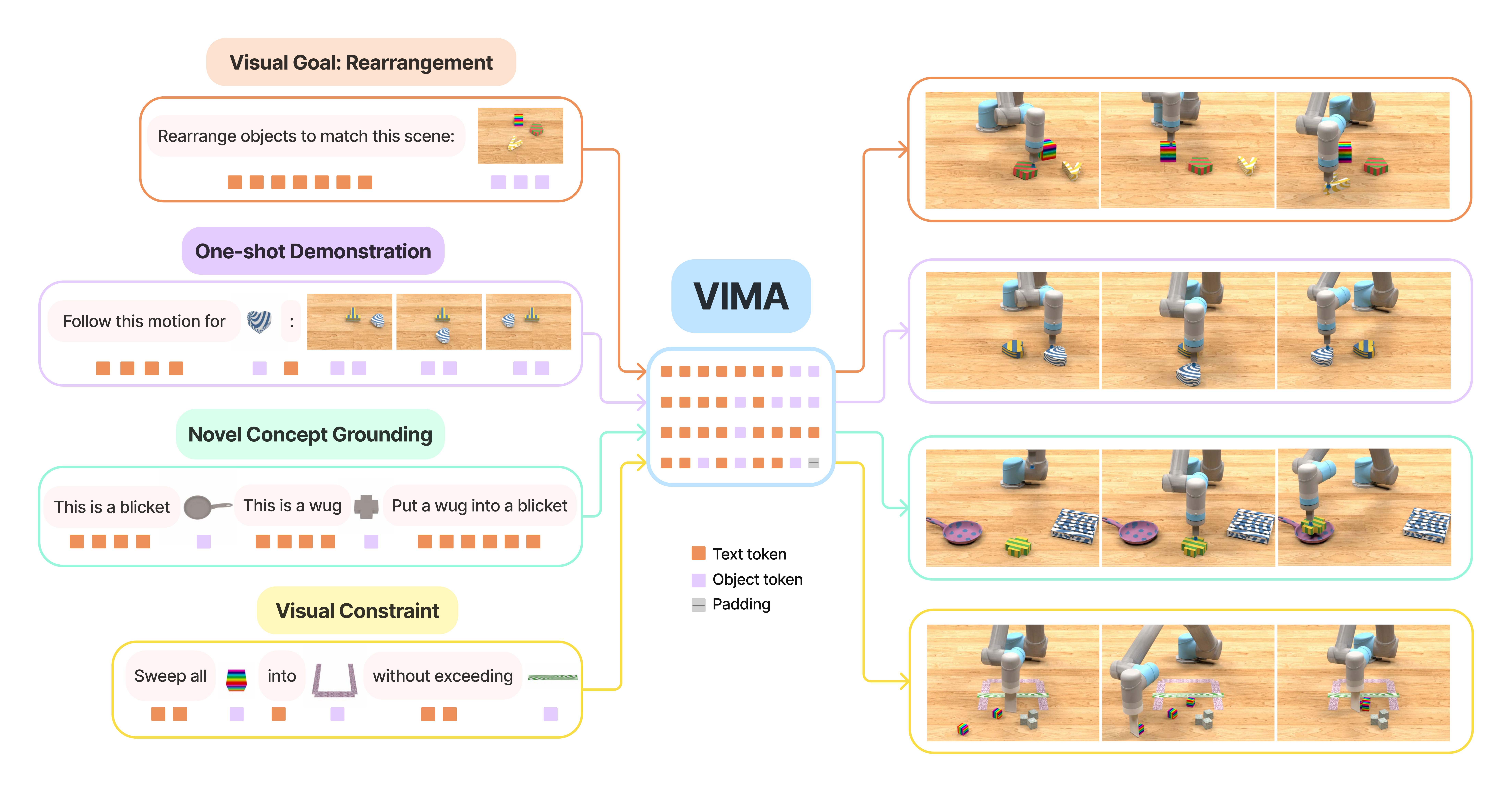}}
    \caption{\textbf{Multimodal prompts for task specification.} We observe that many robot manipulation tasks can be expressed as \textit{multimodal prompts} that interleave language and image/video frames. We introduce \vima, an embodied agent capable of processing mulitimodal prompts (left) and controlling a robot arm to solve the task (right).
    }
    \label{fig:pull_fig}
\end{figure*}

We start with the observation that many robot manipulation tasks can be formulated by \textbf{multimodal prompts that interleave language and images or video frames} (Fig.~\ref{fig:pull_fig}).   
For example, Rearrangement~\citep{batra2020rearrangement}, a type of \textit{Visual Goal}, can be formulated as ``Please rearrange objects to match this \texttt{\{scene\_image\}}"; 
\textit{Few-shot Imitation} can embed video snippet in the prompt ``Follow this motion trajectory for the wooden cube: $\{\texttt{frame}_1\}, \{\texttt{frame}_2\}, \{\texttt{frame}_3\}, \{\texttt{frame}_4\}$". 
Multimodal prompts not only have more expressive power than individual modalities but also enable a \textbf{uniform sequence IO interface} for training generalist robots. Previously, different robot manipulation tasks required distinct policy architectures, objective functions, data pipelines, and training procedures \citep{aceituno2021videoimitation,eskin2022rearrangement,lynch2021language}, leading to siloed robot systems that cannot be easily combined for a rich set of use cases. Instead, our multimodal prompt interface allows us to harness the latest advances in large transformer models~\citep{lin2021survey,tay2020efficient,khan2021transformers} for developing scalable multi-task robot learners.

\begin{figure*}[t]
    \centering
    \makebox[\textwidth][c]{\includegraphics[width=0.9\textwidth]{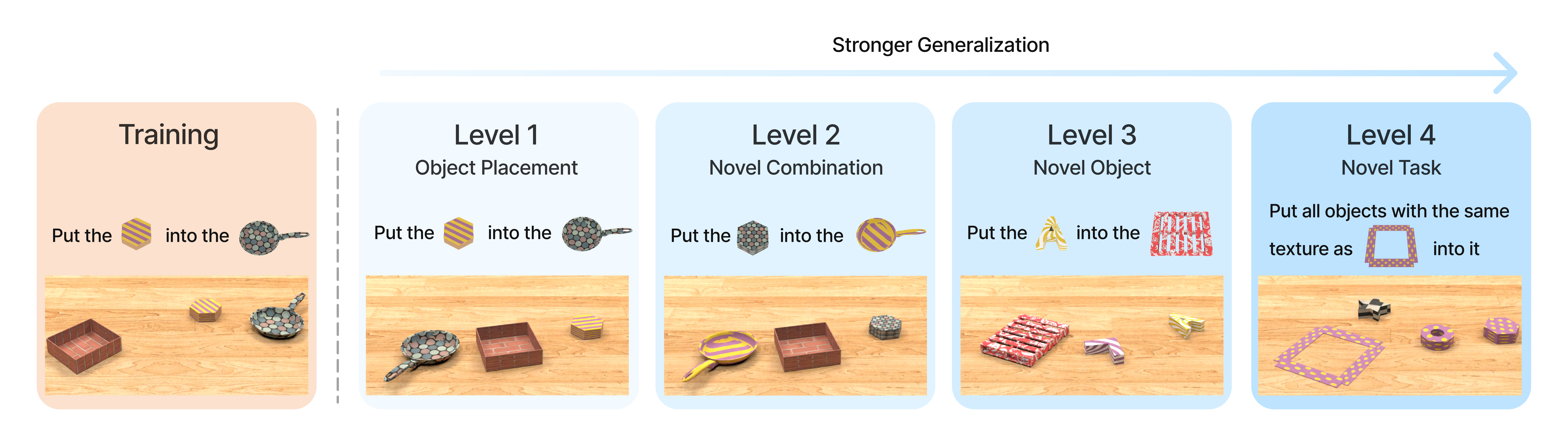}}
    \caption{\textbf{Evaluation Protocol in \vimabench.} We design 4 levels of evaluation settings to systematically measure the zero-shot generalization capability of an agent. Each level deviates more from the training distribution, and thus is strictly more challenging than the previous level.}
    \label{fig:eval-protocol}
\end{figure*}

To systematically evaluate agents with multimodal prompts, we develop a new benchmark, named \vimabench, built on the Ravens simulator~\citep{zeng2020transporter, shridhar2021cliport}. We provide 17 representative tasks with multimodal prompt templates. Each task can be procedurally instantiated into thousands of instances by various combinations of textures and tabletop objects. \vimabench establishes a four-level protocol to evaluate progressively stronger generalization capabilities, from randomized object placement to novel tasks (Fig.~\ref{fig:eval-protocol}).

To this end, we introduce the \textbf{Vi}suo\textbf{M}otor \textbf{A}ttention agent (\vima) to learn robot manipulation from multimodal prompts. The model architecture follows the encoder-decoder transformer design proven to be effective and scalable in NLP \citep{raffel2020t5}. \vima encodes an input sequence of interleaving textual and visual prompt tokens with a pre-trained language model \citep{deepmind2021frozen} and decodes robot control actions autoregressively for each environment interaction step. The transformer decoder is conditioned on the prompt via cross-attention layers that alternate with the usual causal self-attention.  
Instead of operating on raw images, \vima adopts an object-centric approach. We parse all images in the prompt or observation into objects by off-the-shelf then domain fine-tuned detectors~\citep{he2017mask} and flatten them into sequences of object tokens.
To demonstrate the scalability of \vima, we train a spectrum of 7 models ranging from 2M to 200M parameters.
Our approach outperforms other design alternatives, such as image patch tokens~\citep{reed2022gato}, image Perceiver~\citep{jaegle2021perceiver,alayrac2022flamingo}, and decoder-only conditioning~\citep{radford2018gpt}.
\vima obtains consistent performance gains across all four levels of zero-shot generalization and all model capacities, in some cases by a large margin (up to $2.9\times$ task success rate given the same amount of training data, and $2.7\times$ better even with $10\times$ less data).
We open-source the simulation environment, training dataset, algorithm code, and pre-trained model checkpoints to ensure reproducibility and facilitate future work from the community.
These materials along with video demos are available at \webpage.

\begin{figure*}[t]
    \centering
    \makebox[\textwidth][c]{\includegraphics[width=0.8\textwidth]{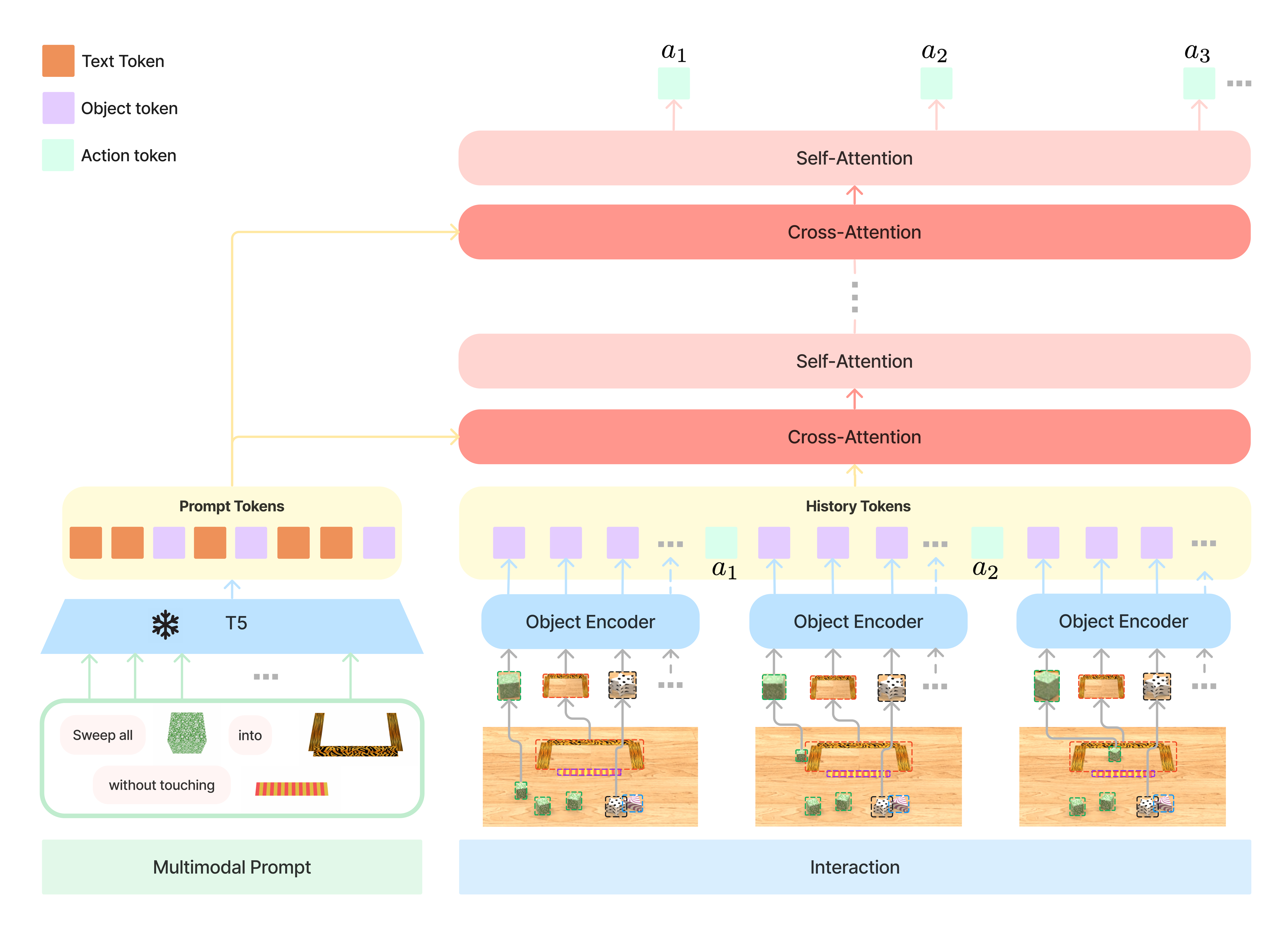}}
    \caption{\textbf{\vima Architecture.} We encode the multimodal prompts with a pre-trained T5 model, and condition the robot controller on the prompt through cross-attention layers. The controller is a causal transformer decoder consisting of alternating self and cross attention layers that predicts motor commands conditioned on prompts and interaction history.}
    \label{fig:arch}
\end{figure*}
\section{Multimodal Prompts for Task Specification}
\label{sec:benchmark}
A central and open problem in robot learning is task specification \citep{agrawal2022task}. In prior literature~\citep{stepputtis2020langintructedmanipulation,dasari2020transformers,brunke2021saferl}, different tasks often require diverse and incompatible interfaces, resulting in siloed robot systems that do not generalize well across tasks. Our key insight is that various task specification paradigms (such as goal conditioning, video demonstration, natural language instruction) can all be instantiated as multimodal prompts (Fig.~\ref{fig:pull_fig}). Concretely, a multimodal prompt $\mathcal{P}$ of length $l$ is defined as an ordered sequence of arbitrarily interleaved texts and images $\mathcal{P} \vcentcolon= \begin{bmatrix} x_1, x_2, \ldots, x_{l}\end{bmatrix}$, where each element $x_i \in \{\text{text}, \text{image}\}$. 

\para{Task Suite.}
The flexibility afforded by multimodal prompts allows us to specify and build models for a variety of task specification formats. Here we consider the following six categories.
\begin{enumerate}
    \item{\textbf{Simple object manipulation.} Simple tasks like ``put \texttt{<object>} into \texttt{<container>}", where each image in the prompt corresponds to a single object;}
    \item{\textbf{Visual goal reaching.} Manipulating objects to reach a goal configuration, \textit{e.g.}, \textit{Rearrangement}~\citep{batra2020rearrangement};}
    \item{\textbf{Novel concept grounding.} The prompt contains unfamiliar words like ``dax" and ``blicket", which are explained by in-prompt images and then immediately used in an instruction. This tests the agent's ability to rapidly internalize new concepts;}
    \item{\textbf{One-shot video imitation.} Watching a video demonstration and learning to reproduce the same motion trajectory for a particular object;}
    \item{\textbf{Visual constraint satisfaction.} The robot must manipulate the objects carefully and avoid violating the (safety) constraints;}
    \item{\textbf{Visual reasoning.} Tasks that require reasoning skills, such as appearance matching ``move all objects with same textures as \texttt{<object>} into \texttt{<container>}'', and visual memory ``put \texttt{<object>} in \texttt{<container>} and then restore to their original position''.}
\end{enumerate}

Note that these six categories are not mutually exclusive. For example, a task may introduce a previously unseen verb (\textit{Novel Concept}) by showing a video demonstration, or combine goal reaching with visual reasoning. More details about the task suite are discussed in Appendix, Sec.~\ref{supp:sec:task_suite}.

\section{\vimabench: Benchmark for Multimodal Robot Learning}

\para{Simulation Environment.}
Existing benchmarks are generally geared towards a particular task specification. To our knowledge, there is no benchmark that provides a rich suite of multimodal tasks and a comprehensive testbed for targeted probing of agent capabilities. To this end, we introduce a new benchmark suite for multimodal robot learning called \vimabench. We build our benchmark by extending the Ravens robot simulator~\citep{zeng2020transporter}. \vimabench supports extensible collections of objects and textures to compose multimodal prompts and to procedurally generate a large number of tasks. Specifically, we provide 17 tasks with multimodal prompt templates, which can be instantiated into thousands of task instances. Each task belongs to one or more of the 6 task categories mentioned above. \vimabench can generate large quantities of imitation learning data via scripted oracle agents. More details are elaborated in Appendix, Sec.~\ref{supp:sec:simulator}.

\para{Observation and Actions.}
The observation space of our simulator includes RGB images rendered from both frontal view and top-down view. Ground-truth object segmentation and bounding boxes are also provided for training object-centric models (Sec.~\ref{sec:method}). We inherit the high-level action space from \citet{zeng2020transporter}, which consists of primitive motor skills like ``pick and place'' and ``wipe''. These are parameterized by poses of the end effector. Our simulator also features scripted oracle programs that can generate expert demonstrations by using privileged simulator state information, such as the precise location of all objects, and the ground-truth interpretation of the multimodal instruction. 

\para{Training Dataset.}
We leverage oracles to generate a large offline dataset of expert trajectories for imitation learning. Our dataset includes 50K trajectories per task, and 650K successful trajectories in total. We hold out a subset of objects and textures for evaluation and designate 4 out of 17 tasks as a testbed for zero-shot generalization.

\para{Evaluating Zero-Shot Generalization.}
Each task in \vimabench has a binary success criterion and does not provide partial reward. 
During test time, we execute agent policies in the simulator for multiple episodes to compute a percentage success rate.
The average success rate over all evaluated tasks will be the final reported metric.

We design a four-level evaluation protocol (Fig.~\ref{fig:eval-protocol}) to systematically probe the generalization capabilities of learned agents.
Each level deviates more from the training distribution, and is thus strictly harder than the previous one.
\begin{enumerate}
    \item{\textbf{Placement generalization.} All prompts are seen verbatim during training, but only the placement of objects on the tabletop is randomized at testing;}
    \item{\textbf{Combinatorial generalization.} All textures and objects are seen during training, but new combinations of them appear in testing;}
    \item{\textbf{Novel object generalization.} Test prompts and the simulated workspace include novel textures and objects;}
    \item{\textbf{Novel task generalization.} New tasks with novel prompt templates at test time.}
\end{enumerate}

\begin{figure*}[t]
    \centering
    \makebox[\textwidth][c]{\includegraphics[width=0.9\textwidth]{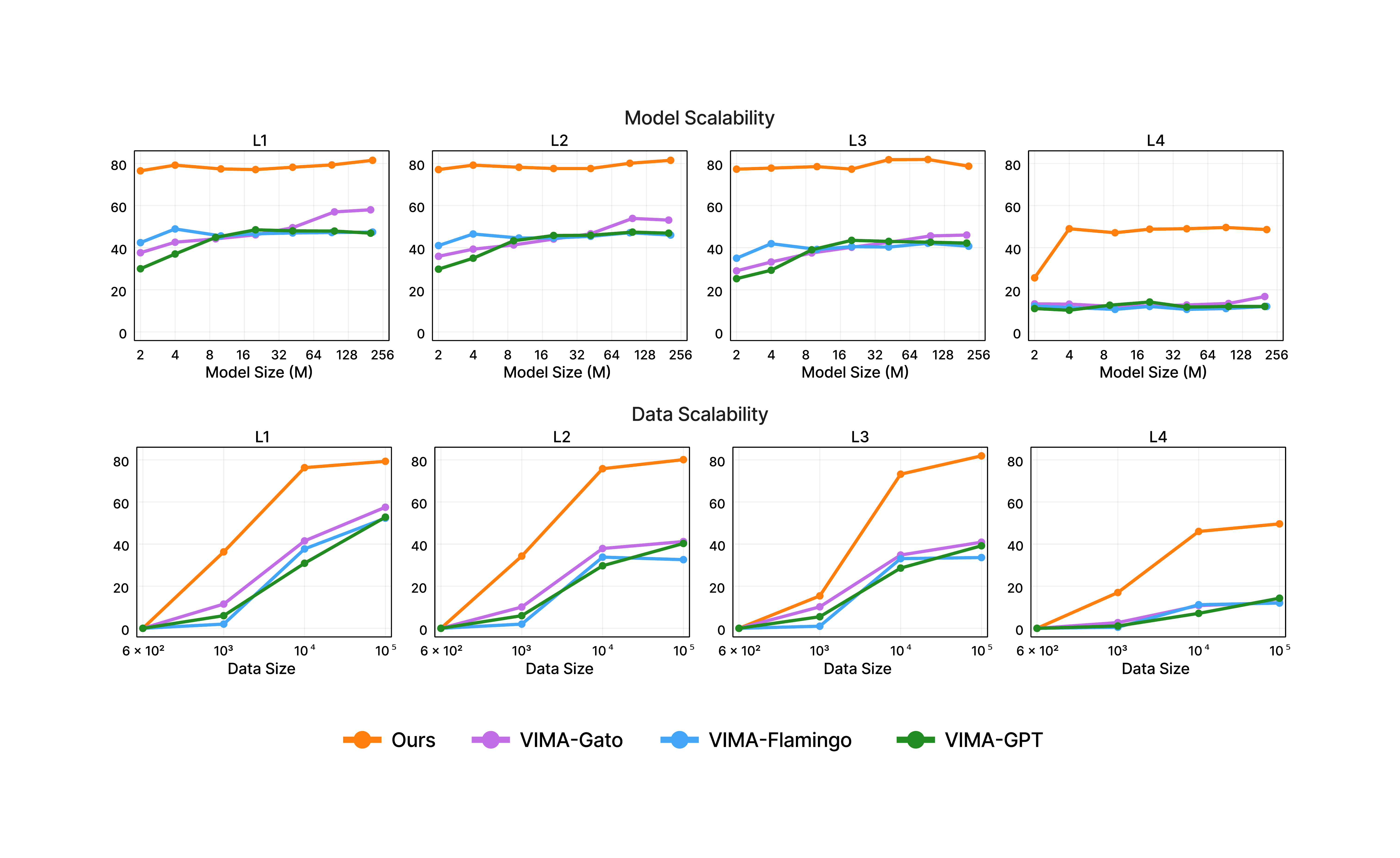}}
    \caption{\textbf{Scaling model and data.} \textit{Top}: We compare performance of different methods with model sizes ranging from 2M to 200M parameters. Across all model sizes and generalization levels, \vima outperforms baseline variants. \textit{Bottom}: For a fixed model size of 92M parameters we compare the effect of imitation learning dataset size with $0.1\%$, $1\%$, $10\%$, and full data. \vima is extremely sample efficient and can achieve performance comparable to other methods with $10\times$ less data.}
    \label{fig:scalability}
\end{figure*}

\section{\vima: Visuomotor Attention Agent}
\label{sec:method}
Our goal is to build a robot agent capable of performing any task specified by multimodal prompts.
There is no prior method that works out of the box with multimodal prompts.
To learn an effective multi-task robot policy, we propose \vima, a robot agent with a multi-task encoder-decoder architecture and object-centric design (Fig.~\ref{fig:arch}). Concretely, we learn a robot policy $\pi(a_t \vert \mathcal{P}, \mathcal{H})$, where $\mathcal{H} \vcentcolon= \begin{bmatrix} o_1, a_1, o_2, a_2, \ldots, o_t \end{bmatrix} $ denotes the past interaction history, and $o_t \in \mathcal{O}, a_t \in \mathcal{A}$ are observations and actions at each interaction steps. 
We encode multimodal prompts via a \textit{frozen} pre-trained language model and decode robot waypoint commands conditioned on the encoded prompts via cross-attention layers.
Unlike prior work~\citep{florence2019selfsupervised,sieb2019graphstructured,zhu2022viola}, \vima adopts an object-centric representation that computes tokens from bounding box coordinates and cropped RGB patches.

\para{Tokenization.}
There are $3$ formats of raw input in the prompt --- text, image of a single object, and image of a full tabletop scene (\textit{e.g.}, for \textit{Rearrangement} or imitation from video frames).
For \textbf{text inputs}, we use pre-trained T5 tokenizer and word embedding to obtain word tokens.
For \textbf{images of full scenes}, we first extract individual objects using domain fine-tuned Mask R-CNN~\citep{he2017mask} (Appendix, Sec.~\ref{supp:sec:mask_rcnn}).
Each object is represented as a bounding box and a cropped image. We then compute object tokens by encoding them with a bounding box encoder and a ViT~\citep{dosovitskiy2020image}, respectively. Since Mask R-CNN is imperfect, the bounding boxes can be noisy and the cropped images may have irrelevant pixels. For \textbf{images of single objects}, we obtain tokens in the same way except with a dummy bounding box.
Prompt tokenization produces a sequence of interleaved textual and visual tokens. We then follow the practice in \citet{deepmind2021frozen} and encode the prompt via a pre-trained T5 encoder~\mbox{\citep{raffel2020t5}}. Since T5 has been pre-trained on large text corpora, \vima inherits the semantic understanding capability and robustness properties. To accommodate tokens from new modalities, we insert MLPs between non-textual tokens and T5.

\para{Robot Controller.}
A challenging aspect of designing a multi-task policy is to select a suitable conditioning mechanism. In our schema (Fig.~\ref{fig:arch}), the robot controller (decoder) is conditioned on the prompt sequence $\mathcal{P}$ by a series of cross-attention layers between $\mathcal{P}$ and the trajectory history sequence $\mathcal{H}$.
We compute key $K_\mathcal{P}$ and value $V_\mathcal{P}$ sequences from the prompt and query $Q_\mathcal{H}$ from the trajectory history, following the encoder-decoder convention in \citet{raffel2020t5}.
Each cross-attention layer then generates an output sequence $\mathcal{H}^\prime = \text{softmax}\left(\frac{Q_\mathcal{H} K^\intercal_{\mathcal{P}} }{\sqrt{d}}   \right)V_\mathcal{P}$, where $d$ is the embedding dimension.
Residual connections are added to connect higher layers with the input rollout trajectory sequence.
The cross-attention design enjoys three advantages: 1) strengthened connection to prompt; 2) intact and deep flow of the original prompt tokens; and 3) better computational efficiency.  \vima decoder consists of $L$ alternating cross-attention and self-attention layers. Finally, we follow common practice \citep{openai2022vpt} to map predicted action tokens to discretized poses of the robot arm.
See Appendix, Sec.~\mbox{\ref{supp:sec:model_vima}} for more details.

\para{Training.}
We follow behavioral cloning to train our models by minimizing the negative log-likelihood of predicted actions. Concretely, for a trajectory with $T$ steps, we optimize $\min_\theta \sum_{t=1}^T - \log \pi_\theta (a_t \vert \mathcal{P}, \mathcal{H})$. The entire training is conducted on an offline dataset with no simulator access.
To make \vima robust to detection inaccuracies and failures, we apply \emph{object augmentation} by randomly injecting \emph{false-positive} detection outputs.
After training, we select model checkpoints for evaluation based on the aggregated accuracy on a held-out validation set. The evaluation involves interacting with the physics simulator.
We follow the best practices to train Transformer models.
See Appendix, Sec.~\ref{supp:sec:training} for comprehensive training hyperparameters.

\section{Experiments}
\label{sec:experiments}
In this section, we aim to answer three main questions:
\begin{enumerate}
    \item{What is the best recipe for building multi-task transformer-based robot agents with multimodal prompts?}
    \item{What are the \textbf{scaling properties} of our approach in model capacity and data size?}
    \item{How do different components, such as visual tokenizers, prompt conditioning, and prompt encoding, affect robot performance?}
\end{enumerate}

\subsection{Baselines}
Because there is no prior method that works out of the box with our multimodal prompting setup, we make our best effort to select a number of representative transformer-based agent architectures as baselines, and re-interpret them to be compatible with \vimabench:

\textbf{Gato}~\citep{reed2022gato} introduces a decoder-only model that solves tasks from multiple domains where tasks are specified by prompting the model with the observation and action subsequence. For a fair comparison, we provide the same conditioning as \vima, \textit{i.e.}, our multimodal encoded prompts. Input images are divided into patches and encoded by a ViT model to produce observation tokens.
This variant is referred to as ``\textbf{\vima-Gato}''.

\textbf{Flamingo}~\citep{alayrac2022flamingo} is a vision-language model that learns to generate textual completion in response to multimodal prompts. It embeds a variable number of prompt images into a fixed number of tokens via Perceiver~\citep{jaegle2021perceiver}, and conditions the language decoder on the encoded prompt by cross-attention.  
Flamingo does not work with embodied agents out of the box.
We adapt it to support decision-making by replacing the output layer with robot action heads.
We denote the method as ``\textbf{\vima-Flamingo}''.

\textbf{\vima-GPT} is a decoder-only architecture conditioned on tokenized multimodal prompts. It autoregressively decodes the next actions given instructions and interaction histories. Similar to prior work~\citep{chen2021decisiontransformer,janner2021onebigsequence}, it encodes an image into a single \textit{state} token by a ViT encoder and prepends the rollout trajectory with prompt tokens. This baseline does not use cross-attention.

A more detailed comparison between these variants can be found in Appendix, Sec.~\ref{supp:sec:method_compare}.

\subsection{Evaluation Results}
\label{sec:experiments-scalability}
We compare \vima against the baseline variants on four levels of generalization provided in our benchmark for different model and training dataset sizes.
Our empirical results demonstrate that \vima's choice of object tokens combined with cross-attention conditioning is the most effective recipe among the model designs we consider.

\para{Model Scaling.}
We train all methods for a spectrum of model capacities from 2M to 200M parameters, evenly spaced on the log scale (Fig.~\ref{fig:scalability}). The encoder size is kept constant (T5-Base, 111M) for all methods and excluded from the parameter count. Across \textit{all} levels of zero-shot generalization, we find that \vima strongly outperforms other alternatives. Although models like \vima-Gato and \mbox{\vima-Flamingo} show improved performance with bigger model sizes, \vima consistently achieves superior performance over \textit{all} model sizes. We note that this can only be achieved with \textit{both} cross-attention and object token sequence representations --- altering any component will significantly degrade the performance, especially in the low model capacity regime (ablations in Sec.~\ref{sec:experiments-ablation}).

\begin{figure}[t]
    \centering
    \includegraphics[width=0.4\textwidth]{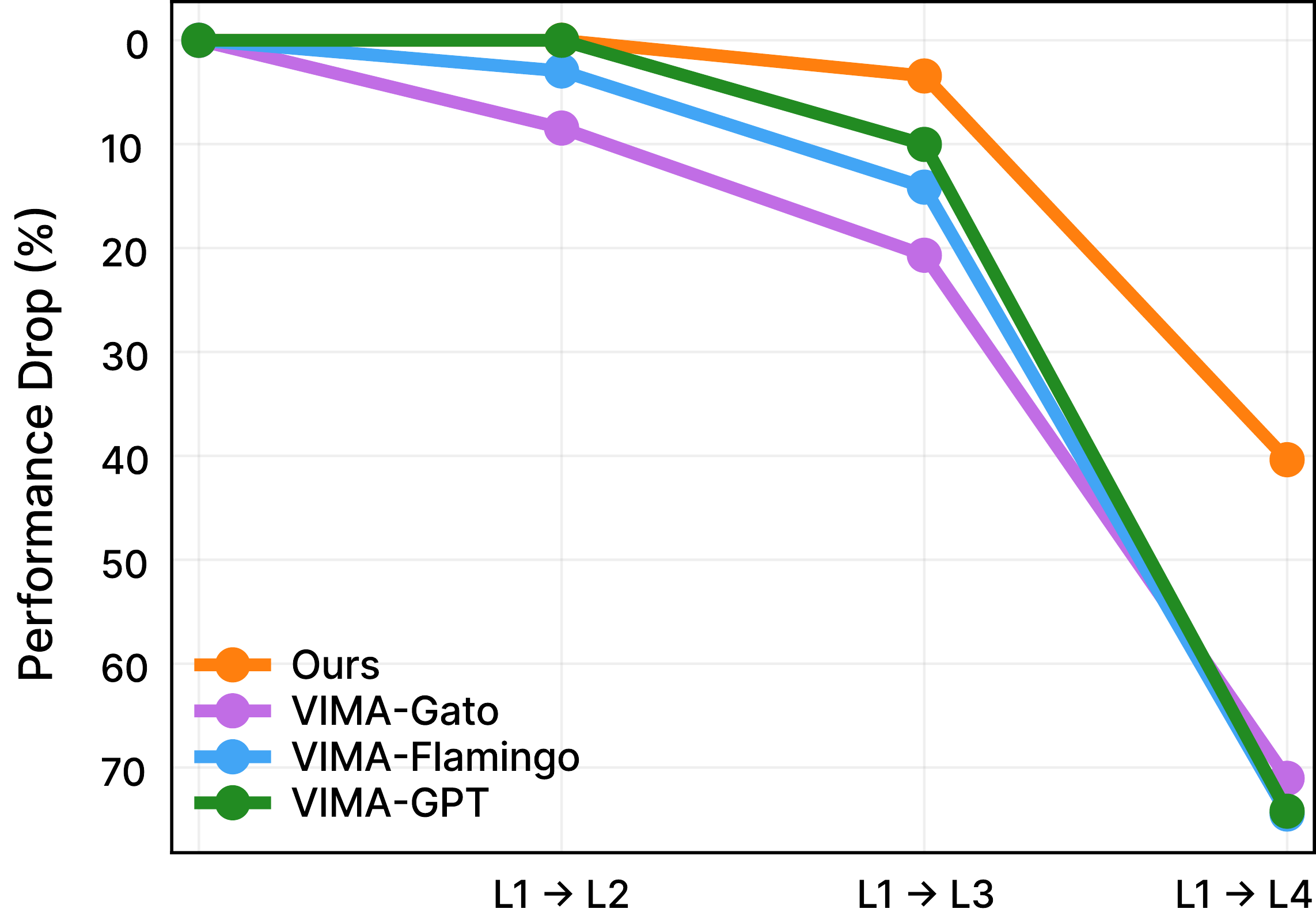}
    \caption{\vima incurs much less performance drop than baselines as we evaluate on progressively harder settings.}
    \label{fig:performance_decrease}
\end{figure}

\begin{figure*}[t]
    \centering
    \makebox[\textwidth][c]{\includegraphics[width=0.85\textwidth]{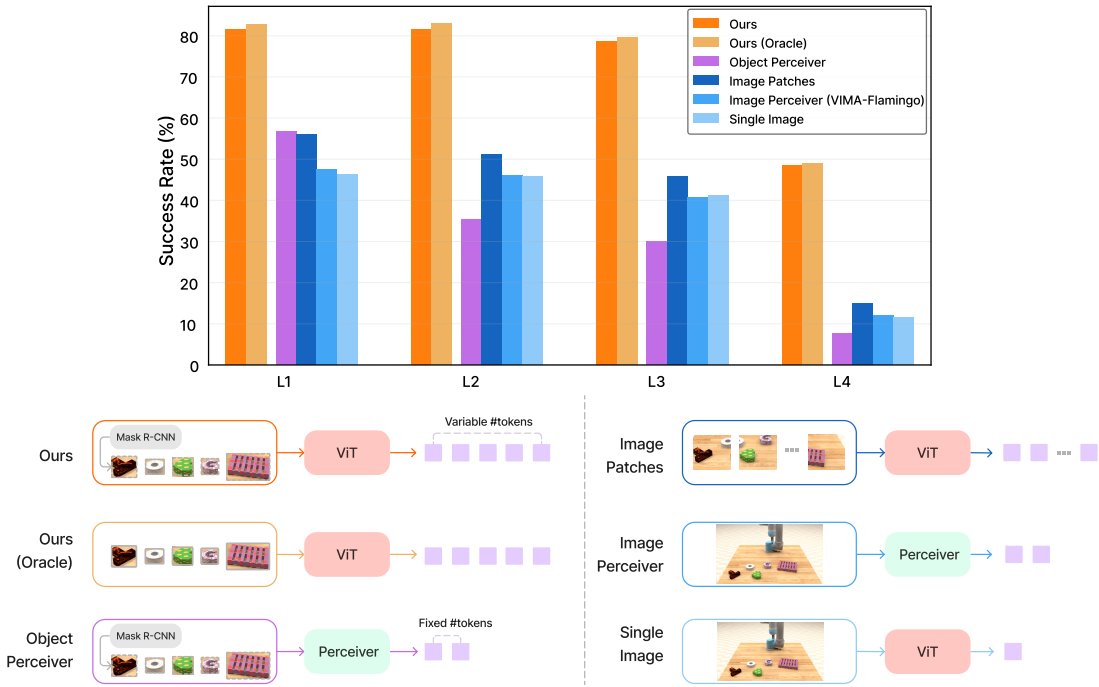}}
    \caption{\textbf{Ablation on visual tokenizers.} We compare the performance of \vima-200M model across different visual tokenizers. Our proposed object tokens outperform all methods that learn directly from raw pixels, and \textit{Object Perceiver} that downsamples the object sequence to a fixed number of tokens.}
    \label{fig:ablation_input_process}
\end{figure*}

\para{Data Scaling.}
Next we investigate how different methods scale with varying dataset sizes. We compare model performance at $0.1\%$, $1\%$, $10\%$ and full imitation learning dataset provided in \vimabench (Fig.~\ref{fig:scalability}).
Note that to ensure all methods are fairly pre-trained on the same amount of data, we initialize baseline variants that directly learn from raw pixels with MVP pre-trained ViT~\citep{xiao2022masked,Radosavovic2022}. It is further MAE fine-tuned~\citep{he2021mae}, using the \emph{same} in-domain data as for the Mask R-CNN object detector. See Appendix, Sec.~\ref{supp:sec:data_scalability} for detailed setup.
\vima is extremely sample efficient and, with just $1\%$ of the data, can achieve performance similar to baseline methods trained with $10\times$ more data on L1 and L2 levels of generalization. In fact, for L4 we find that with just $1\%$ of training data, \vima already surpasses other variants trained with \textit{entire} dataset. Finally, across all levels with just $10\%$ of the data, \vima can outperform other architectures trained with the full dataset by a significant margin. We hypothesize that the data efficiency can be attributed to the object-centric representation employed in the \vima recipe, which is less prone to overfitting than learning directly from pixels in the low-data regime. This is consistent with findings from \citet{sax2018midlevelvision}, which demonstrates that embodied agents conditioned on mid-level visual representations tend to be significantly more sample-efficient than end-to-end control from raw pixels.

\para{Progressive Generalization.}
Finally, we compare the relative performance degradation as we test the models on progressively challenging zero-shot evaluation levels without further fine-tuning (Fig.~\ref{fig:performance_decrease}). Our method exhibits a minimal performance regression, especially between $L1 \rightarrow L2$ and $L1 \rightarrow L3$. In contrast, the baselines can degrade as much as $20\%$, particularly in more difficult generalization scenarios. Although all methods degrade significantly when evaluated on $L4$ (\textit{Novel Tasks}), the performance drop for \vima is only \textit{half} as severe as all other baselines. These results suggest that \vima has developed a more generalizable policy and robust representations than the alternative approaches.

\subsection{Ablation Studies}
\label{sec:experiments-ablation}

Through extensive experiments, we ablate different design choices in \vima and study their impact on robot decision making.
We focus on four aspects:
visual tokenization, prompt conditioning, prompt-encoding language models, and policy robustness against distractions and corruptions.

\begin{figure*}[t]
    \centering
    \makebox[\textwidth][c]{\includegraphics[width=0.9\textwidth]{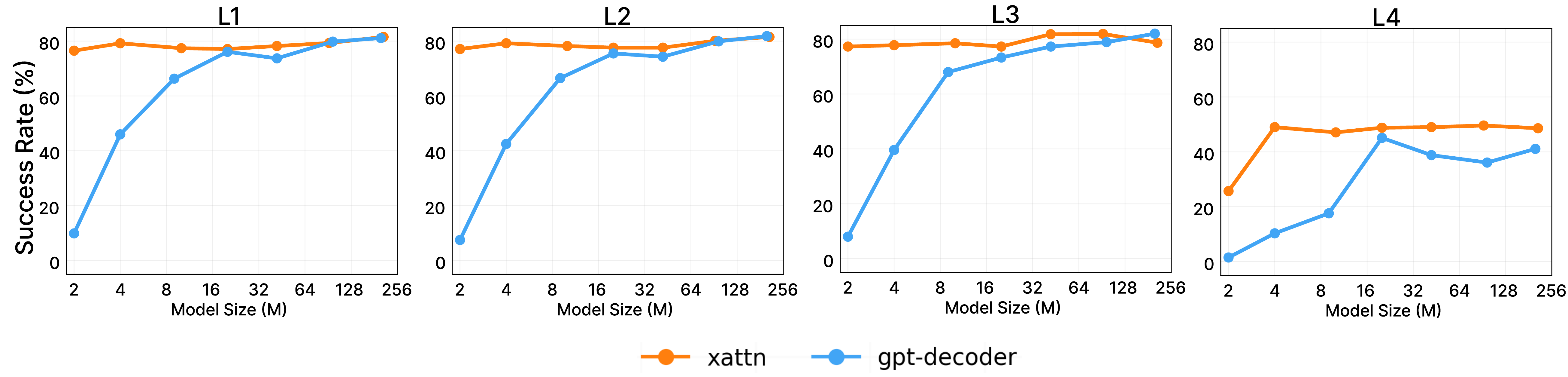}}
    \caption{\textbf{Ablation on prompt conditioning.} We compare our method (\textit{xattn}: cross-attention prompt conditioning) with a vanilla transformer decoder (\textit{gpt-decoder}) across different model sizes. Cross-attention is especially helpful in low-parameter regime and for harder generalization tasks.}
    \label{fig:ablation_seq_conditioning}
\end{figure*}
\para{Visual Tokenization.}
As explained in Sec.~\ref{sec:method}, \vima processes the prompt and observation images into a variable number of object tokens with a domain fine-tuned Mask R-CNN implementation. How important is this particular choice of visual tokenizer? We study 5 different variants and empirically evaluate their 4 levels of generalization performance on \vimabench. 
1) \textbf{Ours (Oracle)}: instead of using Mask R-CNN, we directly read out the ground-truth bounding box from the simulator. In other words, we use a perfect object detector to estimate the upper bound on the performance of this study;
2) \textbf{Object Perceiver}: we apply a Perceiver module to convert the variable number of objects detected in each frame to a \textit{fixed} number of tokens. Perceiver is more computationally efficient because it reduces the average sequence length;
3) \textbf{Image Perceiver}: the same architecture as the \textit{Perceiver Resampler} in \vima-Flamingo, which converts an image to a small, fixed number of tokens; 
4) \textbf{Image patches}: following \vima-Gato, we divide an RGB frame into square patches, and extract ViT embedding tokens. The number of patches is greater than the output of Image Perceiver;
5) \textbf{Single image}: \vima-GPT's tokenizer, which encodes one image into a single token. 

Fig.~\ref{fig:ablation_input_process} shows the ablation results. We highlight a few findings. First, we note that our Mask R-CNN detection pipeline (Appendix, Sec.~\ref{supp:sec:mask_rcnn}) \textbf{incurs a minimal performance loss} compared to the oracle bounding boxes, thanks to the object augmentation (Sec.~\ref{sec:method}) that boosts robustness during training. 
Second, tokenizing from raw pixels (Image Perceiver, patches, or single embedding) consistently underperforms our object-centric format. We hypothesize that these tokenizers have to allocate extra internal capacity to parse the objects from low-level pixels, which likely impedes learning. \citet{sax2018midlevelvision} echoes our finding that using mid-level vision can greatly improve agent generalization compared to an end-to-end pipeline. 
Third, even though \textit{Ours} and \textit{Object Perceiver} both use the same object bounding box inputs, the latter is significantly worse in decision making. We conclude that it is important to directly pass the \textbf{variable-length object sequence} to the robot controller rather than downsampling to a fixed number of tokens.

\para{Prompt Conditioning.}
\vima conditions the robot controller (decoder) on the encoded prompt by cross-attention. A simple alternative is to concatenate the prompt $\mathcal{P}$ and interaction history $\mathcal{H}$ into one big sequence, and then apply a decoder-only transformer like GPT~\citep{radford2018gpt} to predict actions. 
In this ablation, we keep the object tokenizer constant and only switch the conditioning mechanism to causal sequence modeling. Note that this variant is conceptually ``\mbox{\vima}-Gato with object tokens''. Fig.~\ref{fig:ablation_seq_conditioning} shows the comparison of \vima (\texttt{xattn}) and the \texttt{gpt-decoder} variant across 4 generalization levels. While the variant achieves comparable performance in larger models, cross-attention still dominates in the small-capacity range and generalizes better in the most challenging L4 (\textit{Novel Task}) setting. Our hypothesis is that cross-attention helps the controller stay better focused on the prompt instruction at each interaction step. This bears a resemblance to the empirical results in \citet{sanh2021t0,wang2022languagearchstudy}, which show that well-tuned encoder-decoder architectures can outperform GPT-3 in zero-shot generalization.

\para{Prompt Encoding.}
We vary the size of the pre-trained T5 encoder to study the effect of prompt encoding. We experiment with three T5 capacities: \texttt{small} (30M), \texttt{base} (111M), and \texttt{large} (368M). We further fix the parameter count of the decision-making part to be 200M. For all T5 variants, we fine-tune the last two layers and freeze all other layers. We find no significant difference among the variants (Appendix, Sec.~\ref{supp:sec:t5}), thus we set \texttt{base} as default for all our models.

\para{Policy Robustness.}
We study the policy robustness against increasing number of distractors and corrupted task specifications, including incomplete prompts (randomly masking out words with \texttt{<UNK>} token) and corrupted prompts (randomly swapping words, which could have changed the task meaning altogether).
See Appendix, Sec.~\ref{supp:sec:robustness} for exact setup and results. \vima exhibits minimal performance degradation with increased distractors and minor decrease with corrupted prompts. 
We attribute this robustness to the high-quality pre-trained T5 backbone.

\section{Related Work}
\label{sec:related}
\para{Multi-Task Learning by Sequence Modeling.}
Transformers~\citep{vaswani2017attention} have enabled task unification across many AI domains~\citep{brown2020gpt3,chen2022pix2seq,chen2022pix2seqv2,lu2022unifiedio,wang2022ibeit3}.
For example, in \textbf{NLP},
the Natural Language Decathlon~\citep{mccann2018natural} adopts a consistent question-answering format for a suite of 10 NLP tasks.
T5~\citep{raffel2020t5} unifies all language problems into the same text-to-text format. GPT-3~\citep{brown2020gpt3} and Megatron~\citep{shoeybi2019megatron} demonstrate emergent behaviours of intuitive task specifications by zero-shot prompting. 
In \textbf{computer vision}, Pix2Seq~\citep{chen2022pix2seqv2} casts many vision problems into a unified sequence format.
Florence~\citep{yuan2021florence}, BiT~\citep{kolesnikov2020bit}, and MuST~\citep{ghiasi2021must} pre-train shared backbone models at scale for general visual representations and transfer them to downstream tasks.
In \textbf{multimodal learning},
Perceiver~\citep{jaegle2021perceiver,jaegle2021perceiverio} proposes an efficient architecture to handle structured inputs and outputs.
Flamingo~\citep{alayrac2022flamingo} and Frozen~\citep{deepmind2021frozen} design a universal API that ingests interleaving sequences of images and text and generates free-form text. Gato~\citep{reed2022gato} is a massively multi-task model across NLP, vision, and embodied agents. Our work is most similar in spirit to Gato, but we focus primarily on enabling an intuitive multimodal prompting interface for a generalist robot agent.

\para{Foundation Models for Embodied Agents.}
Foundation models~\citep{bommasani2021opportunities} have demonstrated strong emergent properties. There are many ongoing efforts to replicate this success for embodied agents~\citep{yang2023foundation}, focusing on 3 aspects.
1) \textbf{Transformer agent architecture}: Decision Transformer and Trajectory Transformer~\citep{chen2021decisiontransformer,janner2021onebigsequence,zheng2022onlinedt,xu2022prompting,xu2023hyperdecision} leverage the powerful self-attention models for sequential decision making. CLIPort~\citep{shridhar2021cliport}, Perceiver-Actor~\citep{shridhar2022perceiveractor}, and RT-1~\citep{brohan2022rt1} apply large transformers to robot manipulation tasks.
BeT~\citep{shafiullah2022behavior} and C-BeT~\citep{cui2022play} design novel techniques to learn from demonstrations with multiple modes with transformers.
2) \textbf{Pre-training for better representations}: MaskViT~\citep{gupta2022maskvit}, R3M~\citep{nair2022r3m}, VIP~\citep{ma2022vip}, and VC-1~\citep{majumdar2023search} pre-train general visual representations for robotic perception. \citet{li2022pretrained} fine-tunes from LLM checkpoints to accelerate policy learning. MineDojo~\citep{fan2022minedojo} and Ego4D \citep{grauman2021ego4d} provide large-scale multimodal databases to facilitate scalable policy training. 
3) \textbf{LLMs for robot learning}: \mbox{SayCan}~\citep{ahn2022saycan} leverages PaLM~\citep{chowdhery2022palm} for zero-shot concept grounding.
\citet{huang2022language}, Inner Monologue~\citep{huang2022inner} and LM-Nav~\citep{shah2022lmnav} apply LLMs to long-horizon robot planning.
PaLM-E~\citep{driess2023palme} is instead a multimodal language model that can be repurposed for sequential robotic manipulation planning.
Ours differs from these works in our novel multimodal prompting formulation, which existing LLMs do not easily support.  

\para{Robot Manipulation and Benchmarks.}  
A wide range of robot manipulation tasks require different skills and task specification formats, such as instruction following~\citep{stepputtis2020langintructedmanipulation}, one-shot imitation~\citep{finn2017oneshot,duan2017oneshot}, rearrangement~\citep{batra2020rearrangement}, constraint satisfaction~\citep{brunke2021safe}, and reasoning~\citep{shridhar2020alfred}.
Multiple physics simulation benchmarks are introduced to study the above tasks. For example, 
iGibson~\citep{shen2020igibson,li2021igibson2.0,srivastava2021behavior,li2022behaviork} simulates interactive household scenarios.
Ravens~\citep{zeng2020transporter} and Robosuite~\citep{zhu2020robosuite,fan2021secant} design various tabletop manipulation tasks with realistic robot arms.
CALVIN~\mbox{\citep{mees2021calvin}} develops long-horizon language-conditioned tasks.
Meta-World~\citep{yu2019metaworld} is a widely used simulator benchmark studying robotics manipulation with tabletop settings.
CausalWorld~\citep{ahmed2021causalworld} is a benchmark for causal structure and transfer learning in manipulation, requiring long-horizon planning and precise low-level motor control.
AI2-THOR~\citep{rhsani2021ManipulaTHOR,deitke2022procthor} is a framework that supports visual object manipulation and procedural generation of environments.
Our \vimabench is the first robot learning benchmark to support multimodal-prompted tasks. We also standardize the evaluation protocol to systematically measure an agent's generalization capabilities. 

\textbf{An extended review can be found in Appendix, Sec.~\ref{supp:sec:extended_related_work}.}
\section{Conclusion}

In this work, we introduce a novel \textit{multimodal} prompting formulation that converts diverse robot manipulation tasks into a uniform sequence modeling problem.
We instantiate this formulation in \vimabench, a diverse benchmark with multimodal tasks and systematic evaluation protocols for generalization.
We propose \vima, a conceptually simple transformer-based agent capable of solving tasks such as visual goal reaching, one-shot video imitation, and novel concept grounding with a single model.
Through comprehensive experiments, we show that \vima exhibits strong model scalability and zero-shot generalization. Therefore, we recommend our agent design as a solid starting point for future work.

\section*{Acknowledgement}
We are extremely grateful to Shyamal Buch, Jonathan Tremblay, Ajay Mandlekar, Chris Choy, De-An Huang, Silvio Savarese, Fei Xia, Josiah Wong, Abhishek Joshi, Soroush Nasiriany, and many other colleagues and friends for their helpful feedback and insightful discussions.
We also thank the anonymous reviewers for offering us highly constructive advice and kind encouragement during the review period.
NVIDIA provides the necessary computing resource and infrastructure for this project. This work is done during Yunfan Jiang and Guanzhi Wang's internships at NVIDIA. Guanzhi Wang is supported by the Kortschak fellowship in Computing and Mathematical Sciences at Caltech.

\bibliography{bibliography}
\bibliographystyle{icml2023}

\newpage
\appendix
\onecolumn
\renewcommand{\thefigure}{A.\arabic{figure}}
\setcounter{figure}{0}

\section{Simulator Details}
\label{supp:sec:simulator}

\begin{figure}[h]
    \centering
    \makebox[\textwidth][c]{\includegraphics[width=0.8\textwidth]{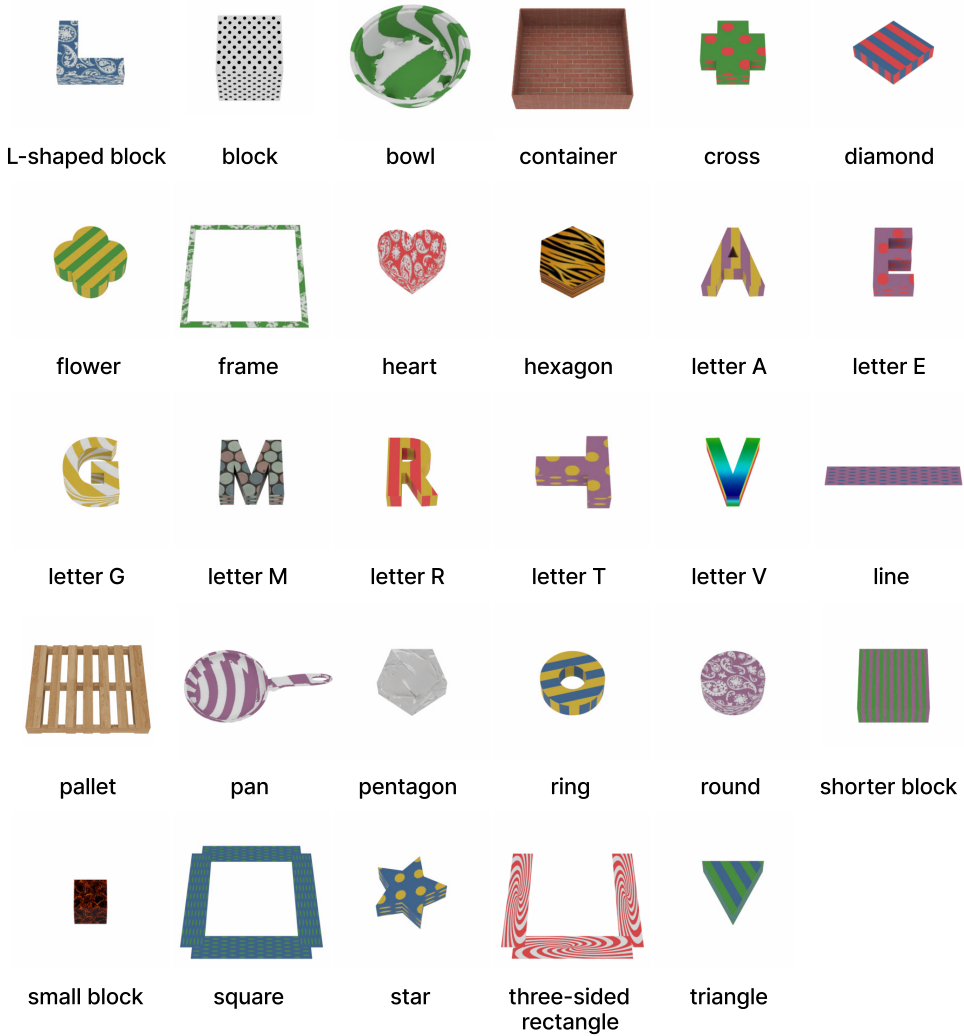}}
    \caption{\textbf{Object Gallery in \vimabench} textured with random textures. Bowl and pan are from Google Scanned Objects~\citep{downs2022google}, while others are from Ravens~\citep{zeng2020transporter}.}
    \label{supp:fig:objs}
\end{figure}

\begin{figure}[!h]
    \centering
    \makebox[\textwidth][c]{\includegraphics[width=1.0\textwidth]{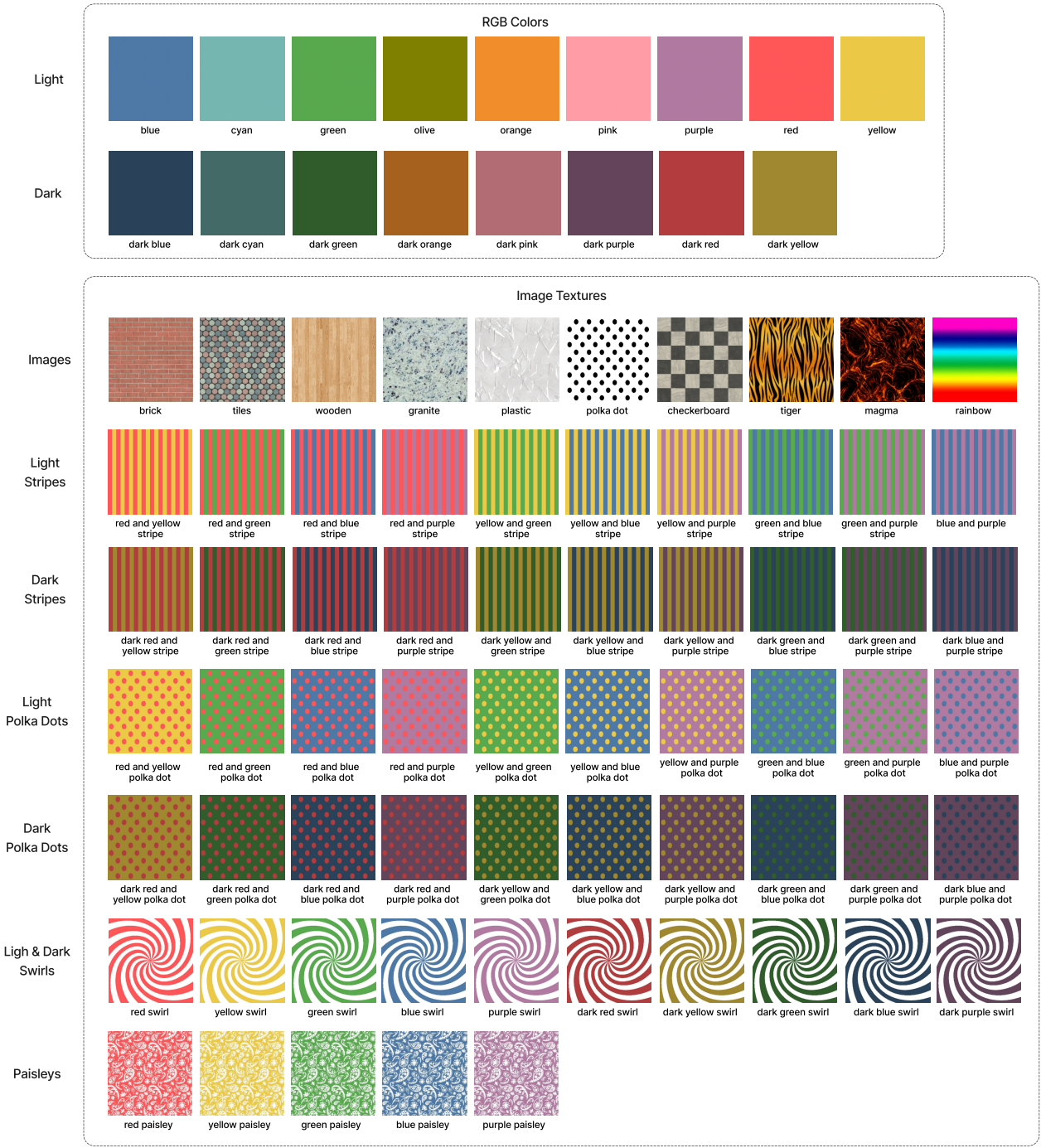}}
    \caption{\textbf{Texture Gallery in \vimabench}. The first row of image-based textures is from Blender Cloud Libraries~\citep{weikert2022blendertexture}, while others are hard-coded.}
    \label{supp:fig:textures}
\end{figure}

We build our \vimabench simulation suite upon the Ravens physics simulator \citep{zeng2020transporter,shridhar2021cliport}. Specifically, it is supported by PyBullet~\citep{coumans2016pybullet} with a Universal Robot UR5 arm. The size of the tabletop workspace is $0.5 \times 1$m.
Our benchmark contains extensible sets of 3D objects and textures. Instantiated from an object-texture combination, all object instances can be rendered as RGB images appeared in multimodal prompts. Figure~\ref{supp:fig:objs} displays all 3D objects. Figure~\ref{supp:fig:textures} displays all textures.

The observation space of \vimabench includes RGB images from both frontal and top-down views. It also includes a one-hot vector $\in \{0, 1\}^2$ to indicate type of the end-effector $\in \{\text{suction cup, spatula}\}$. While a suction cup is equipped in most manipulation tasks, a spatula is used in particular for visual constraint tasks, where an agent is asked to ``wipe'' objects. \vimabench inherits the same action space from \citet{zeng2020transporter} and \citet{shridhar2021cliport}, which consists of primitive actions of ``pick and place'' for tasks with a suction cup as the end effector, or ``push'' for tasks with a spatula. Both primitive actions contain two poses $\in$ $\mathbf{SE}(2)$ specifying target poses of the end effector. For the ``pick and place'' primitive, they represent the pick pose and the place pose. For the ``push'' primitive, they represent the push starting pose and push ending pose.

Similar to prior work~\citep{zeng2020transporter,shridhar2021cliport}, \vimabench provides scripted oracles to generate successful demonstrations for all tasks. We leverage them to construct an offline imitation dataset for behavioral cloning. Given a prompt, these programmed bots can access privileged information, such as the correct object to pick and target location to place.
\clearpage
\section{Task Suite}
\label{supp:sec:task_suite}

We develop 17 task templates that belong to 6 diverse categories. Thousands of individual task instances and their corresponding multimodal prompts can be procedurally generated from these task templates. 
We use PyBullet~\citep{coumans2016pybullet} as our backend and the default renderer to produce the RGB frames for training data and interactive test environments. For demonstration purpose, we apply the  NVISII~\citep{morrical20nvisii} ray tracing to enhance the visual quality. We elaborate on each task in the following subsections.

\subsection{Simple Object Manipulation}

This task category asks agents to follow basic instructions specified by multimodal prompts.

\begin{figure}[!htbp]
    \centering
    \makebox[\textwidth][c]{\includegraphics[width=1\textwidth]{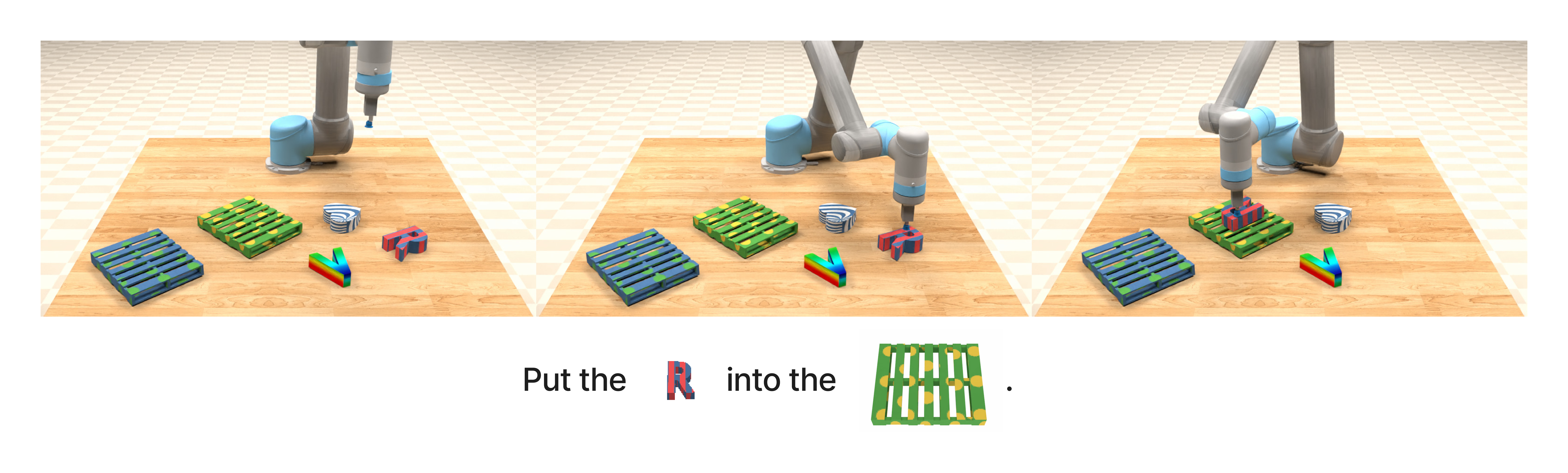}}
    \caption{Simple Object Manipulation: Task 01}
    \label{supp:fig:traj:simple_manipulation}
\end{figure}

\paragraph{\underline{Task 01:}} Pick the specified object(s) and place it (them) into the specified container. 
\begin{itemize}
    \item{\textbf{Prompt:} \texttt{Put the \{object\}$_1$ into the \{object\}$_2$.}}
    \item{\textbf{Description:} The image placeholder \texttt{\{object\}}$_1$ is the object to be picked and the \texttt{\{object\}}$_2$ is the container object. The agent requires to recognize the objects with the correct color-shape combinations. To extend the difficulties, it supports more than one object to be picked or placed. For example, the prompt ``\texttt{Put the \{object\}$_1$ and \{object\}$_2$ into the \{object\}$_3$}'' asks to pick two different objects and place into a target container. 
    We uniformly sample different color-shape combos for objects to be picked and containers.}
    \item{\textbf{Success Criteria:} All specified object(s) to pick are within the bounds of the container object(s), with specified shapes and textures provided in the prompt.}
    \item{\textbf{Oracle Trajectory:} Shown in Fig.~\ref{supp:fig:traj:simple_manipulation} with its multimodal prompt.}
\end{itemize}

\begin{figure}[!htbp]
    \centering
    \makebox[\textwidth][c]{\includegraphics[width=1\textwidth]{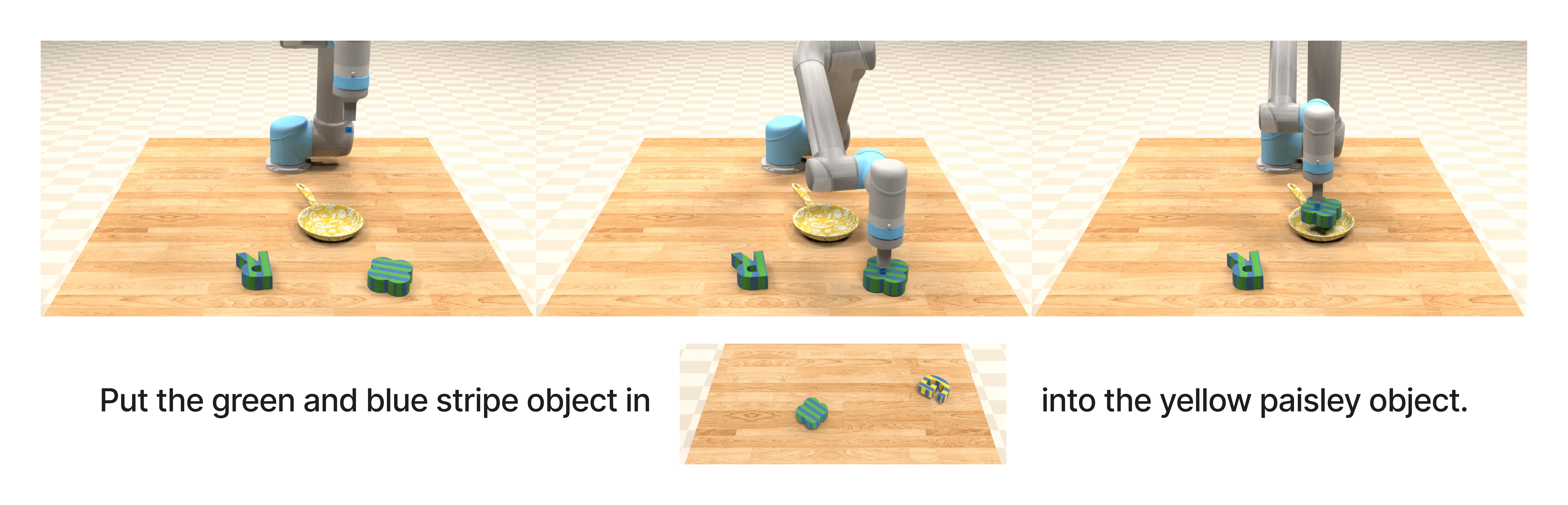}}
    \caption{Simple Object Manipulation: Task 02}
    \label{supp:fig:traj:scene_understanding}
\end{figure}

\paragraph{\underline{Task 02:}} In the workspace, put the objects with a specified texture shown in the scene image in the prompt into container object(s) with a specified color. This task requires the agent to find the correct object to manipulate by grounding the textural attributes from both natural language descriptions and the visual scene images.
\begin{itemize}
    \item{\textbf{Prompt:} \texttt{Put the \{texture\}$_1$ object in \{scene\} into the \{texture\}$_2$ object.}}
    \item{\textbf{Description:} The text placeholder \texttt{\{texture\}}$_1$ and \texttt{\{texture\}}$_2$ are sampled textures for objects to be picked and the container objects, respectively. The number of dragged objects with the same texture can be varied. \texttt{\{scene\}} is the workspace-like image placeholder. There is a designated number of distractors with different textures (and potentially different shapes) in the scene. For each distractor in the workspace, it has $50\%$ chance to be either dragged or container distractor object with different textures from those specified in the prompt.}
    \item{\textbf{Success Criteria:} All objects in the workspace with \texttt{\{texture\}}$_1$ are within the bounds of the container object with \texttt{\{texture\}}$_2$.}
    \item{\textbf{Oracle Trajectory:} Shown in Fig.~\ref{supp:fig:traj:scene_understanding} with its multimodal prompt.}
\end{itemize}

\begin{figure}[!htbp]
    \centering
    \makebox[\textwidth][c]{\includegraphics[width=0.9\textwidth]{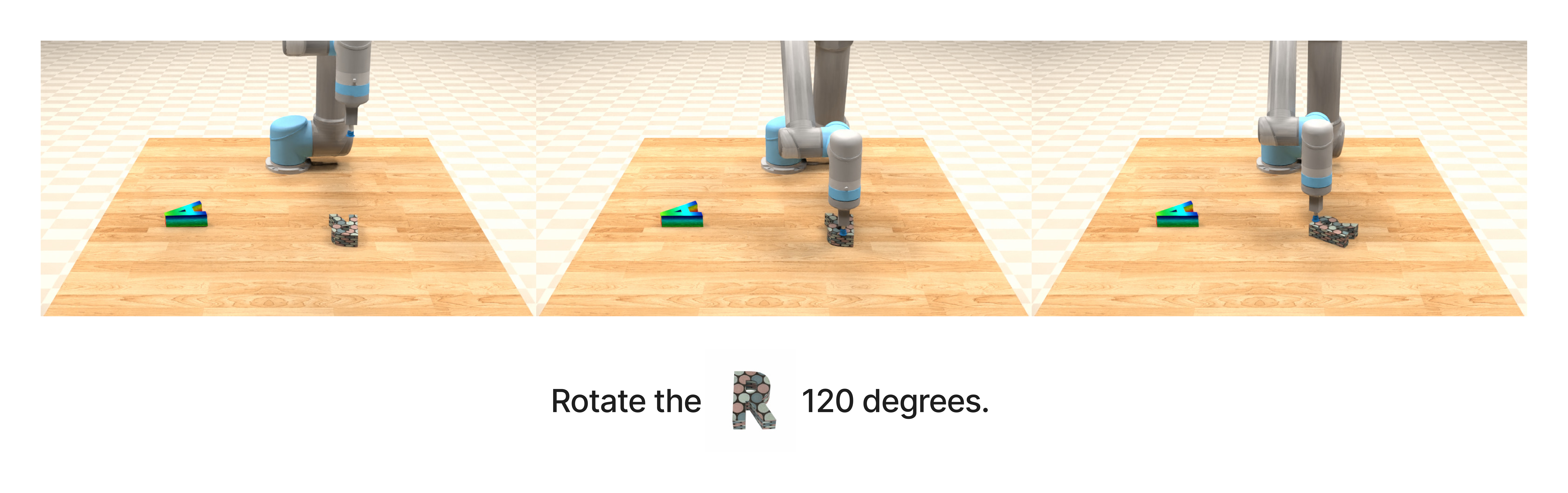}}
    \caption{Simple Object Manipulation: Task 03}
    \label{supp:fig:traj:rotate}
\end{figure}

\paragraph{\underline{Task 03:}} Rotate objects clockwise by certain degrees along $z$-axis. Only rotationally asymmetric objects are considered in this task.
\begin{itemize}
    \item{\textbf{Prompt:} \texttt{Rotate the \{object\}$_1$ \{angles\} degrees.}}
    \item{\textbf{Description:} The agent is required to rotate all objects in the workspace specified by the image placeholder \texttt{\{object\}}$_1$. There are also objects with different color-shape combinations in the workspace as distractors. \texttt{\{angles\}} is the sampled degree that needs to be rotated. A target angle is sampled from $30^{\circ}$, $60^{\circ}$, $90^{\circ}$, $120^{\circ}$, and $150^{\circ}$.}
    \item{\textbf{Success Criteria:} The position of the specified object matches its original position, and the orientation matches the orientation after rotating specific angles.}
    \item{\textbf{Oracle Trajectory:} Shown in Fig.~\ref{supp:fig:traj:rotate} with its multimodal prompt.}
\end{itemize}

\clearpage
\subsection{Visual Goal Reaching}

This task category requires agents to manipulate objects in the workspace to reach goal states represented as images shown in prompts.

\begin{figure}[!htbp]
    \centering
    \makebox[\textwidth][c]{\includegraphics[width=0.9\textwidth]{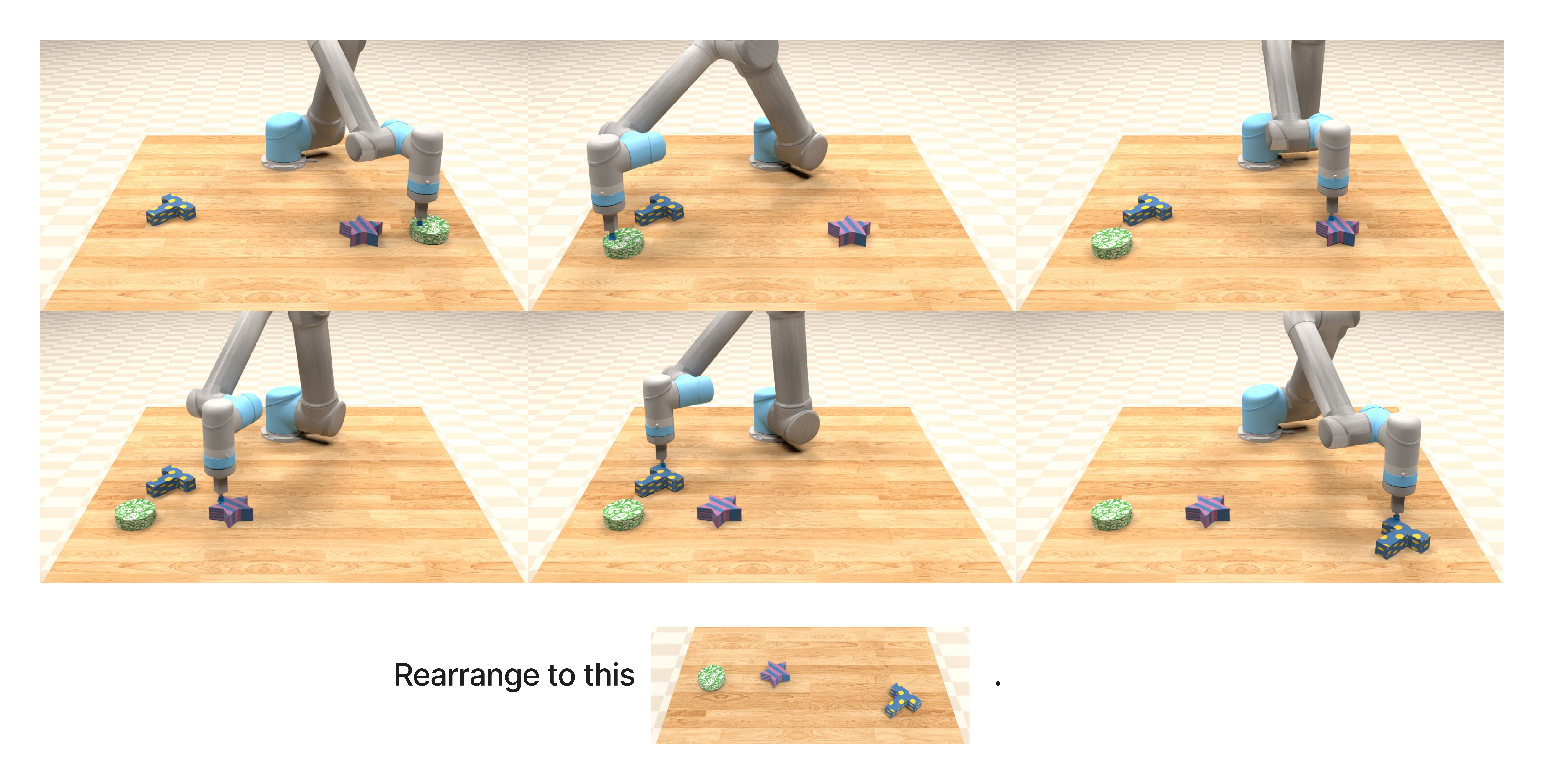}}
    \caption{Visual Goal Reaching: Task 04}
    \label{supp:fig:traj:rearrange}
\end{figure}

\paragraph{\underline{Task 04:}} Rearrange target objects in the workspace to match goal configuration shown in prompts. Note that to achieve the goal configuration, distractors may need to be moved away first.
\begin{itemize}
    \item{\textbf{Prompt:} \texttt{Rearrange to this \{scene\}.}}
    \item{\textbf{Description:} Objects in the scene placeholder \texttt{\{scene\}} are target objects to be manipulated and rearranged. In the workspace, the same target objects are spawned randomly, potentially with distractors randomly spawned as well. With a pre-defined distractor conflict rate, the position of each distractor has this probability to occupy the position of any target object such that the rearrangement can only succeed if moving away that distractor first.}
    \item{\textbf{Success Criteria:} The configuration of target objects in the workspace matches that specified in the prompt.}
    \item{\textbf{Oracle Trajectory:} Shown in Fig.~\ref{supp:fig:traj:rearrange} with its multimodal prompt.}
\end{itemize}

\begin{figure}[ht]
    \centering
    \makebox[\textwidth][c]{\includegraphics[width=1\textwidth]{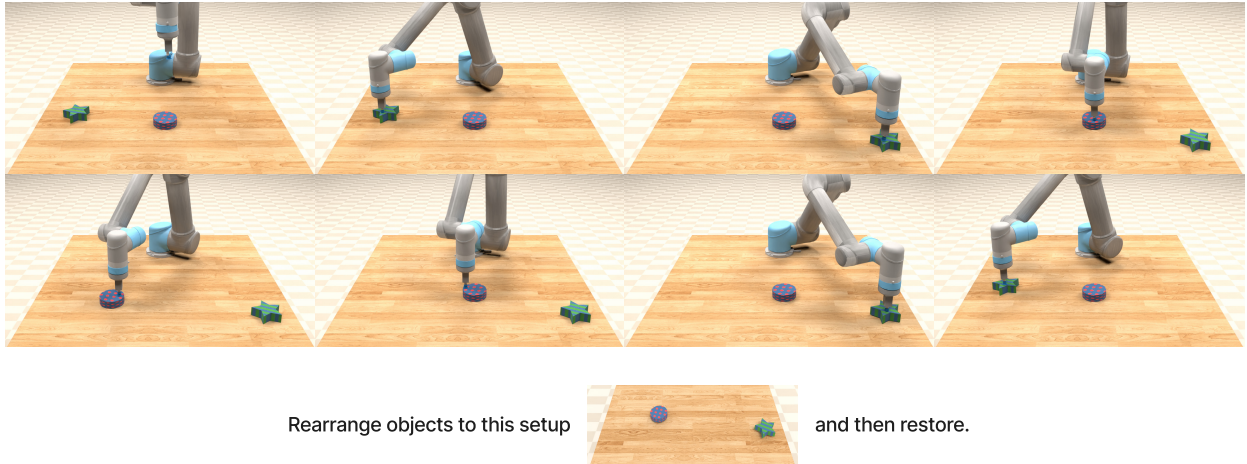}}
    \caption{Visual Goal Reaching: Task 05}
    \label{supp:fig:traj:rearrange_then_restore}
\end{figure}

\paragraph{\underline{Task 05:}} Extend the task \textit{04} by requiring the agent to restore rearranged objects to the initial setup after the ``rearranging'' phase.
\begin{itemize}
    \item{\textbf{Prompt:} \texttt{Rearrange objects to this setup \{scene\} and then restore.}}
    \item{\textbf{Description:} Same as the task \textit{04}, except introducing the instruction ``restore''.}
    \item{\textbf{Success Criteria:} Meet the success criteria of the task \textit{04}, and then within the allowed max steps restore all target objects to their initial configurations.}
    \item{\textbf{Oracle Trajectory:} Shown in Fig.~\ref{supp:fig:traj:rearrange_then_restore} with its multimodal prompt.} 
\end{itemize}

\subsection{Novel Concept Grounding}
\label{supp:sec:novel_concept_grounding}

This task category requires agents to ground new concepts of adjectives, nouns, or verbs via visual perception and language understanding.
Similar task design can be found in prior work~\citep{hill2021grounded}. Completing these tasks are challenging, because the model should a) first understand prompts with interleaved texts, images, and even video frames; b) quickly internalize new concepts that are different across task instances, which even tests the ability to meta-learn; and c) do complicated reasoning such as comparing between ``taller'' vs ``less taller'' vs ``shorter'' and then ground this reasoning into the robot action space.

Prompts consist of two parts: a definition part followed by an instruction part. In the definition part, novel concepts are defined by multimodal illustrations with multiple support examples. In the instruction part, agents are asked to achieve the goal by properly applying concepts from the definition part. The assignment of dummy object names is varied and independent for each task instance such that tasks can only be solved if the agent applies the reasoning correctly. This ability is also referred to as \textit{fast-mapping}~\citep{heibeck1987word}.

\begin{figure}[!htbp]
    \centering
    \makebox[\textwidth][c]{\includegraphics[width=1\textwidth]{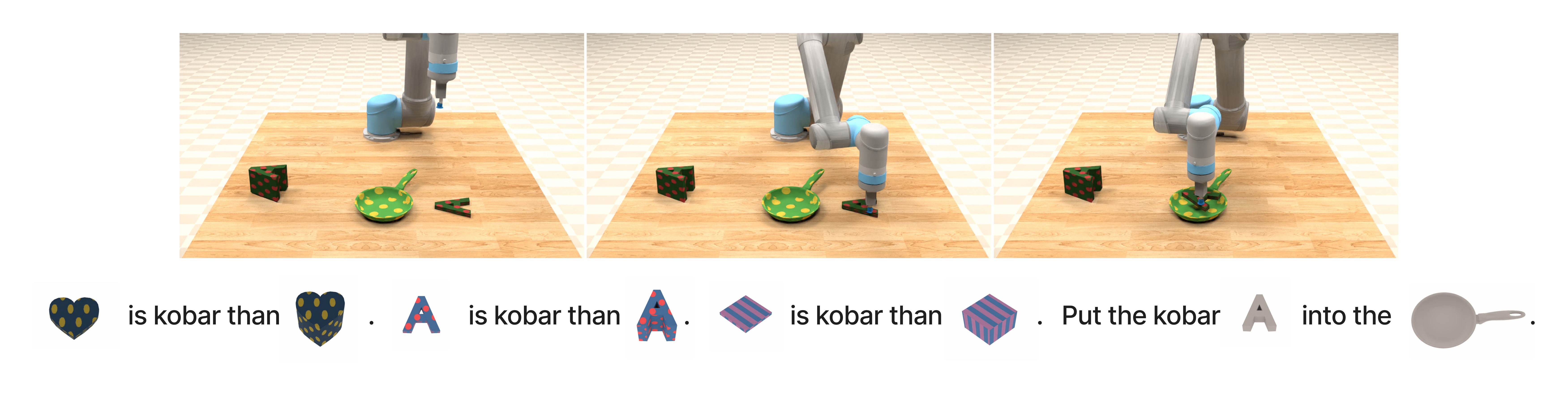}}
    \caption{Novel Concept Grounding: Task 06}
    \label{supp:fig:traj:novel_adj}
\end{figure}

\paragraph{\underline{Task 06:}}  
Ground comparative adjectives by comparing the size or the textural saturation of objects and manipulating the correct object(s) instructed in the prompt. 
\begin{itemize}
    \item{\textbf{Prompt:} \texttt{\{demo\_object\}$_1$ is \{novel\_adj\} than \{demo\_object\}$_2$. Put the \{adv\} \{novel\_adj\} \{object\}$_1$ into the \{object\}$_2$.}}
    \item{\textbf{Description:} The sampled adjective \texttt{\{novel\_adj\}} is a dummy adjective placeholder for agent to ground. By default, the novel adjective set is \texttt{\{daxer, blicker, modier, kobar\}}. The real meaning can be related to size (smaller/larger) or textural saturation (lighter/darker texture). The image placeholders \texttt{\{demo\_object\}}$_1$ and \texttt{\{demo\_object\}}$_2$ illustrate how the novel adjective is defined. For example, if the real comparison is "taller", then the sampled object in \texttt{\{demo\_object\}}$_1$ is taller than \texttt{\{demo\_object\}}$_2$. The choices of the novel adjective and the real meaning are independently sampled for different task instances. For the instruction part, this task is similar to task \textit{01}, where the agent is required to pick the specified object(s) with the novel adjective attribute and then place it into the specified container object. To avoid revealing the correct object to manipulate, we use a neutral texture for objects appeared in the instruction part.}
    \item{\textbf{Success Criteria:} All target objects with the specified adjective attribute are within the bounds of the specified container object.}
    \item{\textbf{Oracle Trajectory:} Shown in Fig.~\ref{supp:fig:traj:novel_adj} with its multimodal prompt.}
\end{itemize}

\begin{figure}[!htbp]
    \centering
    \makebox[\textwidth][c]{\includegraphics[width=1\textwidth]{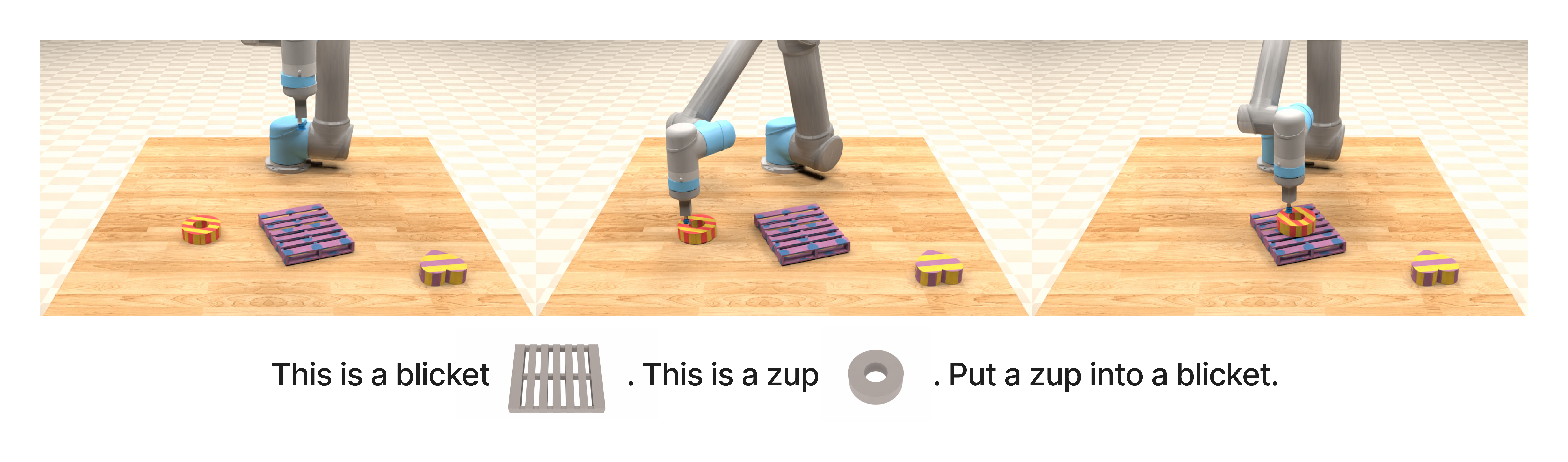}}
    \caption{Novel Concept Grounding: Task 07}
    \label{supp:fig:traj:novel_noun}
\end{figure}

\paragraph{\underline{Task 07:}} 
Orthogonal to task \textit{06} by requiring to learn mappings of novel nouns.
\begin{itemize}
    \item \textbf{Prompt:} $\texttt{This is a \{novel\_name\}}_1 \texttt{ \{object\}}_1 \texttt{. This is a \{novel\_name\}}_2 $ $\texttt{ \{object\}}_2 \texttt{. Put \{novel\_name\}}_1 \texttt{ into a \{novel\_name\}}_2.$
    \item \textbf{Description:} Novel noun words are defined with the text placeholders $\texttt{\{novel\_name\}}_1$  and $\texttt{\{novel\_name\}}_2$, following their image placeholders $\texttt{\{object\}}_1$ and $\texttt{\{object\}}_2$, for the target object and container object, respectively. Novel nouns are sampled from \texttt{\{dax, blicket, wug, zup\}}. In the instruction part, objects are expressed as novel nouns defined in the previous definition part. Distractors are defined the same as task \textit{01}. 
    \item \textbf{Success Criteria:} All target object(s) are within the bounds of the container object(s).
    \item \textbf{Oracle Trajectory:}  Shown in Fig.~\ref{supp:fig:traj:novel_noun} with its multimodal prompt. 
\end{itemize}

\begin{figure}[!htbp]
    \centering
    \makebox[\textwidth][c]{\includegraphics[width=1\textwidth]{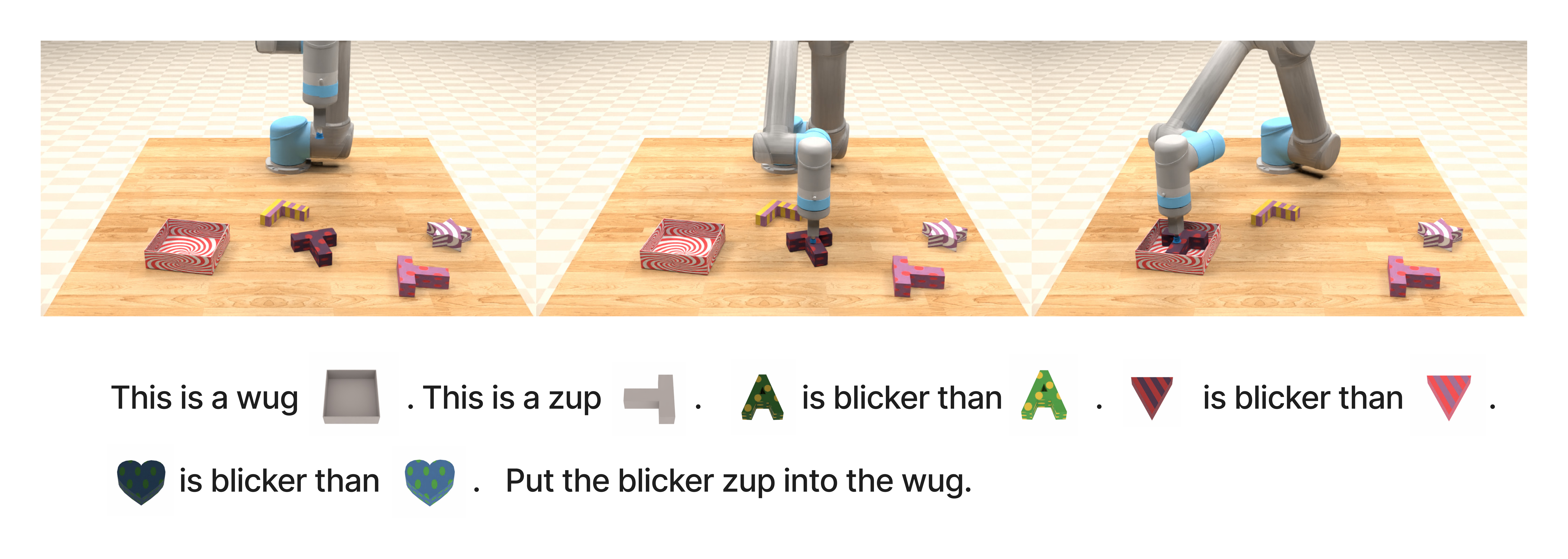}}
    \caption{Novel Concept Grounding: Task 08}
    \label{supp:fig:traj:novel_adj_and_noun}
\end{figure}

\paragraph{\underline{Task 08:}} 
Combination of tasks \textit{06} and \textit{07}.
\begin{itemize}
    \item \textbf{Prompt:} $\texttt{This is a \{novel\_name\}}_1 \texttt{ \{object\}}_1 \texttt{. This is a \{novel\_name\}}_2 $ $ \texttt{ \{object\}}_2  \texttt{. \{demo\_object\}}_1 \texttt{ is \{adj\} than \{demo\_object\}}_2.\texttt{ Put the  } $ $\texttt{\{adv\} \{novel\_adj\} \{novel\_name\}}_1 \texttt{ into the \{novel\_name\}}_2 \texttt{.}$
    \item \textbf{Description:} See task description for task \textit{06} and task \textit{07}. 
    \item \textbf{Success Criteria:} Similar as tasks \textit{06} and \textit{07}.
    \item \textbf{Oracle Trajectory:} Shown in Fig.~\ref{supp:fig:traj:novel_adj_and_noun} with its multimodal prompt. 
\end{itemize}

\begin{figure}[!htbp]
    \centering
    \makebox[\textwidth][c]{\includegraphics[width=1\textwidth]{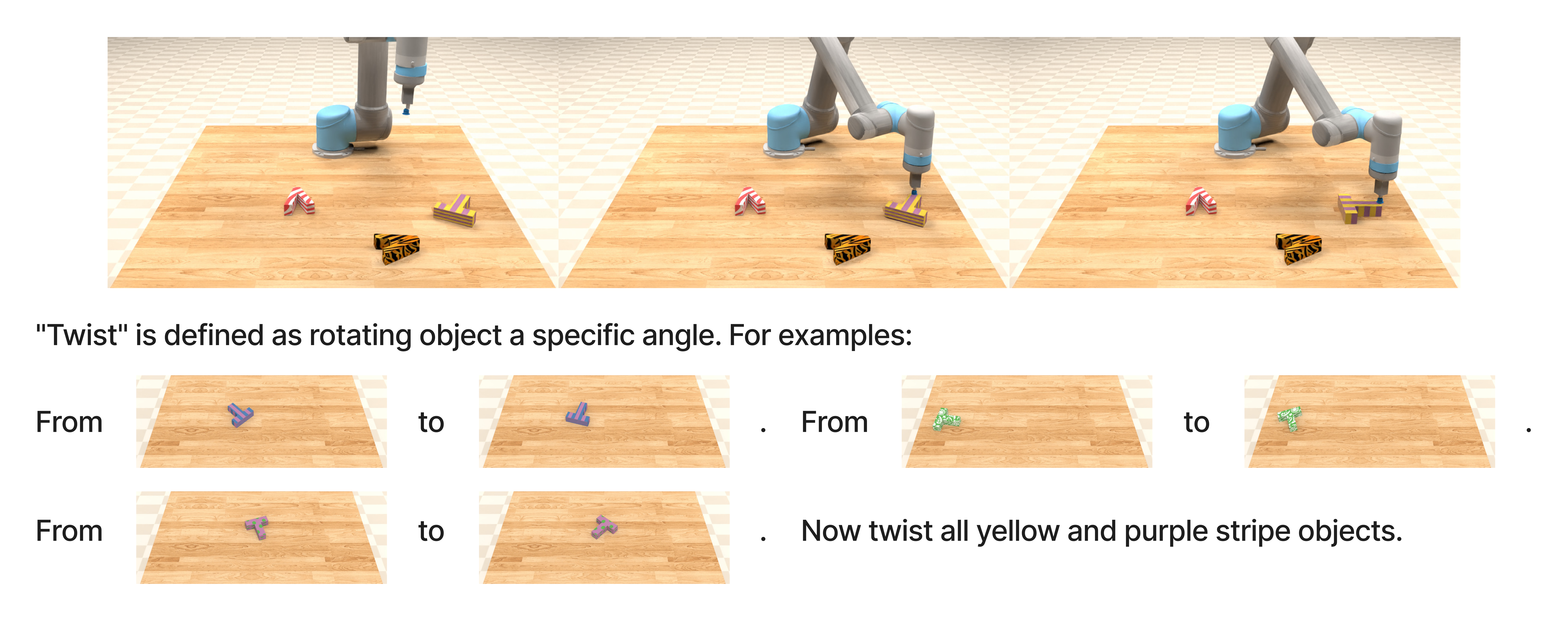}}
    \caption{Novel Concept Grounding: Task 09}
    \label{supp:fig:traj:twist}
\end{figure}

\paragraph{\underline{Task 09:}}  
A novel verb ``twist" is defined as rotating a specific angle illustrated by several examples. This task is similar to task \textit{03}, but it requires the agent to infer what is the exact angle to rotate from the prompt and to ground novel verbs that are semantically similar but different in exact definitions.
\begin{itemize}
    \item \textbf{Prompt:} $\texttt{"Twist" is defined as rotating object a specific angle. }$ $ \texttt{For examples: From \{before\_twist\}}_i$ \texttt{to} $\texttt{\{after\_twist\}}_i.$ $\texttt{ Now twist}$ $\texttt{ all \{texture\} objects.}$
    \item \textbf{Description:} Both $\texttt{\{before\_twist\}}_i$ and $\texttt{\{after\_twist\}}_i$ are scene placeholders where $\texttt{\{before\_twist\}}_i$ shows a randomly sampled object before ``twisting'' and $\texttt{\{after\_twist\}}_i$  shows the same object pose after ``twisting''. All examples illustrate the same sampled angle to rotate. In the workspace, the target objects have the texture specified by $\texttt{\{texture\}}$ and randomly sampled shapes.    \item \textbf{Success Criteria:} Same as the task \textit{03}.
    \item \textbf{Oracle Trajectory:} Shown in Fig.~\ref{supp:fig:traj:twist} with its multimodal prompt. 
\end{itemize}

\subsection{One-Shot Video Imitation}
\label{supp:sec:one_shot_imitation}
This task category requires agents to imitate motions demonstrated through videos shown in prompts.
We follow prior works~\citep{finn2017oneshot,dasari2020transformers,duan2017oneshot} to formulate the problem by giving one video demonstration (represented as key frames in prompts), then test the learned imitator\mbox{’}s ability to produce target trajectories. This setup is challenging because a) only one demonstration is available to the agent; b) the model needs to understand video frames interleaved with textual instructions; and c) missing correspondences between demonstrations and target trajectories since demonstrations only show partial key frames.

\begin{figure}[!htbp]
    \centering
    \makebox[\textwidth][c]{\includegraphics[width=1\textwidth]{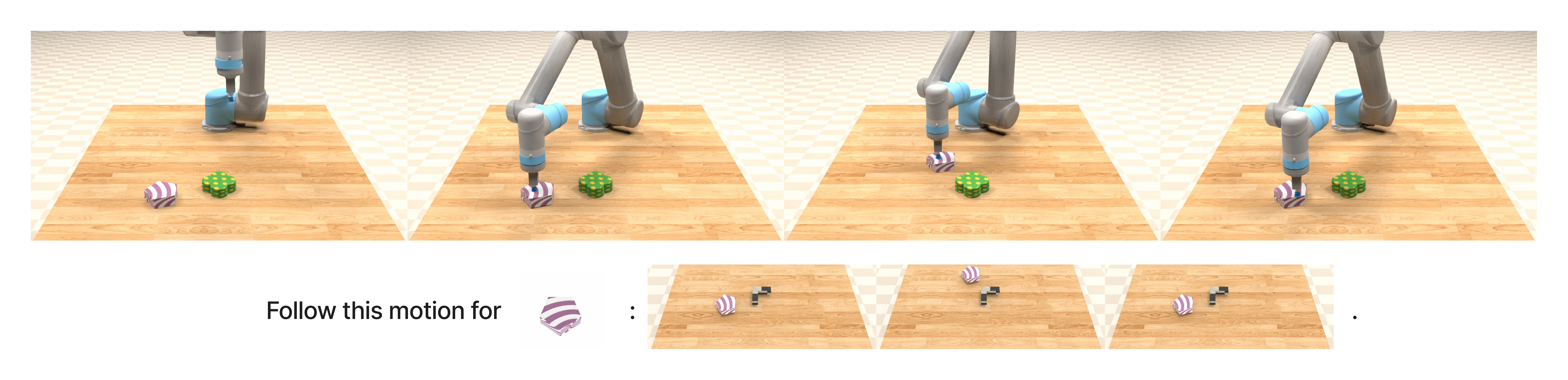}}
    \caption{One-shot video imitation: Task 10}
    \label{supp:fig:traj:follow_motion}
\end{figure}

\paragraph{\underline{Task 10:}}  
Follow motions for specific objects.
\begin{itemize}
    \item \textbf{Prompt:} $\texttt{Follow this motion for \{object\}} \texttt{: \{frame\}}_1 \texttt{...\{frame\}}_i \texttt{...}$ $ \texttt{\{frame\}}_n\texttt{.}$
    \item \textbf{Description:} Image placeholder $\texttt{\{object\}}$ is the target object to be manipulated and $\texttt{\{\{frame\}}_i\}$ is set of workspace-like scene placeholders to represent a video trajectory, where $n$ is the trajectory length. There is an object spawned at the center in both the workspace and the prompt video but with different textures as a distractor. The initial position of the target object matches that in $\texttt{\{frame\}}_1$.
    \item \textbf{Success Criteria:} In each step, the pose of the target object matches the pose in the corresponding video frame. Incorrect manipulation sequences are considered as failures. 
    \item \textbf{Oracle Trajectory:}  Shown in Fig.~\ref{supp:fig:traj:follow_motion} with its multimodal prompt. 
\end{itemize}

\begin{figure}[!htbp]
    \centering
    \makebox[\textwidth][c]{\includegraphics[width=1\textwidth]{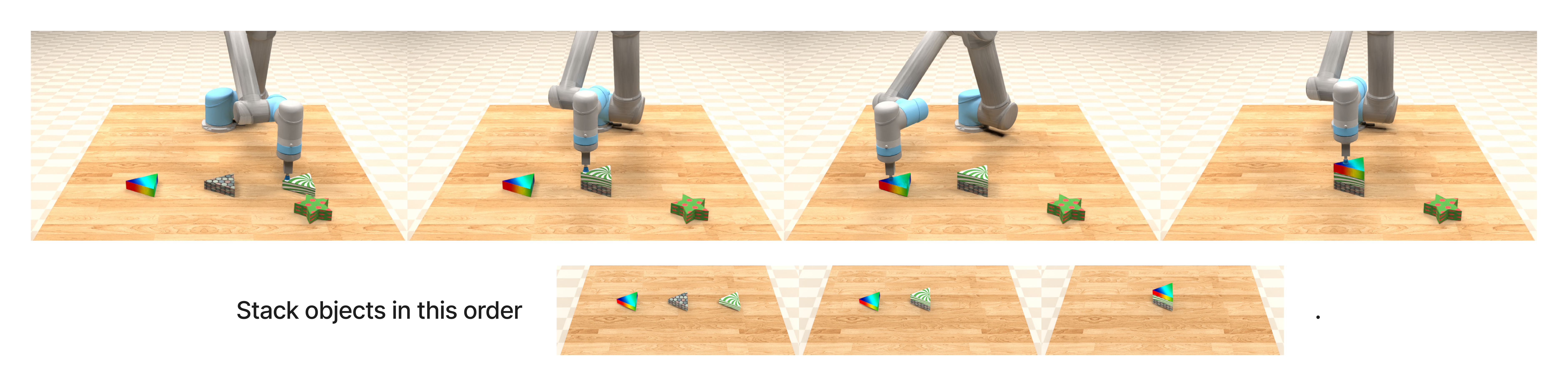}}
    \caption{One-shot video imitation: Task 11}
    \label{supp:fig:traj:follow_order}
\end{figure}

\paragraph{\underline{Task 11:}} 
Stack objects with the order illustrated in the prompt video. 
\begin{itemize}
    \item \textbf{Prompt:} $\texttt{Stack objects in this order \{frame\}}_1 \texttt{...\{frame\}}_i \texttt{...\{frame\}}_n\texttt{.}$
    \item \textbf{Description:} There are multiple objects with the same shape but different textures spawned in the workspace without any stacking initially. Distractor objects with different shapes are spawned in the workspace but not in the prompt video. At each step of the prompt video, one object is stacked over another or put at an empty position.
    \item \textbf{Success Criteria:} Similar as task 10.
    \item \textbf{Oracle Trajectory:} Shown in Fig.~\ref{supp:fig:traj:follow_order} with its multimodal prompt. 
\end{itemize}

\subsection{Visual Constraint Satisfaction}
This task category requires agents to wipe a specific number of objects in the workspace to a goal region while also satisfy the given visual constraint. 

\begin{figure}[!htbp]
    \centering
    \makebox[\textwidth][c]{\includegraphics[width=1\textwidth]{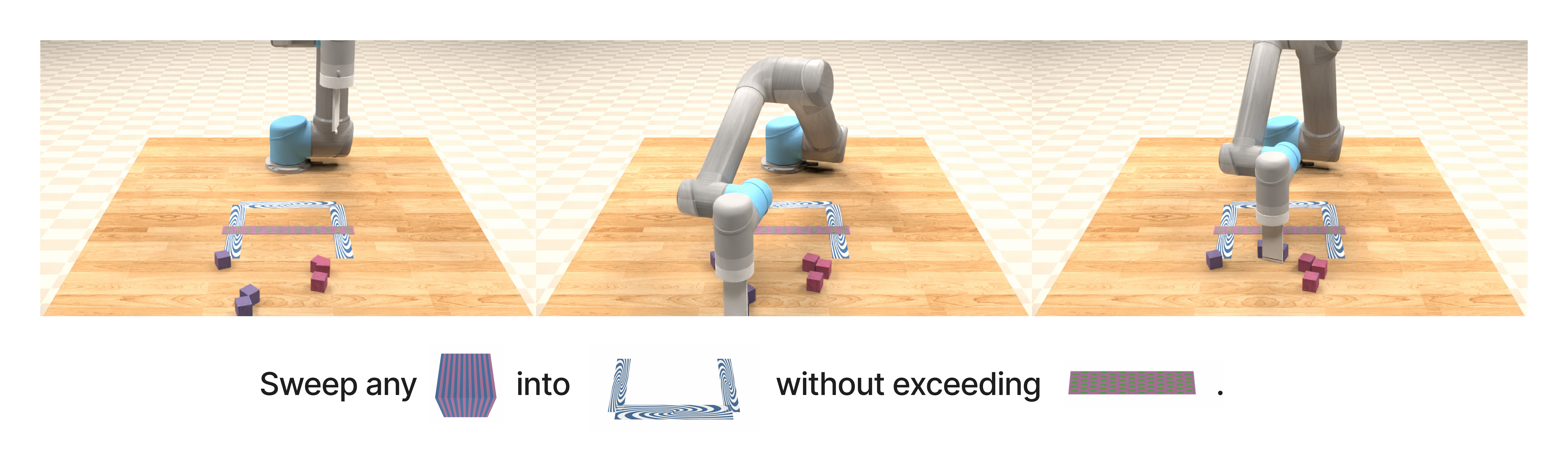}}
    \caption{Visual Constraint Satisfaction: Task 12}
    \label{supp:fig:traj:sweep_without_exceeding}
\end{figure}

\paragraph{\underline{Task 12:}}
Sweep the designated number of objects into a specified region without exceeding the boundary.
\begin{itemize}
    \item \textbf{Prompt:} \texttt{Sweep \{quantifier\} \{object\} into \{bounds\} without exceeding \{constraint\}.} 
    \item \textbf{Description:} \texttt{\{object\}} is the image placeholder of the target object to be swept spawned with a random amount in the workspace. Distractors have the same amount, same shape, but different color from target objects. \texttt{\{quantifier\}} is the text placeholder to determine the target quantity of objects to be wiped, sampled from \texttt{any}, \texttt{one}, \texttt{two}, \texttt{three}, and \texttt{all}. \texttt{\{bounds\}} is the image placeholder for a three-sided rectangle as the goal region. \texttt{\{constraint\}} is the constraint line.
    \item \textbf{Success Criteria:} The exact number of target objects to be swept are all inside the specified region. Potential failure cases include 1) any distractor being wiped into the region, 2) target object exceeding the constraint, or 3) incorrect number of target objects being swept into the goal region.
    \item \textbf{Oracle Trajectory:}  Shown in Fig.~\ref{supp:fig:traj:sweep_without_exceeding} with its multimodal prompt. 
\end{itemize}

\begin{figure}[!htbp]
    \centering
    \makebox[\textwidth][c]{\includegraphics[width=1\textwidth]{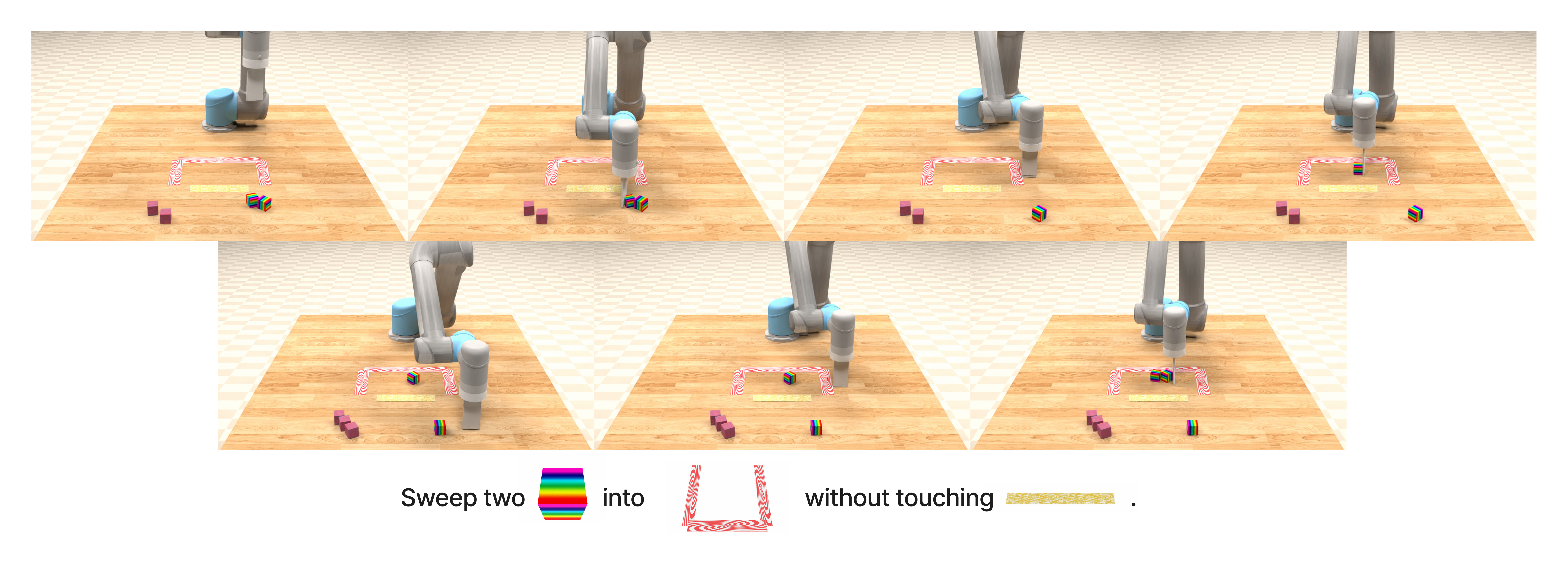}}
    \caption{Visual Constraint Satisfaction: Task 13}
    \label{supp:fig:traj:sweep_without_touching}
\end{figure}

\paragraph{\underline{Task 13:}} 
 Sweep the designated number of objects into a specified region without touching the constraint.
\begin{itemize}
    \item \textbf{Prompt:} \texttt{Sweep \{quantifier\} \{object\} into \{bounds\} without touching \{constraint\}.} 
    \item \textbf{Description:} Similar as task \textit{12} but requiring a different way to satisfy the constraint. The agent has to learn to avoid contacting the constraint line in this case. 
    \item \textbf{Success Criteria:} Similar as task \textit{12} except that the constraint is to not touch the red line.
    \item \textbf{Oracle Trajectory:} Shown in Fig.~\ref{supp:fig:traj:sweep_without_touching} with its multimodal prompt.  
\end{itemize}

\subsection{Visual Reasoning}

This task category requires agents to make decisions by reasoning over or memorizing information conveyed through multimodal prompts.

\begin{figure}[!htbp]
    \centering
    \makebox[\textwidth][c]{\includegraphics[width=1\textwidth]{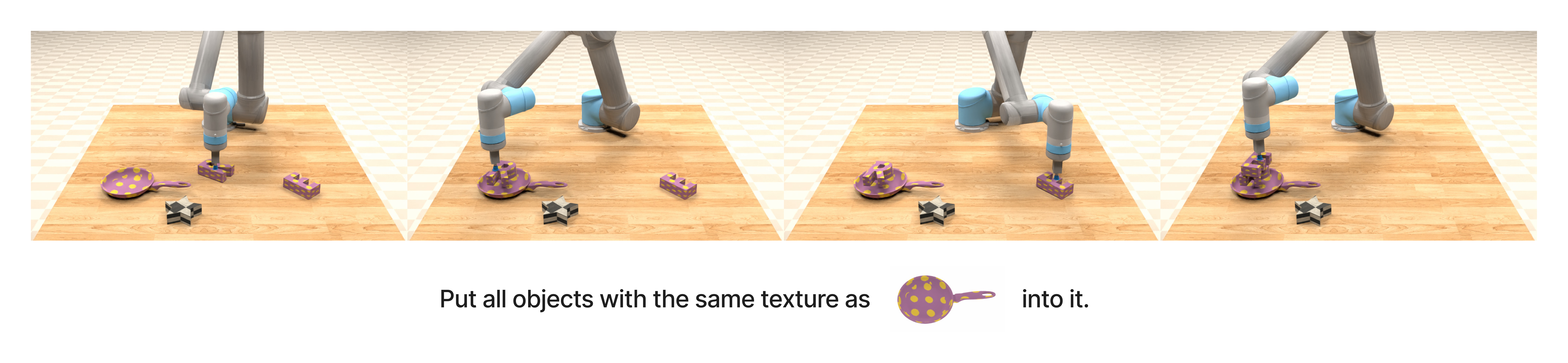}}
    \caption{Visual Reasoning: Task 14}
    \label{supp:fig:traj:same_color}
\end{figure}

\paragraph{\underline{Task 14:}} 
By reasoning the ``same texture'', the agent is required to pick all objects in the workspace with the same texture as the container objects specified in the prompt and place them into it. 
\begin{itemize}
    \item \textbf{Prompt:} \texttt{Put all objects with the same texture as \{object\} into it.}
    \item \textbf{Description:} \texttt{\{object\}} is the sampled goal container object. In the workspace, there are objects with the same texture as the container but potentially different shapes. Distractors with different textures are spawned.
    \item \textbf{Success Criteria:} All objects with the same texture as the goal container are within the bounds of the container. 
    \item \textbf{Oracle Trajectory:}  Shown in Fig.~\ref{supp:fig:traj:same_color} with its multimodal prompt. 
\end{itemize}

\begin{figure}[!htbp]
    \centering
    \makebox[\textwidth][c]{\includegraphics[width=1\textwidth]{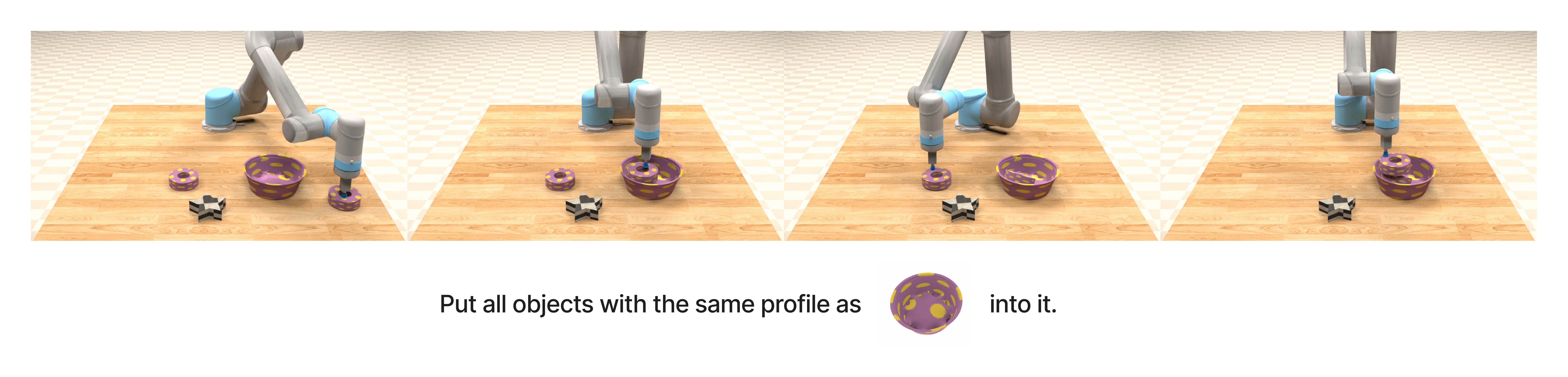}}
    \caption{Visual Reasoning: Task 15}
    \label{supp:fig:traj:same_profile}
\end{figure}

\paragraph{\underline{Task 15:}} 
By reasoning the ``same shape'', the agent is required to pick all objects in the workspace with the same top-down profile as the goal container specified in the prompt and place them into it. For example, blocks and boxes have the same rectangular profile. 
\begin{itemize}
    \item \textbf{Prompt:} \texttt{Put all objects with the same profile as \{object\} into it.}
    \item \textbf{Description:} Similar to the task \textit{14} except the objects to be picked and placed have the same shape. There are three different shapes: \textit{rectangular-like} (e.g. block and pallet), \textit{circle-like} (e.g. ring and bowl), and \textit{undetermined} for the rest.  
    \item \textbf{Success Criteria:} All objects with the same shape as the container are within the container.
    \item \textbf{Oracle Trajectory:} Shown in Fig.~\ref{supp:fig:traj:same_profile} with its multimodal prompt.
\end{itemize}

\begin{figure}[!htbp]
    \centering
    \makebox[\textwidth][c]{\includegraphics[width=1\textwidth]{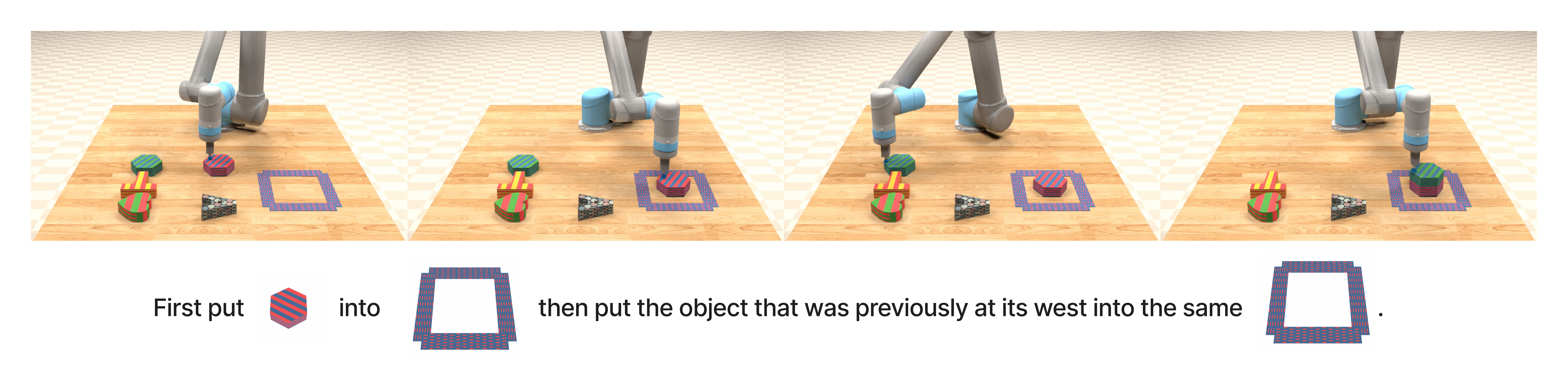}}
    \caption{Visual Reasoning: Task 16}
    \label{supp:fig:traj:manipulate_old_neighbor}
\end{figure}

\paragraph{\underline{Task 16:}} 
Put the target object into the container, and then put one of its old neighbors into the same container.
\begin{itemize}
    \item \textbf{Prompt:} \texttt{First put \{object\}}\textsubscript{1} \texttt{ into \{object\}}\textsubscript{2} \texttt{ then put the object } \texttt{that was previously at its \{direction\} into the same \{object\}}\textsubscript{2}.
    \item \textbf{Description:} Objects in image placeholders $\texttt{\{object\}}_1$ and $\texttt{\{object\}}_2$ are the target object to be picked and the container, respectively. We then ask the agent to put one of old neighbors of the previous target object into the same container. The old neighboring object is specified through cardinal directions \texttt{\{north, south, west, east\}}. 
    \item \textbf{Success Criteria:} The target object and the correct neighboring object are inside the container.
    \item \textbf{Oracle Trajectory:} Shown in Fig.~\ref{supp:fig:traj:manipulate_old_neighbor} with its multimodal prompt. 
\end{itemize}

\begin{figure}[!htbp]
    \centering
    \makebox[\textwidth][c]{\includegraphics[width=1\textwidth]{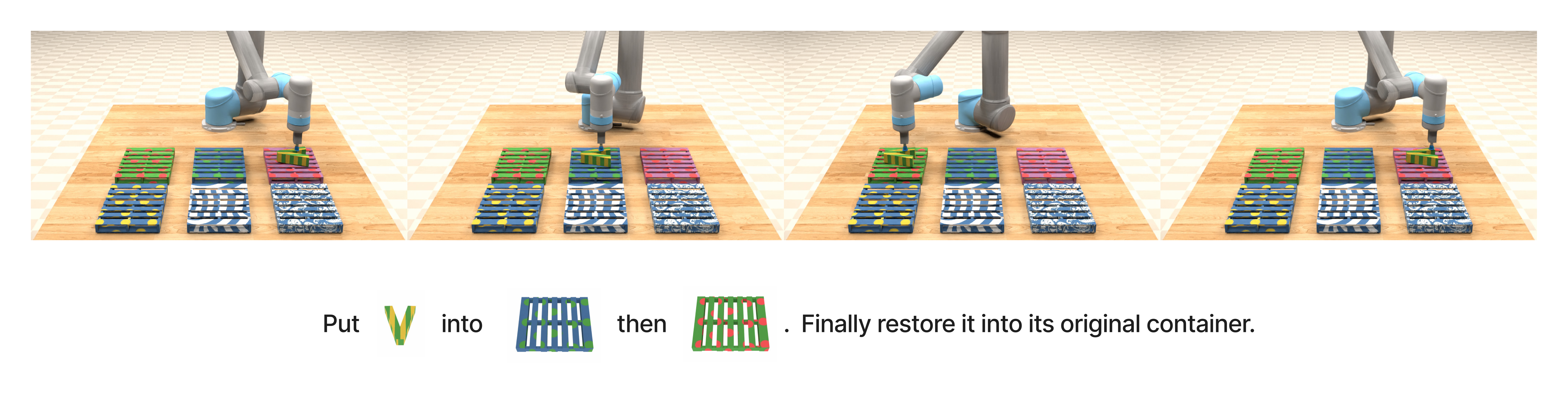}}
    \caption{Visual Reasoning: Task 17}
    \label{supp:fig:traj:pick_in_order_then_restore}
\end{figure}

\paragraph{\underline{Task 17:}} 
Pick and place the target object specified in the prompt into different containers in order then restore to the initial container. 
\begin{itemize}
    \item \textbf{Prompt:} \texttt{Put \{object\}}\textsubscript{1} \texttt{ into \{object\}}\textsubscript{2} \texttt{. Finally restore it into its original container.}
    \item \textbf{Description:} The object in the image placeholder \texttt{\{object\}}\textsubscript{1} is the target object to be manipulated across the task. There are more than one target containers (e.g. ``\texttt{Put \{object\}}\textsubscript{1} \texttt{ into \{object\}}\textsubscript{2} \texttt{ then \{object\}}\textsubscript{3}. \texttt{Finally restore it into its original container}'' for two target containers to be placed in order). The rest of spawned containers naturally becomes distractors.
    \item \textbf{Success Criteria:} The target object is first put into multiple containers following the specific order. Finally it should be restored into its original container.
    \item \textbf{Oracle Trajectory:} Shown in Fig.\ref{supp:fig:traj:pick_in_order_then_restore} with its multimodal prompt. 
\end{itemize}

\clearpage
\section{Model Architecture}
\label{supp:sec:model}

In this section, we provide comprehensive details about \vima model architecture as well as other adapted baseline methods. We implement all models in PyTorch~\citep{Paszke2019PyTorch} and adapt Transformer-related implementation from \citet{wolf2019huggingfaces}.

\subsection{Summary of Different Methods}
\label{supp:sec:method_compare}
We summarize differences between \vima and other baseline variants in Table~\ref{supp:table:method_compare}. In the column ``Prompt Conditioning'', an alternative to cross-attention is to first concatenate prompt and interaction into a big sequence, then repetitively apply transformer decoders to predict actions. It is referred to as ``Direct modeling''. The relative computation cost is quadratically proportional to the number of observation tokens.
\begin{table}[!htbp]
\caption{Comparison of different methods.}
\centering
\resizebox{1.0\textwidth}{!}{
\begin{tabular}{@{}llll@{}}
\toprule
                                                                     & \textbf{Visual Tokenizer}                                                                                                         & \textbf{Prompt  Conditioning} & \textbf{Number of Observation Tokens per Step}        \\ \midrule
Ours                                                                 & \begin{tabular}[c]{@{}l@{}}Object tokens consisting of \\ cropped images and bounding boxes\end{tabular}                 & Cross-attention      & Equal to number of objects, typically 3 to 8 \\
                                                                     &                                                                                                                          &                      &                                              \\
\begin{tabular}[c]{@{}l@{}}\vima-Gato\\\citep{reed2022gato}\end{tabular}                 & Image patch tokens encoded by a ViT                                                                                      & Direct modeling      & Equal to number of image patches, 16         \\
                                                                     &                                                                                                                          &                      &                                              \\
\begin{tabular}[c]{@{}l@{}}\vima-Flamingo\\\citep{alayrac2022flamingo}\end{tabular}       & \begin{tabular}[c]{@{}l@{}}Image patch tokens encoded by a ViT,\\ further downsampled by a Perceiver module\end{tabular} & Cross-attention      & Equal to number of learned query vectors, 4  \\
                                                                     &                                                                                                                          &                      &                                              \\
\begin{tabular}[c]{@{}l@{}}\vima-GPT\\\citep{brown2020gpt3}\end{tabular} & Single image token encoded by a ViT                                                                                      & Direct modeling      & Single visual feature, 1                     \\ \bottomrule
\end{tabular}
}
\label{supp:table:method_compare}
\end{table}

\subsection{\vima Architecture}
\label{supp:sec:model_vima}
\subsubsection{Multimodal Prompt Tokenization}
\label{supp:sec:model_vima_tokenization}

As introduced in Section~\ref{sec:method}, there are 3 types of input formats in multimodal prompts, namely (1) \textbf{text inputs}, (2) \textbf{images of full scenes}, and (3) \textbf{images of single objects}.

For \textbf{text inputs}, we follow the standard pipeline in NLP to first tokenize raw language to discrete indices through pre-trained \texttt{t5-base} tokenizer. We then obtain corresponding word tokens from the embedding look-up of the pre-trained \texttt{t5-base} model. For \textbf{images of full scenes}, we first parse the scene through a fine-tuned Mask R-CNN detection model~\citep{he2017mask,wu2019detectron2} to extract individual objects. Each object representation contains a bounding box and a cropped image. The bounding box is in the format of $\begin{bmatrix}x_{\text{center}}, y_{\text{center}}, \text{height}, \text{width} \end{bmatrix}$. We normalize it to be within $[0, 1]$ by dividing each dimension with corresponding upper-bound value. We then pass it through a bounding box encoder MLP and obtain a feature vector. To process the cropped image, we first pad non-square image to a square by padding along the shorter dimension. We then resize it to a pre-configured size and pass it through a ViT (trained from scratch) to obtain the image feature. Finally, an object token is obtained by concatenating the bounding box feature and the image feature and mapping to the embedding dimension. For \textbf{images of single objects}, we obtain tokens in the same way except with a dummy bounding box. Detailed model hyperparameters about tokenization are listed in Table~\ref{supp:table:model_hyperparam_prompt}. 

\begin{table}[ht]
\centering
\caption{Model hyperparameters for multimodal prompt tokenization.}
\begin{tabular}{@{}ll@{}}
\toprule
\textbf{Hyperparameter}          & \textbf{Value}             \\ \midrule
\multicolumn{2}{c}{Text Tokenization}       \\ \midrule
Tokenizer               & \texttt{t5-base} tokenizer \\
Embedding Dimension     & 768               \\ \midrule
\multicolumn{2}{c}{Image Tokenization}      \\ \midrule
ViT Input Image Size    & 32 $\times$ 32           \\
ViT Patch Size          & 16                \\
ViT Width               & 768               \\
ViT Layer               & 4                 \\
ViT Number of Heads     & 24                \\ \midrule
\multicolumn{2}{c}{Bounding Box MLP}        \\ \midrule
Hidden Dimension        & 768               \\
Hidden Depth            & 2                 \\ \midrule
\multicolumn{2}{c}{Prompt Encoding}         \\ \midrule
Pre-Trained LM          & \texttt{t5-base}           \\
Unfrozen Last $N$ Layers  & 2                 \\
Positional Embedding    & Absolute          \\
Token Adapter MLP Depth & 2                 \\ \bottomrule
\end{tabular}
\label{supp:table:model_hyperparam_prompt}
\end{table}

After obtaining a sequence of prompt tokens, we follow \citet{deepmind2021frozen} to pass it through a pre-trained \texttt{t5-base} encoder to obtain encoded prompt. Note that we add adapter MLP between object tokens and the T5 encoder.
To prevent catastrophic forgetting, \vima only fine-tunes the last two layers of the language encoder with layer-wise learning rate decay~\citep{he2021mae} but freezes all other layers.
We adopt learned absolute positional embedding. Model hyperparameters are listed in Table~\ref{supp:table:model_hyperparam_prompt} as well.

\subsubsection{Observation Encoding}
Since all RGB observations are images of full scenes, we follow the same procedure discussed above to obtain flattened object tokens. Because we provide RGBs from two views (frontal and top-down), we order object tokens by following the order of $\begin{bmatrix} \text{frontal, top-down} \end{bmatrix}$. We one-hot encode the state of the end effector. We then concatenate object tokens with the end-effector state and transform to observation tokens. We adopt learned absolute positional embedding. Detailed model hyperparameters about observation encoding is provided in Table~\ref{supp:table:obs_encd_hyperparams}.

\begin{table}[ht]
\caption{Model hyperparameters for observation encoding.}
\centering
\begin{tabular}{@{}ll@{}}
\toprule
\textbf{Hyperparameter}                   & \textbf{Value}    \\ \midrule
Observation Token Dimension      & 768      \\
End Effector Embedding Dimension & 2        \\
Positional Embedding             & Absolute \\ \bottomrule
\end{tabular}
\label{supp:table:obs_encd_hyperparams}
\end{table}

\subsubsection{Action Encoding}
Since our model is conditioned on observation-action interleaved history, we also tokenize past actions. We follow common practice in \citet{chen2021decisiontransformer,zheng2022onlinedt} to encode past actions with a two-layer MLP. It has a hidden dimension of 256. We then map outputs to token dimension and obtain action tokens.

\subsubsection{Sequence Modeling}
\label{supp:sec:model_vima_controller}
The robot controller in \vima is a causal decoder that autoregressively predicts actions. To condition the decoder on prompt tokens, we perform cross-attention between history tokens and prompt tokens (Figure~\ref{fig:arch}). Concretely, we pass history tokens as the query sequence and prompt tokens as the key-value sequence into cross-attention blocks. The output prompt-aware trajectory tokens then go through causal self-attention blocks. We alternate cross-attention and self-attention $L$ times. This procedure is technically described in Pseudocode~\ref{supp:code:xattn}.

\begin{minipage}{\linewidth}
\begin{lstlisting}[language=Python,caption={Cross-attention operation that conditions the trajectory history on prompt. We repetitively alternate cross-attention and self-attention to model the trajectory given a specific task.},label={supp:code:xattn}]
def xattn_sequence_modeling(
    prompt_tokens,      # the [L, d] prompt tokens (L=prompt length)
    obs_tokens,         # the [T, d] obs tokens (T=time step)
    act_tokens,         # the [T-1, d] action tokens
    traj_pos_embd,      # learned positional embedding for trajectory
    prompt_pos_embd,    # learned positional embedding for prompt
):
    # interleave obs and action tokens
    traj_tokens = interleave(obs_tokens, act_tokens)  # [2T-1, d]
    # add positional embedding to trajectory tokens
    x = traj_tokens + traj_pos_embd
    # add positional embedding to prompt tokens
    prompt_tokens = prompt_tokens + prompt_pos_embd

    # apply xattn and causal self-attn
    for i in range(num_layers):
        # cross-attention
        x = x + attn_i(q=x, kv=prompt_tokens)
        # feed forward
        x = x + ffw_xattn_i(x)
        # self-attention
        x = x + causal_attn_i(q=x, kv=x)
        # feed forward
        x = x + ffw_i(x)
        
    # the last token is the predicted action token
    predicted_act_token = x[-1]
    return predicted_act_token
\end{lstlisting}
\end{minipage}

\subsubsection{Action Decoding}
After obtaining the predicted action token, we map it to the action space $\mathcal{A}$ and obtain the predicted action. This is achieved though a group of action heads. Since the action space consists of two $\mathbf{SE}(2)$ poses, for each pose we use six independent heads to decode discrete actions (two for xy coordinate and four for rotation represented in quaternion). These discrete actions are then integrated and mapped to continuous actions through affine transformation. The two poses are modeled independently. Early ablations show that this independent modeling is equally good as alternative techniques, such as autoregressive decoding~\citep{vinyals2019alphastar,openai2019dota}.
Detailed model hyperparameters are listed in Table~\ref{supp:table:action_decoder}.
\begin{table}[ht]
\caption{Model hyperparameters for action decoders.}
\centering
\begin{tabular}{@{}ll@{}}
\toprule
\textbf{Hyperparameter}         & \textbf{Value} \\ \midrule
Hidden Dimension       & 512   \\
Hidden Depth           & 2     \\
Activation             & ReLU  \\
X-Axis Discrete Bins   & 50    \\
Y-Axis Discrete Bins   & 100   \\
Rotation Discrete Bins & 50    \\ \bottomrule
\end{tabular}
\label{supp:table:action_decoder}
\end{table}

\subsection{Baselines Architectures}
In this section, we elaborate model architectures for adapted baseline methods. Some components such as the action decoder are same across all models. Therefore, we only discuss unique model components. 

\subsubsection{\vima-Gato}
\label{supp:sec:model_gato}

\textbf{Gato}~\citep{reed2022gato} introduces a decoder-only model that solves tasks from multiple domains including robotics, video game, image captioning, language modeling, etc. Different tasks are specified by supplying the model with an initial sequence of corresponding tokens. For example, in tasks involving decision making, these tokens include observation and action tokens. For fair comparison, we provide the same conditioning as \vima, i.e., our multimodal tokenized prompts.
This adapted baseline variant is referred to as ``\textbf{\vima-Gato}''.
Similar to our method, \vima-Gato also predicts actions in an autoregressive manner. \vima-Gato and our method share the same training philosophy to only optimize the causal behavior cloning objective. However, unlike our method that adopts an object-centric representation to treat individual objects as observation tokens, \vima-Gato divides input images into patches and encodes them by a ViT~\citep{dosovitskiy2020image} to produce observation tokens. Furthermore, \vima-Gato relies on causal self-attention to model entire trajectory sequences starting with prompt tokens. Hyperparameters of \vima-Gato's ViT is listed in Table~\ref{supp:table:baseline_vit}. The transformer-decoder style sequence modeling is technically illustrated in Pseudocode~\ref{supp:code:selfattn}.

\begin{table}[h]
\caption{Model hyperparameters for ViT used in baseline methods.}
\centering
\begin{tabular}{@{}ll@{}}
\toprule
\textbf{Hyperparameter} & \textbf{Value}    \\ \midrule
Image Size     & 64 $\times$ 128 \\
Patch Size     & 32       \\
ViT Width      & 768      \\
ViT Layers     & 4        \\
ViT Heads      & 24       \\ \bottomrule
\end{tabular}
\label{supp:table:baseline_vit}
\end{table}

\begin{minipage}{\linewidth}
\begin{lstlisting}[language=Python,label={supp:code:selfattn},caption={Plain sequence modeling that directly concatenates prompt and trajectory history and repetitively perform causal self-attention operation.}]
def causal_sequence_modeling(
    prompt_tokens,  # the [L, d] prompt tokens (L=prompt length)
    sep_token,      # the [1, d] learned token to separate prompt and trajectory history
    obs_tokens,     # the [T, d] obs tokens (T=time step)
    act_tokens,     # the [T-1, d] action tokens
    pos_embd,       # learned positional embedding
):
    # interleave obs and action tokens
    traj_tokens = interleave(obs_tokens, act_tokens)  # [2T-1, d]
    # assemble input tokens
    x = concat([prompt_tokens, sep_token, traj_tokens])
    x = x + pos_embd
    
    # apply GPT layers with causal mask
    for i in range(num_layers):
        # self-attention
        x = x + causal_attn_i(q=x, kv=x)
        # feed forward
        x = x + ffw_i(x)
        
    # the last token is the predicted action token
    predicted_act_token = x[-1]
    return predicted_act_token
\end{lstlisting}
\end{minipage}

\subsubsection{\mbox{\vima}-Flamingo}
\label{supp:sec:model_flamingo}

\textbf{Flamingo}~\citep{alayrac2022flamingo} is a vision-language model that learns to generate textual completion in response to multimodal prompts. It embeds a variable number of prompt images into a fixed number of tokens via the Perceiver Resampler module~\citep{jaegle2021perceiver}, and conditions the language decoder on encoded prompts by cross-attention. Flamingo does not work with embodied agents out of the box. We adapt it by replacing the output layer with robot action heads (hyperparameters listed in Table~\ref{supp:table:action_decoder}) and using tokenized rollout histories as inputs.
We thus call it ``\textbf{\vima-Flamingo}''.
We train it end-to-end with causal behavior cloning loss. \vima-Flamingo differs from ours since it processes image observations into a fixed number of visual tokens through a learned Perceiver Resampler. Model hyperparameters for our reimplementation of the Perceiver Resampler is listed in Table~\ref{supp:table:perceiver}.
\begin{table}[htbp]
\caption{Model hyperparameters for Perceiver Resampler used in \vima-Flamingo method.}
\centering
\begin{tabular}{@{}ll@{}}
\toprule
\textbf{Hyperparameter}           & \textbf{Value} \\ \midrule
Number of Latent Queries & 4     \\
Number of Blocks         & 4     \\
Self-Attn per Block      & 4     \\
Self-Attn Heads          & 24    \\
Cross-Attn Heads         & 24    \\ \bottomrule
\end{tabular}
\label{supp:table:perceiver}
\end{table}

\subsubsection{\vima-GPT}
\label{supp:sec:model_dt}
\textbf{\vima-GPT} is a GPT-based behavior cloning agent conditioned on tokenized multimodal prompts with the GPT architecture. It autoregressively decodes next actions given multimodal prompts and interaction histories. We optimize this method end-to-end with causal behavior cloning loss. Similar to prior works of casting RL problems as sequence modeling~\citep{chen2021decisiontransformer,janner2021onebigsequence,zheng2022onlinedt}, it encodes an image into a single ``state'' token through a learned ViT encoder. It also directly models entire trajectory sequences prepended with prompt tokens. Therefore, it differs from our method in the representation of observation tokens and prompt conditioning. For visual tokenizer, we employ a learned ViT with hyperparameters listed in Table~\ref{supp:table:baseline_vit}.

\subsection{Mask R-CNN Detection Model}
\label{supp:sec:mask_rcnn}
Finally, we elaborate on the mask R-CNN model \citep{he2017mask} for scene parsing and object extraction. We fine-tune a pre-trained lightweight mask R-CNN (\texttt{mask\textunderscore{}rcnn\textunderscore{}R\textunderscore{}50\textunderscore{}FPN\textunderscore{}3x}) from \citet{wu2019detectron2} to adapt to scenes and images in our tabletop environment.
We fine-tune it on a subset of agent training dataset. It contains 100 trajectories for each task, resulting in 22,741 images and 61,822 annotations in total. We use learning rate $5\times 10^{-4}$ and train for 10 epochs. During model selection, we particularly favor models with high recall to reduce the number of missed objects. To compensate for resulting false-positives, we adopt object augmentation during agent training (Appendix, Sec.~\ref{supp:sec:training}).

\begin{figure}[!htbp]
    \centering
    \makebox[\textwidth][c]{\includegraphics[width=\textwidth]{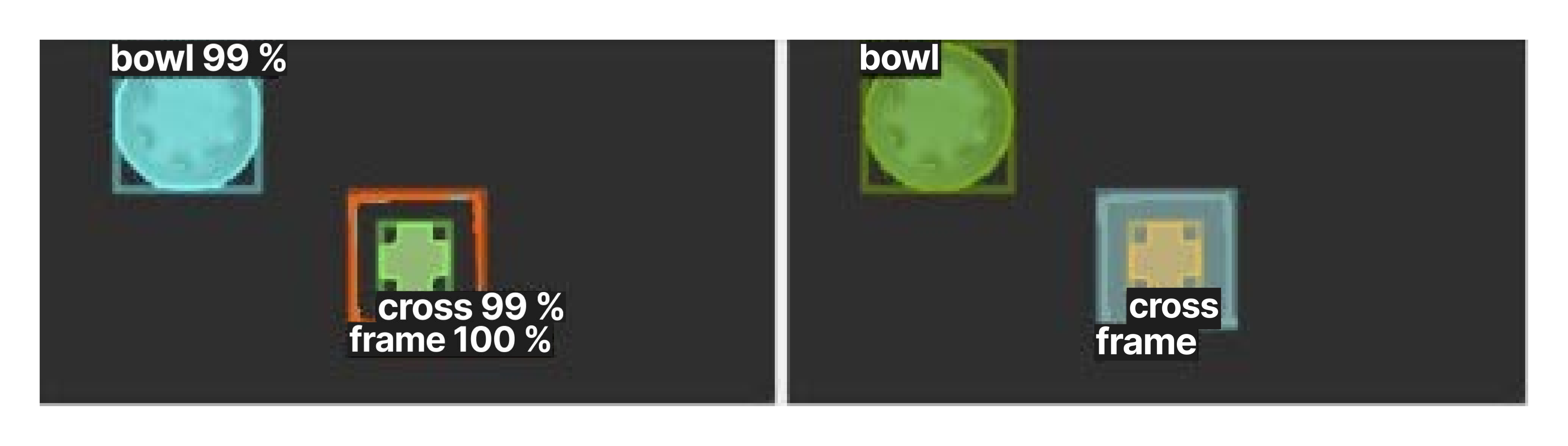}}
    \caption{Visualization of fine-tuned mask R-CNN. \textit{Left}: Prediction from the detection model. \textit{Right}: Ground-truth scene parsing. The detection model agrees well with ground-truth objects.}
    \label{supp:fig:mask_rcnn}
\end{figure}

A visualization of its output is provided in Figure~\ref{supp:fig:mask_rcnn}. We do not use the predicted object names in our models.

\clearpage
\section{\vima Training Details}
\label{supp:sec:training}

We follow the best practice to train Transformer models using the AdamW optimizer~\citep{loshchilov2019decoupled}, learning rate warm-up, cosine annealing~\citep{loshchilov17cosinelr}, etc. Training hyperparameters are provided in Table~\ref{supp:table:train_hyperparams}. We use GEGLU activation~\citep{shazeer2020glu} inside Transformer models across all methods.

\begin{table}[!htbp]
\caption{Hyperparameters used during training.}
\centering
\begin{tabular}{@{}ll@{}}
\toprule
\textbf{Hyperparameter}            & \textbf{Value}  \\ \midrule
Learning Rate             & 0.0001 \\
Warmup Steps              & 7K     \\
LR Cosine Annealing Steps & 17K    \\
Weight Decay              & 0      \\
Dropout                   & 0.1    \\
Gradient Clip Threshold   & 1.0    \\ \bottomrule
\end{tabular}
\label{supp:table:train_hyperparams}
\end{table}

To make trained models robust to detection inaccuracies and failures, we apply \emph{object augmentation} by randomly injecting \emph{false-positive} detection outputs. Concretely, for observation at each time step, we sample number of augmented objects i.i.d. $n_\text{augmented objects} \sim \text{Cat}(K, \mathbf{p})$, where $\text{Cat}(\cdot)$ denotes a categorical distribution with $K$ supports parameterized by $\mathbf{p}$. For each augmented object, we then randomly sample a bounding box and corresponding cropped image to add to object tokens. In our experiments, we set $\mathbf{p} = \{0: 0.95, 1: 0.05\}$ with $K=2$.

\subsection{Vary Model Capacity}
We train a spectrum of 7 models ranging from 2M to 200M parameters. To vary the model capacity, we follow prior work~\citep{chowdhery2022palm} to change embedding dimension and number of layers. We list configurations for methods with cross-attention prompt conditioning (i.e., ours and \vima-Flamingo) in Table~\ref{supp:table:vary_size_xattn}, and configurations for methods only with causal self-attention (i.e., \vima-Gato and \vima-GPT) in Table~\ref{supp:table:vary_size_selfattn}.

\begin{table}[!htbp]
\caption{Configurations for differently sized models with cross-attention prompt conditioning.}
\centering
\begin{tabular}{@{}ccccc@{}}
\toprule
\textbf{Model Size (M)} & \textbf{Embedding Dimension} & \textbf{Num Blocks} & \textbf{X-Attn Heads} & \textbf{Self-Attn Heads} \\ \midrule
2              & 256                 & 1          & 8            & 8               \\
4              & 256                 & 2          & 8            & 8               \\
9              & 320                 & 3          & 10           & 10              \\
20             & 384                 & 4          & 12           & 12              \\
43             & 512                 & 5          & 16           & 16              \\
92             & 640                 & 7          & 20           & 20              \\
200            & 768                 & 11         & 24           & 24              \\ \bottomrule
\end{tabular}
\label{supp:table:vary_size_xattn}
\end{table}

\begin{table}[!htbp]
\caption{Configurations for differently sized models with causal self-attention prompt conditioning.}
\centering
\begin{tabular}{@{}cccc@{}}
\toprule
\textbf{Model Size (M)} & \textbf{Embedding Dimension} & \textbf{Num Blocks} & \textbf{Self-Attn Heads} \\ \midrule
2              & 64                  & 1          & 2               \\
4              & 96                  & 2          & 3               \\
9              & 192                 & 3          & 6               \\
20             & 320                 & 4          & 10              \\
43             & 512                 & 5          & 16              \\
92             & 768                 & 7          & 24              \\
200            & 768                 & 18         & 24              \\ \bottomrule
\end{tabular}
\label{supp:table:vary_size_selfattn}
\end{table}

\clearpage
\section{Extended Experiment Results}
\label{supp:sec:results}

\subsection{Training Time and Compute}
All experiments are conducted on cluster nodes, each with 8 NVIDIA V100 GPUs. The largest experiment takes approximately one day. We utilize DDP (distributed data parallel) to accelerate the training.

\subsection{Model Scaling}
\subsubsection{Numerical Results}
We present numerical results that constitute Fig.~\ref{fig:scalability} in Table~\ref{supp:table:model_scalability_numerical_results}. The claim of ``up to $2.9 \times$ improvement'' made in Abstract and Sec.~\ref{sec:introduction} is calculated as follows. The best competing variant is \vima-Gato. On the hardest L4, our method shows the most significant relative improvement with a model size of 20M. We compute the performance gap, divide by \vima-Gato's performance, and only keep the first digit after decimal to obtain the result.

\begin{table}[h]
\caption{Model scaling numerical results that constitute Fig.~\ref{fig:scalability}. Numbers in the first row indicate robot controller parameter count.}
\label{supp:table:model_scalability_numerical_results}
\centering
\begin{tabular}{c|r|ccccccc}
\toprule 
\textbf{Level} & \textbf{Method} & \textbf{2M} & \textbf{4M} & \textbf{9M} & \textbf{20M} & \textbf{43M} & \textbf{92M} & \textbf{200M} \\ \midrule 

\multirow{4}{*}{L1} & Ours & $\bestscore{\hphantom{0}76.5}$ & $\bestscore{\hphantom{0}79.2}$ & $\bestscore{\hphantom{0}77.4}$ & $\bestscore{\hphantom{0}77.1}$ & $\bestscore{\hphantom{0}78.2}$ & $\bestscore{\hphantom{0}79.3}$ & $\bestscore{\hphantom{0}81.5}$ \\
  & \vima-Gato & \hphantom{0}37.6 & \hphantom{0}42.6 & \hphantom{0}44.2 & \hphantom{0}46.1 & \hphantom{0}49.5 & \hphantom{0}57.0 & \hphantom{0}58.0 \\
  & \vima-Flamingo & \hphantom{0}42.4 & \hphantom{0}48.9 & \hphantom{0}45.6 & \hphantom{0}46.6 & \hphantom{0}47.0 & \hphantom{0}47.2 & \hphantom{0}47.4 \\
  & \vima-GPT & \hphantom{0}30.0 & \hphantom{0}37.0 & \hphantom{0}44.9 & \hphantom{0}48.5 & \hphantom{0}48.0 & \hphantom{0}47.9 & \hphantom{0}46.9 \\
\midrule 

\multirow{4}{*}{L2} & Ours & $\bestscore{\hphantom{0}77.1}$ & $\bestscore{\hphantom{0}79.2}$ & $\bestscore{\hphantom{0}78.2}$ & $\bestscore{\hphantom{0}77.6}$ & $\bestscore{\hphantom{0}77.6}$ & $\bestscore{\hphantom{0}80.1}$ & $\bestscore{\hphantom{0}81.5}$ \\
  & \vima-Gato & \hphantom{0}35.9 & \hphantom{0}39.3 & \hphantom{0}41.3 & \hphantom{0}44.1 & \hphantom{0}46.6 & \hphantom{0}53.9 & \hphantom{0}53.1 \\
  & \vima-Flamingo & \hphantom{0}41.0 & \hphantom{0}46.5 & \hphantom{0}44.6 & \hphantom{0}44.6 & \hphantom{0}45.4 & \hphantom{0}47.1 & \hphantom{0}46.0 \\
  & \vima-GPT & \hphantom{0}29.8 & \hphantom{0}35.0 & \hphantom{0}43.3 & \hphantom{0}45.8 & \hphantom{0}45.9 & \hphantom{0}47.4 & \hphantom{0}46.9 \\
\midrule 

\multirow{4}{*}{L3} & Ours & $\bestscore{\hphantom{0}77.3}$ & $\bestscore{\hphantom{0}77.8}$ & $\bestscore{\hphantom{0}78.5}$ & $\bestscore{\hphantom{0}77.3}$ & $\bestscore{\hphantom{0}81.8}$ & $\bestscore{\hphantom{0}81.9}$ & $\bestscore{\hphantom{0}78.7}$ \\
  & \vima-Gato & \hphantom{0}29.0 & \hphantom{0}33.2 & \hphantom{0}37.5 & \hphantom{0}40.2 & \hphantom{0}42.5 & \hphantom{0}45.6 & \hphantom{0}46.0 \\
  & \vima-Flamingo & \hphantom{0}35.0 & \hphantom{0}41.9 & \hphantom{0}39.2 & \hphantom{0}40.5 & \hphantom{0}40.3 & \hphantom{0}42.1 & \hphantom{0}40.7 \\
  & \vima-GPT & \hphantom{0}25.3 & \hphantom{0}29.3 & \hphantom{0}39.0 & \hphantom{0}43.5 & \hphantom{0}43.0 & \hphantom{0}42.6 & \hphantom{0}42.2 \\
\midrule 

\multirow{4}{*}{L4} & Ours & $\bestscore{\hphantom{0}25.7}$ & $\bestscore{\hphantom{0}49.0}$ & $\bestscore{\hphantom{0}47.1}$ & $\bestscore{\hphantom{0}48.8}$ & $\bestscore{\hphantom{0}49.0}$ & $\bestscore{\hphantom{0}49.6}$ & $\bestscore{\hphantom{0}48.6}$ \\
  & \vima-Gato & \hphantom{0}13.3 & \hphantom{0}13.2 & \hphantom{0}12.2 & \hphantom{0}12.3 & \hphantom{0}12.8 & \hphantom{0}13.5 & \hphantom{0}16.8 \\
  & \vima-Flamingo & \hphantom{0}12.3 & \hphantom{0}11.6 & \hphantom{0}10.7 & \hphantom{0}12.1 & \hphantom{0}10.7 & \hphantom{0}11.1 & \hphantom{0}12.1 \\
  & \vima-GPT & \hphantom{0}11.1 & \hphantom{0}10.3 & \hphantom{0}12.7 & \hphantom{0}14.2 & \hphantom{0}11.8 & \hphantom{0}12.1 & \hphantom{0}12.1 \\

\bottomrule

\end{tabular}
\end{table}

\subsection{Data Scaling}
\label{supp:sec:data_scalability}
\subsubsection{Detailed Setup}
To ensure all methods are fairly pre-trained on the same amount of data (i.e., they have roughly the same amount of built-in information, thus the x-axis in Fig.~\ref{fig:scalability} faithfully corresponds to the extra bits of information seen during further training), we initialize variants that directly learn from raw pixels with MVP pre-trained ViT~\citep{xiao2022masked,Radosavovic2022}. It is further MAE fine-tuned~\citep{he2021mae}, using the \emph{same} in-domain data as for the Mask R-CNN object detector. Note that the MVP pre-trained then domain fine-tuned ViT also updates weights jointly with robot controllers later on. We use the \texttt{ViT-B} backbone from MVP. The in-domain data for fine-tuning include 100 trajectories for each task.

\subsubsection{Numerical Results}
We present numerical results that constitute Fig.~\ref{fig:scalability} in Table~\ref{supp:table:data_scalability_numerical_results}. The claim of ``$2.7 \times$ improvement'' made in Abstract and Sec.~\ref{sec:introduction} is calculated as follows. The best competing variant is \vima-Gato that achieves $12.2\%$ average success rate trained with full data on L4. Our method trained with $10\%$ data achieves $46\%$ average success rate on the same level. We compute the performance gap, divide by \vima-Gato's performance, and only keep the first digit after decimal to obtain the result.

\begin{table}
\caption{Data scaling numerical results that constitute Fig.~\ref{fig:scalability}. Numbers in the first row indicate the size of training dataset.}
\label{supp:table:data_scalability_numerical_results}
\centering
\begin{tabular}{c|r|cccc}
\toprule 

\textbf{Level} & \textbf{Method} & \textbf{\hphantom{000}0.1\%} & \textbf{\hphantom{00}1\%} & \textbf{\hphantom{00}10\%} & \textbf{Full (100\%)} \\ \midrule 

\multirow{4}{*}{L1} & Ours & $\bestscore{\hphantom{00}0.0}$ & $\bestscore{\hphantom{0}36.3}$ & $\bestscore{\hphantom{0}76.3}$ & $\bestscore{\hphantom{0}79.3}$ \\
  & \vima-Gato & $\bestscore{\hphantom{00}0.0}$ & \hphantom{0}11.5 & \hphantom{0}41.5 & \hphantom{0}57.5 \\
  & \vima-Flamingo & $\bestscore{\hphantom{00}0.0}$ & \hphantom{00}2.0 & \hphantom{0}37.7 & \hphantom{0}52.3 \\
  & \vima-GPT & $\bestscore{\hphantom{00}0.0}$ & \hphantom{00}6.0 & \hphantom{0}30.9 & \hphantom{0}52.8 \\
\midrule 

\multirow{4}{*}{L2} & Ours & $\bestscore{\hphantom{00}0.0}$ & $\bestscore{\hphantom{0}34.3}$ & $\bestscore{\hphantom{0}75.8}$ & $\bestscore{\hphantom{0}80.1}$ \\
  & \vima-Gato & $\bestscore{\hphantom{00}0.0}$ & \hphantom{0}10.1 & \hphantom{0}37.9 & \hphantom{0}41.2 \\
  & \vima-Flamingo & $\bestscore{\hphantom{00}0.0}$ & \hphantom{00}2.0 & \hphantom{0}33.8 & \hphantom{0}32.6 \\
  & \vima-GPT & $\bestscore{\hphantom{00}0.0}$ & \hphantom{00}6.0 & \hphantom{0}29.7 & \hphantom{0}40.3 \\
\midrule 

\multirow{4}{*}{L3} & Ours & $\bestscore{\hphantom{00}0.0}$ & $\bestscore{\hphantom{0}15.4}$ & $\bestscore{\hphantom{0}73.2}$ & $\bestscore{\hphantom{0}81.9}$ \\
  & \vima-Gato & $\bestscore{\hphantom{00}0.0}$ & \hphantom{0}10.2 & \hphantom{0}34.8 & \hphantom{0}40.9 \\
  & \vima-Flamingo & $\bestscore{\hphantom{00}0.0}$ & \hphantom{00}1.0 & \hphantom{0}33.1 & \hphantom{0}33.6 \\
  & \vima-GPT & $\bestscore{\hphantom{00}0.0}$ & \hphantom{00}5.5 & \hphantom{0}28.6 & \hphantom{0}39.2 \\
\midrule 

\multirow{4}{*}{L4} & Ours & $\bestscore{\hphantom{00}0.0}$ & $\bestscore{\hphantom{0}17.0}$ & $\bestscore{\hphantom{0}46.0}$ & $\bestscore{\hphantom{0}49.6}$ \\
  & \vima-Gato & $\bestscore{\hphantom{00}0.0}$ & \hphantom{00}2.7 & \hphantom{0}10.8 & \hphantom{0}12.2 \\
  & \vima-Flamingo & $\bestscore{\hphantom{00}0.0}$ & \hphantom{00}0.5 & \hphantom{0}11.2 & \hphantom{0}12.0 \\
  & \vima-GPT & $\bestscore{\hphantom{00}0.0}$ & \hphantom{00}1.1 & \hphantom{00}7.1 & \hphantom{0}14.3 \\

\bottomrule

\end{tabular}
\end{table}

\subsubsection{What If Baseline Variants' ViT Is Trained from Scratch?}
\begin{figure}[h]
    \vspace{-0.1in}
    \centering
    \makebox[\textwidth][c]{\includegraphics[width=1\textwidth]{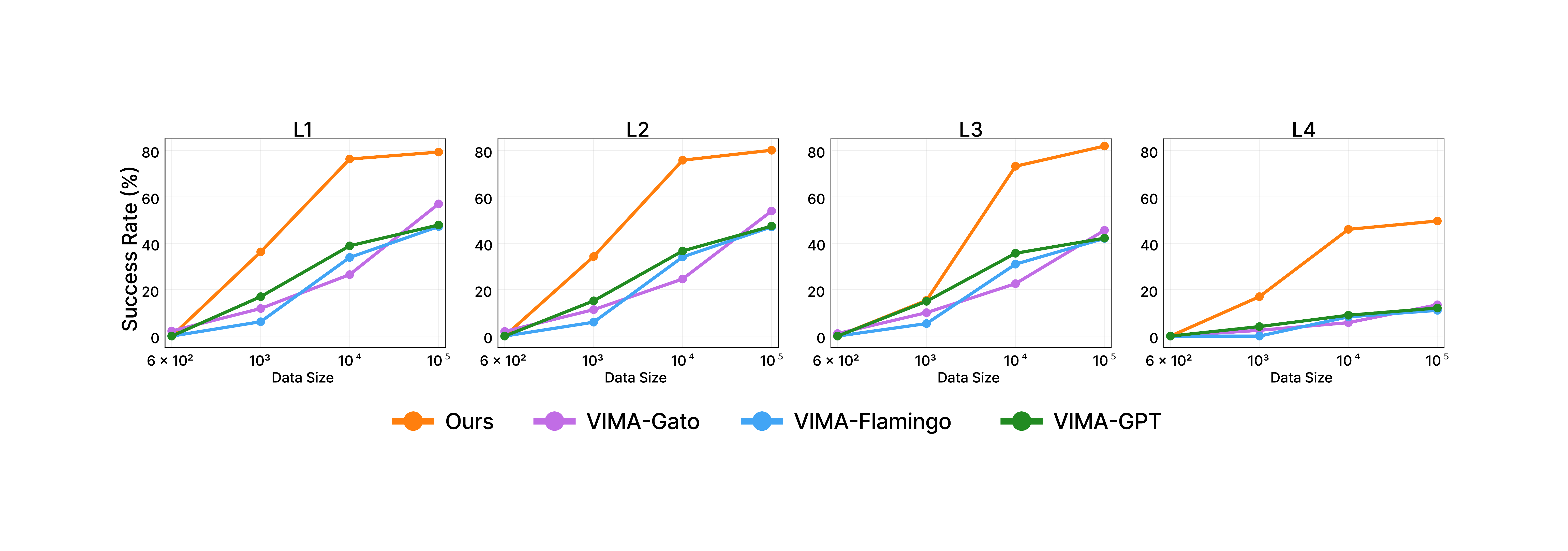}}
    \caption{Data scaling when baseline variants' ViT is trained from scratch. In settings with reasonable amounts of in-domain data (i.e., the 1\%, 10\%, and 100\% scenarios), our recommended recipe always outperforms other variants.}
    \label{supp:fig:data_scalability_baseline_vit_from_scratch}
    \vspace{-0.1in}
\end{figure}

We further investigate what if baseline variants' ViT is trained from scratch and end-to-end with the robot controllers. We visualize the results in Fig.~\ref{supp:fig:data_scalability_baseline_vit_from_scratch} and numerically present them in Table~\ref{supp:table:data_scalability_compared_to_vit_from_scratch}. We annotate with arrows to indicate performance increase ($\uparrow$) and decrease ($\downarrow$). We highlight two findings.

First, MVP pre-trained ViT is most beneficial in the setting with sufficient in-domain training data (i.e., the 10\% data scenario). It boosts the performance for the most competing baseline variant \vima-Gato. However, in other settings with abundant in-domain data (i.e., the full data scenario) or insufficient in-domain data (i.e., 1\% and 0.1\% scenarios), the advantage of MVP pre-trained ViT diminishes and it even becomes detrimental. This aligns with the finding in previous empirical studies~\citep{hansen2022pretraining}. Second, in settings with reasonable amounts of in-domain data (i.e., the 1\%, 10\%, and 100\% scenarios), our recommended recipe always outperforms other variants. We notice that such a data demand generally can be satisfied by both simulated robotics data~\citep{mandlekar2021matters} and real robotics data~\citep{dasari19robonet,brohan2022rt1}. Therefore, it demonstrates that our recommended recipe is highly sample-efficient compared to alternative designs, especially under practical settings.

\begin{table}[t]
\caption{Data scaling when baseline variants' ViT is trained from scratch, indicated inside parentheses. $\uparrow$ and $\downarrow$ denote performance increase and decrease. Numbers in the first row represent the size of training dataset.}
\label{supp:table:data_scalability_compared_to_vit_from_scratch}
\centering
\begin{tabular}{c|r|cccc}
\toprule
\textbf{Level} & \textbf{Method} & \textbf{\hphantom{000}0.1\%} & \textbf{\hphantom{00}1\%} & \textbf{\hphantom{00}10\%} & \textbf{Full (100\%)} \\ \midrule 

\multirow{4}{*}{L1} & Ours & \hphantom{00}0.0 & $\bestscore{\hphantom{0}36.3}$ & $\bestscore{\hphantom{0}76.3}$ & $\bestscore{\hphantom{0}79.3}$ \\
  & \vima-Gato & \hphantom{00}0.0 ($\bestscore{2.2} \uparrow$) & \hphantom{0}11.5 (11.9 $\uparrow$) & \hphantom{0}41.5 (26.5 $\downarrow$) & \hphantom{0}57.5 (57.0 $\downarrow$) \\
  & \vima-Flamingo & \hphantom{00}0.0 (0.0) & \hphantom{00}2.0 (6.2 $\uparrow$)\hphantom{0} & \hphantom{0}37.7 (33.9 $\downarrow$) & \hphantom{0}52.3 (47.2 $\downarrow$) \\
  & \vima-GPT & \hphantom{00}0.0 (0.0) & \hphantom{00}6.0 (17.0 $\uparrow$)  & \hphantom{0}30.9 (38.9 $\uparrow$) & \hphantom{0}52.8 (47.9 $\downarrow$) \\
\midrule 

\multirow{4}{*}{L2} & Ours & \hphantom{00}0.0 & $\bestscore{\hphantom{0}34.3}$ & $\bestscore{\hphantom{0}75.8}$ & $\bestscore{\hphantom{0}80.1}$ \\
  & \vima-Gato & \hphantom{00}0.0 ($\bestscore{2.0} \uparrow$) & \hphantom{0}10.1 (11.4 $\uparrow$) & \hphantom{0}37.9 (24.6 $\downarrow$) & \hphantom{0}41.2 (53.9 $\uparrow$) \\
  & \vima-Flamingo & \hphantom{00}0.0 (0.0) & \hphantom{00}2.0 (6.0 $\uparrow$)\hphantom{0} & \hphantom{0}33.8 (34.1 $\uparrow$) & \hphantom{0}32.6 (47.1 $\uparrow$)\\
  & \vima-GPT & \hphantom{00}0.0 (0.0) & \hphantom{00}6.0 (15.2 $\uparrow$) & \hphantom{0}29.7 (36.7 $\uparrow$) & \hphantom{0}40.3 (47.4 $\uparrow$) \\
\midrule 

\multirow{4}{*}{L3} & Ours & \hphantom{00}0.0 & $\bestscore{\hphantom{0}15.4}$ & $\bestscore{\hphantom{0}73.2}$ & $\bestscore{\hphantom{0}81.9}$ \\
  & \vima-Gato & \hphantom{00}0.0 ($\bestscore{1.1}$ $\uparrow$) & \hphantom{0}10.2 (10.1 $\downarrow$) & \hphantom{0}34.8 (22.6 $\downarrow$) & \hphantom{0}40.9 (45.6 $\uparrow$) \\
  & \vima-Flamingo & \hphantom{00}0.0 (0.0) & \hphantom{00}1.0 (5.4 $\uparrow$)\hphantom{0} & \hphantom{0}33.1 (31.0 $\downarrow$) & \hphantom{0}33.6 (42.1 $\uparrow$)\\
  & \vima-GPT & \hphantom{00}0.0 (0.0) & \hphantom{00}5.5 (15.0 $\uparrow$) & \hphantom{0}28.6 (35.7 $\uparrow$ ) & \hphantom{0}39.2 (42.2 $\uparrow$) \\
\midrule 

\multirow{4}{*}{L4} & Ours & \hphantom{00}0.0 & $\bestscore{\hphantom{0}17.0}$ & $\bestscore{\hphantom{0}46.0}$ & $\bestscore{\hphantom{0}49.6}$ \\
  & \vima-Gato & \hphantom{00}0.0 (0.0) & \hphantom{00}2.7 (2.5 $\downarrow$) & \hphantom{0}10.8 (5.8 $\downarrow$) & \hphantom{0}12.2 (13.5 $\uparrow$) \\
  & \vima-Flamingo & \hphantom{00}0.0 (0.0) & \hphantom{00}0.5 (0.0 $\downarrow$) & \hphantom{0}11.2 (8.3 $\downarrow$) & \hphantom{0}12.0 (11.1 $\downarrow$) \\
  & \vima-GPT & \hphantom{00}0.0 (0.0) & \hphantom{00}1.1 (4.1 $\uparrow$) & \hphantom{00}7.1 (9.0 $\uparrow$) & \hphantom{0}14.3 (12.1 $\downarrow$) \\

\bottomrule 
\end{tabular}
\end{table}

\subsection{Vary T5 Encoder Sizes}
\label{supp:sec:t5}
We vary the size of the pre-trained T5 encoder~\citep{raffel2020t5} to study the effect of prompt encoding. We experiment with three T5 model capacities: \texttt{t5-small} (30M), \texttt{t5-base} (111M), and \texttt{t5-large} (368M). For all T5 variants, we fine-tune the last two layers and freeze all other layers. We fix the parameter count of the decision-making part to be 200M. As shown in Table~\ref{supp:table:t5}, we find no significant difference among the variants. Thus we set the standard \texttt{t5-base} as default for all our models.

\begin{table}[!htbp]
\caption{Performances of our method with differently sized pre-trained T5 prompt encoder. We fix the parameter count of the decision-making part to be 200M.}
\centering
\begin{tabular}{@{}cccc@{}}
\toprule
   & \textbf{\texttt{t5-small} (30M)} & \textbf{\texttt{t5-base} (111M)} & \textbf{\texttt{t5-large} (368M)} \\ \midrule
L1 & 78.8     & 81.5    & 80.8     \\
L2 & 79.0     & 81.5    & 81.0     \\
L3 & 80.3     & 78.7    & 81.0     \\
L4 & 49.1     & 48.6    & 49.3     \\ \bottomrule
\end{tabular}
\label{supp:table:t5}
\end{table}

\subsection{Policy Robustness}
\label{supp:sec:robustness}
\para{Increasing Amounts of Distractors.} We study the policy robustness against increasing amounts of distractors in scenes. For all tasks being evaluated, we add one more distractor object. We run our largest \vima model with 200M parameters. The result is presented in Table~\ref{supp:table:more_distractors}.

It turns out that the performance of VIMA degrades minimally with more distractors than the training distribution. This indicates that our agent has learned a reasonably robust policy against objects that are irrelevant to the task.

\begin{table}[!htbp]
\caption{Evaluation results on tasks with increased amounts of distractors. We fix the parameter count of the decision-making part to be 200M.}
\centering
\begin{tabular}{@{}rcccc@{}}
\toprule
                                   & \textbf{L1}   & \textbf{L2}   & \textbf{L3}   & \textbf{L4}   \\ \midrule
Original                           & 81.5 & 81.5 & 78.7 & 48.6 \\
More Distractors                   & 78.5 & 78.6 & 72.9 & 47.8 \\
Relevant Performance Decrease (\%) & \hphantom{0}3.6  & \hphantom{0}3.5  & \hphantom{0}7.3  & \hphantom{0}1.6  \\ \bottomrule
\end{tabular}
\label{supp:table:more_distractors}
\end{table}

\para{Imperfect Prompts.} We then study the policy robustness against imperfect prompts, including incomplete prompts (randomly masking out words with \texttt{<UNK>} token) and corrupted prompts (randomly swapping words, which could have changed the task meaning altogether). We run our largest \vima model with 200M parameters, results are shown in Table~\ref{supp:table:imperfect_prompts}.

Our well-trained model exhibits minimal performance decrease when evaluated on masked prompts and minor decrease on corrupted prompts. We attribute this robustness to the high-quality pre-trained T5 language backbone.

\begin{table}[!htbp]
\caption{Evaluation results with incomplete and corrupted prompts. We fix the parameter count of the decision-making part to be 200M.}
\centering
\begin{tabular}{@{}rcccc@{}}
\toprule
                                                       & \textbf{L1}   & \textbf{L2}   & \textbf{L3}   & \textbf{L4}   \\ \midrule
Original                                               & 81.5 & 81.5 & 78.7 & 48.6 \\
Incomplete Prompts                                          & 80.8 & 81.1 & 77.0 & 48.0 \\
Corrupted Prompts                                       & 78.2 & 78.1 & 73.8 & 45.3 \\
Relevant Performance Decrease w/ Incomplete Prompts (\%)   & \hphantom{0}0.8  & \hphantom{0}0.4  & \hphantom{0}2.1  & \hphantom{0}1.2  \\
Relevant Performance Decrease w/ Corrupted Prompts (\%) & \hphantom{0}4.2  & \hphantom{0}4.3  & \hphantom{0}6.6  & \hphantom{0}7.2  \\ \bottomrule
\end{tabular}
\label{supp:table:imperfect_prompts}
\end{table}

\clearpage
\section{Extended Related Work}
\label{supp:sec:extended_related_work}

In this section, we provide an extended review of related work as complementary to Section~\ref{sec:related}.

\para{Multi-Task Learning by Sequence Modeling.} 
In computer vision, Mask R-CNN~\citep{he2017mask}, UberNet~\citep{kokkinos2016ubernet}, and 12-in-1~\citep{lu2020multi} leverage a single backbone model with multiple independent heads for different tasks. UVim~\citep{kolesnikov2022uvim} is another unified approach for vision that uses a language model to generate the guiding code for a second model to predict raw vision outputs.
In multimodal learning, numerous works \citep{lu2022unifiedio,wang2022ofa,zellers2021merlot,zellers2022merlotreserve,buch2022atp,fu2021violet,yang2022icode} investigate the unification of image, video, audio, and/or language modalities to deliver multi-purpose foundation models, although most of which are not equipped with decision-making capabilities.
BEiT-3~\citep{wang2022ibeit3} performs masked data modeling on images, texts and image-text pairs to pre-train a backbone for various downstream tasks. MetaMorph~\citep{gupta2022metamorph} learns a universal controller over a modular robot design space.

\para{Foundation Models for Embodied Agents.}
Embodied agent research~\citep{duan2022survey, batra2020rearrangement, ravichandar2020recent, collins2021review} is adopting the large-scale pre-training paradigm~\citep{yang2023foundation}, powered by a collection of learning environments~\citep{deepmind2020playroom,shridhar2020alfred,savva2019habitat,puig2018virtualhome,team2021openended,toyama2021androidenv,shi2017wob}.
From the aspect of \textbf{pre-training for better representations}, \citet{reid2022wikipedia} fine-tunes from LLM checkpoints to accelerate policy learning. LaTTe~\citep{bucker2022latte} and Embodied-CLIP~\citep{khandelwal2021simple} leverage the frozen visual and textual representations of CLIP~\citep{radford2021clip} for robotic manipulation.
MaskDP~\citep{liu2022masked} pre-trains bidirectional transformers for various downstream embodied tasks.
From the perspective of leveraging \textbf{transformer as agent architecture}, methods such as \citet{dasari2020transformers} and MOSAIC~\citep{zhao2022towards} achieve superior performance in one-shot video imitation tasks. They both use the self-attention mechanism with auxiliary losses such as inverse dynamics loss~\citep{dasari2020transformers} and contrastive loss~\citep{zhao2022towards} to learn robot controllers.
\textit{Instruct}RL~\citep{liu2022instructrl} leverages jointly pre-trained vision-language models as robot agents to perform manipulation tasks.
From the perspective of \textbf{large language models for robot learning}, Socratic Models~\citep{zeng2022socratic} composes multiple vision and language foundation models for multimodal reasoning in videos.
ROSIE~\citep{yu2023scaling} leverages text-to-image diffusion models to augment existing robotic dataset~\citep{brohan2022rt1} via inpainting.
MOO~\citep{minderer2022simple} adopts a similar object-centric representation as ours for open-world object manipulation.
Furthermore, Voyager~\citep{wang2023voyager} develops a LLM-powered agent operating in an open-ended virtual world~\citep{fan2022minedojo}.

\para{Robot Manipulation and Benchmarks.} There are many prior works that are not mentioned in the main paper that study different robotic manipulation tasks, such as instruction following~\citep{shridhar2021cliport,lynch2021language}, constraint satisfaction~\citep{bharadhwaj2021conservative,srinivasan2020learning,thananjeyan2021recoveryrl}, one-shot imitation~\citep{paine2018oneshot,huang2019oneshot,dasari2020transformers,aceituno2021videoimitation, zhao2022towards}, rearrangement~\citep{weihs2021visual,szot2021habitat2,liu2021ocrtoc,rhsani2021ManipulaTHOR,gan2021threedworld,eskin2022rearrangement},
and reasoning~\citep{gupta2019relay,ahmed2021causalworld,toyer2020magical,lim2021multitask}.
Multiple simulation benchmarks are introduced to study the above tasks: 1) \textbf{Indoor simulation environments}: Habitat~\citep{savva2019habitat,szot2021habitat2} is equipped with a high-performance 3D simulator for fast rendering and proposes a suite of common tasks for assistive robots.
2) \textbf{Tabletop environments}: RLBench~\citep{james2019rlbench} and SURREAL~\citep{fan2018surreal,fan2019surrealsystem} are other widely used simulator benchmarks studying robotics manipulation with tabletop settings.
STRETCH-P\&P~\citep{zhang2023learning} studies generalization across goals for reset-free reinforcement learning.
All these aforementioned simulators and benchmarks do not natively support task specification and prompting with multiple modalities.

\clearpage
\section{Limitations and Further Discussions}
\label{supp:sec:limitations}

\para{Reliance on a separate object detector.} \vima inherits the errors from the standalone object detector, which may struggle in the cases of occlusion or out-of-distribution object forms. 
However, using object detectors is not entirely without merits.
First, it allows us to seamlessly switch to stronger detection models when they become available. For example, we can switch to object detectors that are more robust and open-vocabulary, such as OWL-ViT~\citep{minderer2022simple}. This would enable \vima to transfer to real-world scenarios with minimal modifications.
Second, by leveraging pre-trained vision pipelines, several concurrent works have demonstrated the superiority of object-centric representation in robot manipulation. For example, VIOLA~\citep{zhu2022viola} achieves better performance with a pre-trained Region Proposal Network~\citep{ren2015faster}. MOO~\citep{stone2023openworld} also shows that a robot agent with OWL-ViT~\citep{minderer2022simple} as the object detector significantly outperforms RT-1~\citep{brohan2022rt1}, which directly learns from raw pixels, on various real-world manipulation tasks.
In fact, MOO~\citep{stone2023openworld} includes a baseline called ``\vima-like'' that already demonstrates strong performance on real robots under real-world scenarios.
As we witness image segmentation is becoming more robust and general-purpose~\citep{kirillov2023segment}, we envision such design choice will become more effective and further gain more popularity.

\para{Limited simulator realism and task complexity.} Our goal with \vimabench is to explore the multi-task ability, generalization, and understanding of multi-modality. Therefore, these aspects are not the primary focus of this work. However, we envision future works can combine this formulation with more physically realistic simulators such as \citet{zhu2020robosuite}, \citet{srivastava2021behavior}, and \citet{mittal2023orbit}.

\para{Limited action primitives.} We inherit the same high-level action space from well-established prior works, such as Transporter~\citep{zeng2020transporter}. While ``pick-and-place'' and ``wipe'' seem simple, they do cover a wide range of tabletop manipulation tasks and are crucial to industrial use cases like warehouse robots~\citep{1242127,berscheid2020selfsupervised,devin2020selfsupervised,song2019grasping}. While \vima is currently using these two actions, the algorithm design is general-purpose and does not make assumptions about the particular action choices. For example, \vima would require only minimal modifications to support more low-level action spaces like joint-torque control.
\clearpage
\section{Full Tables}
This section contains more detailed tables that correspond to the results in Figure~\ref{fig:scalability}. Specifically, we show breakdown results on each task that constitute the model scaling results in Tables~\ref{supp:table:l1}, \ref{supp:table:l2}, \ref{supp:table:l3}, and \ref{supp:table:l4}.

\begin{table}[h]
\caption{L1 level generalization results. Model indicates robot controller parameter count. Integers in the first row refer to indices of tasks described in Appendix, Sec.~\ref{supp:sec:task_suite}.}
\label{supp:table:l1}
\centering

\resizebox{1.0\textwidth}{!}{
\begin{tabular}{c|r|ccccccccccccc}
\toprule 

\textbf{Model} & \textbf{Method} & \textbf{01} & \textbf{02} & \textbf{03} & \textbf{04} & \textbf{05} & \textbf{06} & \textbf{07} & \textbf{09} & \textbf{11} & \textbf{12} & \textbf{15} & \textbf{16} & \textbf{17} \\ \midrule 

\multirow{4}{*}{2M} & Ours & $\bestscore{100.0}$ & $\bestscore{100.0}$ & $\bestscore{100.0}$ & $\bestscore{\hphantom{0}96.0}$ & \hphantom{0}37.0 & $\bestscore{100.0}$ & $\bestscore{100.0}$ & $\bestscore{\hphantom{00}9.5}$ & $\bestscore{\hphantom{0}87.0}$ & \hphantom{0}64.0 & $\bestscore{\hphantom{0}93.5}$ & $\bestscore{\hphantom{0}45.0}$ & $\bestscore{\hphantom{0}63.0}$ \\
  & \vima-Gato & \hphantom{0}62.0 & \hphantom{0}61.0 & \hphantom{0}22.5 & \hphantom{0}13.5 & \hphantom{00}7.0 & \hphantom{0}44.5 & \hphantom{0}54.0 & \hphantom{00}4.0 & \hphantom{0}48.0 & $\bestscore{\hphantom{0}85.0}$ & \hphantom{0}44.5 & \hphantom{0}43.0 & \hphantom{00}0.0 \\
  & \vima-Flamingo & \hphantom{0}56.0 & \hphantom{0}56.0 & \hphantom{0}53.5 & \hphantom{0}36.5 & $\bestscore{\hphantom{0}37.5}$ & \hphantom{0}45.0 & \hphantom{0}55.5 & \hphantom{00}3.5 & \hphantom{0}54.0 & \hphantom{0}83.5 & \hphantom{0}40.5 & \hphantom{0}28.5 & \hphantom{00}2.0 \\
  & \vima-GPT & \hphantom{0}59.5 & \hphantom{0}50.5 & \hphantom{00}7.5 & \hphantom{00}7.0 & \hphantom{00}0.5 & \hphantom{0}43.5 & \hphantom{0}49.5 & \hphantom{00}2.0 & \hphantom{0}61.5 & \hphantom{0}76.5 & \hphantom{0}27.5 & \hphantom{00}5.0 & \hphantom{00}0.0 \\
\midrule 

\multirow{4}{*}{4M} & Ours & $\bestscore{100.0}$ & $\bestscore{100.0}$ & $\bestscore{100.0}$ & $\bestscore{\hphantom{0}99.5}$ & $\bestscore{\hphantom{0}45.5}$ & $\bestscore{100.0}$ & $\bestscore{100.0}$ & $\bestscore{\hphantom{0}10.5}$ & $\bestscore{\hphantom{0}90.5}$ & $\bestscore{\hphantom{0}90.0}$ & $\bestscore{\hphantom{0}96.5}$ & $\bestscore{\hphantom{0}46.5}$ & $\bestscore{\hphantom{0}51.0}$ \\
  & \vima-Gato & \hphantom{0}61.0 & \hphantom{0}61.5 & \hphantom{00}8.0 & \hphantom{0}46.0 & \hphantom{0}32.5 & \hphantom{0}45.5 & \hphantom{0}57.0 & \hphantom{00}1.0 & \hphantom{0}64.5 & \hphantom{0}86.0 & \hphantom{0}46.5 & \hphantom{0}42.5 & \hphantom{00}2.0 \\
  & \vima-Flamingo & \hphantom{0}61.0 & \hphantom{0}62.0 & \hphantom{0}57.5 & \hphantom{0}47.5 & \hphantom{0}45.0 & \hphantom{0}49.5 & \hphantom{0}59.5 & \hphantom{00}5.5 & \hphantom{0}80.0 & \hphantom{0}83.5 & \hphantom{0}40.5 & \hphantom{0}43.0 & \hphantom{00}2.0 \\
  & \vima-GPT & \hphantom{0}58.0 & \hphantom{0}55.0 & \hphantom{0}17.5 & \hphantom{0}25.0 & \hphantom{0}12.0 & \hphantom{0}47.5 & \hphantom{0}54.5 & \hphantom{00}3.0 & \hphantom{0}59.5 & \hphantom{0}80.5 & \hphantom{0}27.0 & \hphantom{0}41.5 & \hphantom{00}0.5 \\
\midrule 

\multirow{4}{*}{9M} & Ours & $\bestscore{100.0}$ & $\bestscore{100.0}$ & $\bestscore{100.0}$ & $\bestscore{\hphantom{0}99.5}$ & $\bestscore{\hphantom{0}51.5}$ & $\bestscore{100.0}$ & $\bestscore{100.0}$ & $\bestscore{\hphantom{0}13.0}$ & $\bestscore{\hphantom{0}82.5}$ & \hphantom{0}58.5 & $\bestscore{\hphantom{0}96.0}$ & $\bestscore{\hphantom{0}42.0}$ & $\bestscore{\hphantom{0}63.5}$ \\
  & \vima-Gato & \hphantom{0}59.0 & \hphantom{0}61.0 & \hphantom{0}41.0 & \hphantom{0}50.5 & \hphantom{0}38.5 & \hphantom{0}47.5 & \hphantom{0}59.5 & \hphantom{00}9.5 & \hphantom{0}58.0 & \hphantom{0}80.5 & \hphantom{0}44.0 & \hphantom{0}24.0 & \hphantom{00}2.5 \\
  & \vima-Flamingo & \hphantom{0}58.5 & \hphantom{0}60.0 & \hphantom{0}46.0 & \hphantom{0}49.0 & \hphantom{0}42.5 & \hphantom{0}45.5 & \hphantom{0}60.5 & \hphantom{00}4.0 & \hphantom{0}66.5 & \hphantom{0}81.5 & \hphantom{0}36.5 & \hphantom{0}41.5 & \hphantom{00}1.0 \\
  & \vima-GPT & \hphantom{0}58.5 & \hphantom{0}54.5 & \hphantom{0}40.5 & \hphantom{0}47.5 & \hphantom{0}37.5 & \hphantom{0}47.5 & \hphantom{0}58.5 & \hphantom{00}9.0 & \hphantom{0}72.0 & $\bestscore{\hphantom{0}85.0}$ & \hphantom{0}38.5 & \hphantom{0}34.0 & \hphantom{00}1.0 \\
\midrule 

\multirow{4}{*}{20M} & Ours & $\bestscore{100.0}$ & $\bestscore{100.0}$ & $\bestscore{100.0}$ & $\bestscore{100.0}$ & $\bestscore{\hphantom{0}59.5}$ & $\bestscore{100.0}$ & $\bestscore{100.0}$ & $\bestscore{\hphantom{0}13.5}$ & \hphantom{0}74.0 & \hphantom{0}72.5 & $\bestscore{\hphantom{0}96.5}$ & \hphantom{0}39.5 & $\bestscore{\hphantom{0}47.5}$ \\
  & \vima-Gato & \hphantom{0}61.5 & \hphantom{0}62.0 & \hphantom{0}32.5 & \hphantom{0}49.0 & \hphantom{0}38.0 & \hphantom{0}46.0 & \hphantom{0}60.0 & \hphantom{00}5.0 & \hphantom{0}68.0 & \hphantom{0}83.0 & \hphantom{0}47.0 & $\bestscore{\hphantom{0}46.5}$ & \hphantom{00}2.0 \\
  & \vima-Flamingo & \hphantom{0}63.0 & \hphantom{0}61.5 & \hphantom{0}55.0 & \hphantom{0}50.0 & \hphantom{0}42.5 & \hphantom{0}41.5 & \hphantom{0}58.0 & \hphantom{00}6.0 & \hphantom{0}62.0 & \hphantom{0}83.0 & \hphantom{0}44.0 & \hphantom{0}38.5 & \hphantom{00}1.0 \\
  & \vima-GPT & \hphantom{0}60.5 & \hphantom{0}64.0 & \hphantom{0}50.5 & \hphantom{0}44.0 & \hphantom{0}41.0 & \hphantom{0}48.0 & \hphantom{0}61.5 & \hphantom{00}7.0 & $\bestscore{\hphantom{0}85.0}$ & $\bestscore{\hphantom{0}84.0}$ & \hphantom{0}44.5 & \hphantom{0}39.0 & \hphantom{00}2.5 \\
\midrule 

\multirow{4}{*}{43M} & Ours & $\bestscore{100.0}$ & $\bestscore{100.0}$ & $\bestscore{100.0}$ & $\bestscore{100.0}$ & $\bestscore{\hphantom{0}57.0}$ & $\bestscore{\hphantom{0}99.5}$ & $\bestscore{100.0}$ & $\bestscore{\hphantom{0}15.0}$ & $\bestscore{\hphantom{0}86.0}$ & \hphantom{0}69.5 & $\bestscore{\hphantom{0}99.0}$ & \hphantom{0}40.0 & $\bestscore{\hphantom{0}51.5}$ \\
  & \vima-Gato & \hphantom{0}57.0 & \hphantom{0}65.5 & \hphantom{0}59.0 & \hphantom{0}57.5 & \hphantom{0}43.5 & \hphantom{0}50.0 & \hphantom{0}56.0 & \hphantom{00}5.0 & \hphantom{0}67.0 & $\bestscore{\hphantom{0}83.5}$ & \hphantom{0}63.0 & \hphantom{0}37.0 & \hphantom{00}0.0 \\
  & \vima-Flamingo & \hphantom{0}54.5 & \hphantom{0}57.0 & \hphantom{0}54.5 & \hphantom{0}54.0 & \hphantom{0}45.0 & \hphantom{0}43.5 & \hphantom{0}55.5 & \hphantom{00}6.0 & \hphantom{0}67.5 & \hphantom{0}82.5 & \hphantom{0}49.0 & $\bestscore{\hphantom{0}40.5}$ & \hphantom{00}1.5 \\
  & \vima-GPT & \hphantom{0}58.0 & \hphantom{0}60.5 & \hphantom{0}69.5 & \hphantom{0}53.5 & \hphantom{0}41.5 & \hphantom{0}47.0 & \hphantom{0}55.5 & \hphantom{00}4.0 & \hphantom{0}66.5 & \hphantom{0}81.5 & \hphantom{0}45.0 & $\bestscore{\hphantom{0}40.5}$ & \hphantom{00}1.5 \\
\midrule 

\multirow{4}{*}{92M} & Ours & $\bestscore{100.0}$ & $\bestscore{100.0}$ & $\bestscore{\hphantom{0}99.5}$ & $\bestscore{100.0}$ & $\bestscore{\hphantom{0}58.0}$ & $\bestscore{100.0}$ & $\bestscore{100.0}$ & $\bestscore{\hphantom{0}14.0}$ & $\bestscore{\hphantom{0}80.5}$ & $\bestscore{\hphantom{0}92.0}$ & $\bestscore{\hphantom{0}98.5}$ & \hphantom{0}40.5 & $\bestscore{\hphantom{0}48.5}$ \\
  & \vima-Gato & \hphantom{0}76.5 & \hphantom{0}59.5 & \hphantom{0}90.0 & \hphantom{0}56.5 & \hphantom{0}44.5 & \hphantom{0}48.5 & \hphantom{0}68.5 & $\bestscore{\hphantom{0}14.0}$ & \hphantom{0}64.5 & \hphantom{0}89.5 & \hphantom{0}85.0 & $\bestscore{\hphantom{0}43.0}$ & \hphantom{00}1.5 \\
  & \vima-Flamingo & \hphantom{0}56.0 & \hphantom{0}56.0 & \hphantom{0}65.5 & \hphantom{0}50.5 & \hphantom{0}41.0 & \hphantom{0}48.0 & \hphantom{0}56.0 & \hphantom{00}3.0 & \hphantom{0}70.0 & \hphantom{0}87.0 & \hphantom{0}41.5 & \hphantom{0}38.0 & \hphantom{00}2.0 \\
  & \vima-GPT & \hphantom{0}57.0 & \hphantom{0}57.5 & \hphantom{0}58.5 & \hphantom{0}53.0 & \hphantom{0}45.0 & \hphantom{0}51.0 & \hphantom{0}61.0 & \hphantom{00}8.0 & \hphantom{0}65.5 & \hphantom{0}87.0 & \hphantom{0}46.0 & \hphantom{0}33.0 & \hphantom{00}1.0 \\
\midrule 

\multirow{4}{*}{200M} & Ours & $\bestscore{100.0}$ & $\bestscore{100.0}$ & $\bestscore{\hphantom{0}99.5}$ & $\bestscore{100.0}$ & $\bestscore{\hphantom{0}56.5}$ & $\bestscore{100.0}$ & $\bestscore{100.0}$ & $\bestscore{\hphantom{0}18.0}$ & $\bestscore{\hphantom{0}77.0}$ & $\bestscore{\hphantom{0}93.0}$ & $\bestscore{\hphantom{0}97.0}$ & $\bestscore{\hphantom{0}76.5}$ & $\bestscore{\hphantom{0}43.0}$ \\
  & \vima-Gato & \hphantom{0}79.0 & \hphantom{0}68.0 & \hphantom{0}91.5 & \hphantom{0}57.0 & \hphantom{0}44.5 & \hphantom{0}54.0 & \hphantom{0}74.0 & $\bestscore{\hphantom{0}18.0}$ & \hphantom{0}61.0 & \hphantom{0}88.5 & \hphantom{0}83.5 & \hphantom{0}33.5 & \hphantom{00}2.5 \\
  & \vima-Flamingo & \hphantom{0}56.0 & \hphantom{0}58.5 & \hphantom{0}63.0 & \hphantom{0}48.5 & \hphantom{0}38.0 & \hphantom{0}48.5 & \hphantom{0}62.5 & \hphantom{00}3.5 & \hphantom{0}66.5 & \hphantom{0}86.0 & \hphantom{0}40.0 & \hphantom{0}43.5 & \hphantom{00}2.5 \\
  & \vima-GPT & \hphantom{0}62.0 & \hphantom{0}57.5 & \hphantom{0}41.0 & \hphantom{0}55.5 & \hphantom{0}45.5 & \hphantom{0}47.5 & \hphantom{0}54.5 & \hphantom{00}8.5 & $\bestscore{\hphantom{0}77.0}$ & \hphantom{0}81.5 & \hphantom{0}41.0 & \hphantom{0}38.0 & \hphantom{00}0.5 \\

\bottomrule 
\end{tabular}
}

\end{table}

\begin{table}
\caption{L2 level generalization results. Model indicates robot controller parameter count. Integers in the first row refer to indices of tasks described in Appendix, Sec.~\ref{supp:sec:task_suite}.}
\label{supp:table:l2}
\centering

\resizebox{1.0\textwidth}{!}{
\begin{tabular}{c|r|ccccccccccccc}
\toprule 

\textbf{Model} & \textbf{Method} & \textbf{01} & \textbf{02} & \textbf{03} & \textbf{04} & \textbf{05} & \textbf{06} & \textbf{07} & \textbf{09} & \textbf{11} & \textbf{12} & \textbf{15} & \textbf{16} & \textbf{17} \\ \midrule 

\multirow{4}{*}{2M} & Ours & $\bestscore{100.0}$ & $\bestscore{100.0}$ & $\bestscore{100.0}$ & $\bestscore{\hphantom{0}95.5}$ & $\bestscore{\hphantom{0}37.5}$ & $\bestscore{100.0}$ & $\bestscore{100.0}$ & $\bestscore{\hphantom{0}17.5}$ & $\bestscore{\hphantom{0}87.5}$ & \hphantom{0}67.0 & $\bestscore{\hphantom{0}97.5}$ & $\bestscore{\hphantom{0}46.0}$ & $\bestscore{\hphantom{0}54.5}$ \\
  & \vima-Gato & \hphantom{0}49.5 & \hphantom{0}49.0 & \hphantom{0}23.0 & \hphantom{0}17.5 & \hphantom{00}5.0 & \hphantom{0}47.5 & \hphantom{0}46.5 & \hphantom{00}5.5 & \hphantom{0}50.0 & $\bestscore{\hphantom{0}82.5}$ & \hphantom{0}49.0 & \hphantom{0}42.0 & \hphantom{00}0.5 \\
  & \vima-Flamingo & \hphantom{0}45.5 & \hphantom{0}46.0 & \hphantom{0}56.0 & \hphantom{0}39.5 & \hphantom{0}35.5 & \hphantom{0}49.0 & \hphantom{0}47.0 & \hphantom{00}9.0 & \hphantom{0}53.0 & \hphantom{0}80.0 & \hphantom{0}43.0 & \hphantom{0}29.5 & \hphantom{00}1.0 \\
  & \vima-GPT & \hphantom{0}51.0 & \hphantom{0}45.5 & \hphantom{00}9.5 & \hphantom{00}7.0 & \hphantom{00}0.5 & \hphantom{0}45.5 & \hphantom{0}45.0 & \hphantom{00}0.0 & \hphantom{0}65.0 & \hphantom{0}81.5 & \hphantom{0}32.0 & \hphantom{00}5.0 & \hphantom{00}0.0 \\
\midrule 

\multirow{4}{*}{4M} & Ours & $\bestscore{100.0}$ & $\bestscore{100.0}$ & $\bestscore{100.0}$ & $\bestscore{\hphantom{0}99.5}$ & $\bestscore{\hphantom{0}44.5}$ & $\bestscore{\hphantom{0}99.5}$ & $\bestscore{100.0}$ & $\bestscore{\hphantom{0}14.5}$ & $\bestscore{\hphantom{0}89.5}$ & $\bestscore{\hphantom{0}91.5}$ & $\bestscore{\hphantom{0}95.5}$ & $\bestscore{\hphantom{0}43.0}$ & $\bestscore{\hphantom{0}52.5}$ \\
  & \vima-Gato & \hphantom{0}44.5 & \hphantom{0}52.0 & \hphantom{00}9.0 & \hphantom{0}39.0 & \hphantom{0}28.0 & \hphantom{0}49.5 & \hphantom{0}48.5 & \hphantom{00}2.0 & \hphantom{0}64.0 & \hphantom{0}86.5 & \hphantom{0}44.5 & \hphantom{0}42.5 & \hphantom{00}2.0 \\
  & \vima-Flamingo & \hphantom{0}49.5 & \hphantom{0}50.5 & \hphantom{0}51.0 & \hphantom{0}48.0 & \hphantom{0}43.0 & \hphantom{0}50.5 & \hphantom{0}53.5 & \hphantom{00}5.5 & \hphantom{0}81.5 & \hphantom{0}82.5 & \hphantom{0}48.5 & \hphantom{0}39.5 & \hphantom{00}1.0 \\
  & \vima-GPT & \hphantom{0}50.5 & \hphantom{0}49.5 & \hphantom{0}16.5 & \hphantom{0}25.5 & \hphantom{0}12.0 & \hphantom{0}41.0 & \hphantom{0}47.0 & \hphantom{00}4.0 & \hphantom{0}63.0 & \hphantom{0}79.0 & \hphantom{0}28.5 & \hphantom{0}39.0 & \hphantom{00}0.0 \\
\midrule 

\multirow{4}{*}{9M} & Ours & $\bestscore{100.0}$ & $\bestscore{100.0}$ & $\bestscore{100.0}$ & $\bestscore{100.0}$ & $\bestscore{\hphantom{0}49.5}$ & $\bestscore{100.0}$ & $\bestscore{100.0}$ & $\bestscore{\hphantom{0}19.0}$ & $\bestscore{\hphantom{0}80.5}$ & \hphantom{0}65.0 & $\bestscore{\hphantom{0}95.5}$ & $\bestscore{\hphantom{0}42.0}$ & $\bestscore{\hphantom{0}66.0}$ \\
  & \vima-Gato & \hphantom{0}47.0 & \hphantom{0}44.5 & \hphantom{0}39.5 & \hphantom{0}46.5 & \hphantom{0}37.5 & \hphantom{0}48.5 & \hphantom{0}51.0 & \hphantom{00}5.5 & \hphantom{0}59.0 & \hphantom{0}83.0 & \hphantom{0}51.5 & \hphantom{0}23.5 & \hphantom{00}1.0 \\
  & \vima-Flamingo & \hphantom{0}48.0 & \hphantom{0}47.5 & \hphantom{0}49.0 & \hphantom{0}52.5 & \hphantom{0}42.0 & \hphantom{0}47.5 & \hphantom{0}48.5 & \hphantom{00}8.5 & \hphantom{0}66.0 & \hphantom{0}81.5 & \hphantom{0}45.5 & $\bestscore{\hphantom{0}42.0}$ & \hphantom{00}2.0 \\
  & \vima-GPT & \hphantom{0}48.5 & \hphantom{0}47.0 & \hphantom{0}43.5 & \hphantom{0}47.0 & \hphantom{0}37.0 & \hphantom{0}47.5 & \hphantom{0}45.5 & \hphantom{0}10.5 & \hphantom{0}74.5 & $\bestscore{\hphantom{0}85.0}$ & \hphantom{0}43.5 & \hphantom{0}33.0 & \hphantom{00}1.0 \\
\midrule 

\multirow{4}{*}{20M} & Ours & $\bestscore{100.0}$ & $\bestscore{100.0}$ & $\bestscore{100.0}$ & $\bestscore{100.0}$ & $\bestscore{\hphantom{0}61.0}$ & $\bestscore{100.0}$ & $\bestscore{100.0}$ & $\bestscore{\hphantom{0}16.5}$ & \hphantom{0}75.5 & \hphantom{0}75.0 & $\bestscore{\hphantom{0}96.0}$ & \hphantom{0}37.5 & $\bestscore{\hphantom{0}47.5}$ \\
  & \vima-Gato & \hphantom{0}44.0 & \hphantom{0}51.5 & \hphantom{0}39.0 & \hphantom{0}51.0 & \hphantom{0}38.5 & \hphantom{0}47.5 & \hphantom{0}52.5 & \hphantom{00}6.0 & \hphantom{0}65.5 & $\bestscore{\hphantom{0}84.0}$ & \hphantom{0}52.5 & \hphantom{0}40.5 & \hphantom{00}1.0 \\
  & \vima-Flamingo & \hphantom{0}48.5 & \hphantom{0}49.0 & \hphantom{0}55.5 & \hphantom{0}48.0 & \hphantom{0}42.5 & \hphantom{0}46.5 & \hphantom{0}52.0 & \hphantom{00}6.0 & \hphantom{0}66.0 & \hphantom{0}82.0 & \hphantom{0}47.5 & \hphantom{0}37.0 & \hphantom{00}0.5 \\
  & \vima-GPT & \hphantom{0}50.5 & \hphantom{0}49.5 & \hphantom{0}53.0 & \hphantom{0}44.5 & \hphantom{0}43.5 & \hphantom{0}47.0 & \hphantom{0}46.0 & \hphantom{00}8.0 & $\bestscore{\hphantom{0}83.5}$ & \hphantom{0}80.0 & \hphantom{0}46.5 & $\bestscore{\hphantom{0}41.0}$ & \hphantom{00}2.5 \\
\midrule 

\multirow{4}{*}{43M} & Ours & $\bestscore{100.0}$ & $\bestscore{100.0}$ & $\bestscore{100.0}$ & $\bestscore{100.0}$ & $\bestscore{\hphantom{0}54.5}$ & $\bestscore{100.0}$ & $\bestscore{100.0}$ & $\bestscore{\hphantom{0}14.5}$ & $\bestscore{\hphantom{0}83.5}$ & \hphantom{0}69.0 & $\bestscore{\hphantom{0}98.0}$ & \hphantom{0}38.5 & $\bestscore{\hphantom{0}51.5}$ \\
  & \vima-Gato & \hphantom{0}50.0 & \hphantom{0}51.5 & \hphantom{0}53.0 & \hphantom{0}57.5 & \hphantom{0}42.5 & \hphantom{0}47.0 & \hphantom{0}51.0 & \hphantom{00}8.5 & \hphantom{0}67.0 & $\bestscore{\hphantom{0}83.0}$ & \hphantom{0}63.5 & \hphantom{0}32.0 & \hphantom{00}0.5 \\
  & \vima-Flamingo & \hphantom{0}48.0 & \hphantom{0}46.5 & \hphantom{0}52.0 & \hphantom{0}51.5 & \hphantom{0}43.5 & \hphantom{0}45.0 & \hphantom{0}51.5 & \hphantom{00}5.0 & \hphantom{0}68.0 & \hphantom{0}81.5 & \hphantom{0}52.5 & $\bestscore{\hphantom{0}44.0}$ & \hphantom{00}1.5 \\
  & \vima-GPT & \hphantom{0}45.0 & \hphantom{0}49.0 & \hphantom{0}64.5 & \hphantom{0}53.5 & \hphantom{0}40.0 & \hphantom{0}46.5 & \hphantom{0}48.5 & \hphantom{00}8.5 & \hphantom{0}68.0 & \hphantom{0}82.0 & \hphantom{0}50.0 & \hphantom{0}40.0 & \hphantom{00}1.5 \\
\midrule 

\multirow{4}{*}{92M} & Ours & $\bestscore{100.0}$ & $\bestscore{100.0}$ & $\bestscore{\hphantom{0}99.0}$ & $\bestscore{100.0}$ & $\bestscore{\hphantom{0}57.5}$ & $\bestscore{\hphantom{0}99.5}$ & $\bestscore{100.0}$ & $\bestscore{\hphantom{0}19.5}$ & $\bestscore{\hphantom{0}81.5}$ & \hphantom{0}92.0 & $\bestscore{\hphantom{0}97.5}$ & $\bestscore{\hphantom{0}42.0}$ & $\bestscore{\hphantom{0}53.5}$ \\
  & \vima-Gato & \hphantom{0}64.5 & \hphantom{0}50.0 & \hphantom{0}83.0 & \hphantom{0}56.5 & \hphantom{0}46.0 & \hphantom{0}55.5 & \hphantom{0}54.5 & \hphantom{0}10.5 & \hphantom{0}64.5 & $\bestscore{\hphantom{0}92.5}$ & \hphantom{0}81.0 & $\bestscore{\hphantom{0}42.0}$ & \hphantom{00}1.0 \\
  & \vima-Flamingo & \hphantom{0}53.0 & \hphantom{0}48.5 & \hphantom{0}67.5 & \hphantom{0}53.0 & \hphantom{0}43.0 & \hphantom{0}49.0 & \hphantom{0}53.0 & \hphantom{00}4.5 & \hphantom{0}67.0 & \hphantom{0}84.0 & \hphantom{0}50.0 & \hphantom{0}40.0 & \hphantom{00}1.0 \\
  & \vima-GPT & \hphantom{0}50.5 & \hphantom{0}55.0 & \hphantom{0}55.5 & \hphantom{0}54.5 & \hphantom{0}43.0 & \hphantom{0}51.5 & \hphantom{0}54.5 & \hphantom{0}10.5 & \hphantom{0}68.5 & \hphantom{0}87.0 & \hphantom{0}49.5 & \hphantom{0}34.0 & \hphantom{00}3.0 \\
\midrule 

\multirow{4}{*}{200M} & Ours & $\bestscore{100.0}$ & $\bestscore{100.0}$ & $\bestscore{\hphantom{0}99.5}$ & $\bestscore{100.0}$ & $\bestscore{\hphantom{0}54.5}$ & $\bestscore{100.0}$ & $\bestscore{100.0}$ & $\bestscore{\hphantom{0}17.5}$ & $\bestscore{\hphantom{0}77.0}$ & $\bestscore{\hphantom{0}93.0}$ & $\bestscore{\hphantom{0}98.5}$ & $\bestscore{\hphantom{0}75.0}$ & $\bestscore{\hphantom{0}45.0}$ \\
  & \vima-Gato & \hphantom{0}56.5 & \hphantom{0}53.5 & \hphantom{0}88.0 & \hphantom{0}55.5 & \hphantom{0}43.5 & \hphantom{0}55.5 & \hphantom{0}53.0 & \hphantom{0}14.0 & \hphantom{0}63.0 & \hphantom{0}90.5 & \hphantom{0}81.5 & \hphantom{0}33.0 & \hphantom{00}4.0 \\
  & \vima-Flamingo & \hphantom{0}51.0 & \hphantom{0}52.5 & \hphantom{0}61.5 & \hphantom{0}49.5 & \hphantom{0}38.5 & \hphantom{0}47.5 & \hphantom{0}55.5 & \hphantom{00}5.5 & \hphantom{0}70.5 & \hphantom{0}82.0 & \hphantom{0}42.0 & \hphantom{0}39.0 & \hphantom{00}3.0 \\
  & \vima-GPT & \hphantom{0}52.0 & \hphantom{0}52.0 & \hphantom{0}49.5 & \hphantom{0}54.5 & \hphantom{0}45.5 & \hphantom{0}52.5 & \hphantom{0}51.0 & \hphantom{0}11.0 & \hphantom{0}76.5 & \hphantom{0}84.0 & \hphantom{0}43.0 & \hphantom{0}38.0 & \hphantom{00}0.5 \\

\bottomrule 

\end{tabular}
}

\end{table}

\begin{table}
\caption{L3 level generalization results. Model indicates robot controller parameter count. Integers in the first row refer to indices of tasks described in Appendix, Sec.~\ref{supp:sec:task_suite}.}
\label{supp:table:l3}
\centering

\resizebox{1.0\textwidth}{!}{
\begin{tabular}{c|r|cccccccccccc}
\toprule 

\textbf{Model} & \textbf{Method} & \textbf{01} & \textbf{02} & \textbf{03} & \textbf{04} & \textbf{05} & \textbf{06} & \textbf{07} & \textbf{09} & \textbf{11} & \textbf{15} & \textbf{16} & \textbf{17} \\ \midrule 

\multirow{4}{*}{2M} & Ours & $\bestscore{100.0}$ & $\bestscore{100.0}$ & $\bestscore{100.0}$ & $\bestscore{\hphantom{0}98.0}$ & $\bestscore{\hphantom{0}34.5}$ & $\bestscore{100.0}$ & $\bestscore{\hphantom{0}99.5}$ & $\bestscore{\hphantom{0}17.0}$ & $\bestscore{\hphantom{0}97.5}$ & $\bestscore{\hphantom{0}94.0}$ & $\bestscore{\hphantom{0}48.5}$ & $\bestscore{\hphantom{0}39.0}$ \\
  & \vima-Gato & \hphantom{0}45.5 & \hphantom{0}48.0 & \hphantom{0}28.0 & \hphantom{0}23.0 & \hphantom{00}3.0 & \hphantom{0}45.5 & \hphantom{0}45.0 & \hphantom{00}2.5 & \hphantom{0}40.5 & \hphantom{0}29.5 & \hphantom{0}37.0 & \hphantom{00}1.0 \\
  & \vima-Flamingo & \hphantom{0}41.5 & \hphantom{0}54.5 & \hphantom{0}50.5 & \hphantom{0}39.5 & \hphantom{0}29.0 & \hphantom{0}45.0 & \hphantom{0}49.5 & \hphantom{00}5.5 & \hphantom{0}57.5 & \hphantom{0}22.5 & \hphantom{0}25.0 & \hphantom{00}0.0 \\
  & \vima-GPT & \hphantom{0}48.5 & \hphantom{0}50.0 & \hphantom{00}5.0 & \hphantom{00}7.0 & \hphantom{00}2.5 & \hphantom{0}47.0 & \hphantom{0}45.5 & \hphantom{00}2.0 & \hphantom{0}69.5 & \hphantom{0}22.5 & \hphantom{00}5.0 & \hphantom{00}0.0 \\
\midrule 

\multirow{4}{*}{4M} & Ours & $\bestscore{\hphantom{0}99.5}$ & $\bestscore{100.0}$ & $\bestscore{100.0}$ & $\bestscore{\hphantom{0}98.0}$ & $\bestscore{\hphantom{0}44.0}$ & $\bestscore{\hphantom{0}99.5}$ & $\bestscore{\hphantom{0}99.5}$ & $\bestscore{\hphantom{0}12.0}$ & $\bestscore{\hphantom{0}92.5}$ & $\bestscore{\hphantom{0}98.5}$ & $\bestscore{\hphantom{0}47.0}$ & $\bestscore{\hphantom{0}43.5}$ \\
  & \vima-Gato & \hphantom{0}44.5 & \hphantom{0}55.0 & \hphantom{00}9.5 & \hphantom{0}37.5 & \hphantom{0}24.5 & \hphantom{0}47.0 & \hphantom{0}50.0 & \hphantom{00}3.5 & \hphantom{0}60.0 & \hphantom{0}30.5 & \hphantom{0}37.5 & \hphantom{00}0.0 \\
  & \vima-Flamingo & \hphantom{0}46.0 & \hphantom{0}53.5 & \hphantom{0}59.0 & \hphantom{0}49.5 & \hphantom{0}35.5 & \hphantom{0}47.5 & \hphantom{0}48.0 & \hphantom{00}7.0 & \hphantom{0}87.5 & \hphantom{0}30.5 & \hphantom{0}39.5 & \hphantom{00}0.0 \\
  & \vima-GPT & \hphantom{0}44.0 & \hphantom{0}47.0 & \hphantom{0}14.5 & \hphantom{0}22.0 & \hphantom{00}9.0 & \hphantom{0}39.5 & \hphantom{0}40.0 & \hphantom{00}2.0 & \hphantom{0}62.0 & \hphantom{0}28.5 & \hphantom{0}43.0 & \hphantom{00}1.0 \\
\midrule 

\multirow{4}{*}{9M} & Ours & $\bestscore{\hphantom{0}99.5}$ & $\bestscore{100.0}$ & $\bestscore{100.0}$ & $\bestscore{\hphantom{0}98.5}$ & $\bestscore{\hphantom{0}44.5}$ & $\bestscore{\hphantom{0}99.5}$ & $\bestscore{\hphantom{0}99.5}$ & $\bestscore{\hphantom{0}18.5}$ & $\bestscore{\hphantom{0}88.5}$ & $\bestscore{\hphantom{0}98.5}$ & $\bestscore{\hphantom{0}48.5}$ & $\bestscore{\hphantom{0}46.5}$ \\
  & \vima-Gato & \hphantom{0}44.5 & \hphantom{0}53.5 & \hphantom{0}42.5 & \hphantom{0}52.0 & \hphantom{0}28.0 & \hphantom{0}46.5 & \hphantom{0}51.5 & \hphantom{00}6.0 & \hphantom{0}67.0 & \hphantom{0}35.0 & \hphantom{0}23.0 & \hphantom{00}0.5 \\
  & \vima-Flamingo & \hphantom{0}44.5 & \hphantom{0}53.0 & \hphantom{0}53.0 & \hphantom{0}48.5 & \hphantom{0}33.0 & \hphantom{0}41.0 & \hphantom{0}45.5 & \hphantom{00}8.0 & \hphantom{0}72.5 & \hphantom{0}27.0 & \hphantom{0}44.5 & \hphantom{00}0.5 \\
  & \vima-GPT & \hphantom{0}49.0 & \hphantom{0}50.5 & \hphantom{0}39.0 & \hphantom{0}46.5 & \hphantom{0}30.5 & \hphantom{0}43.0 & \hphantom{0}52.0 & \hphantom{00}6.5 & \hphantom{0}84.0 & \hphantom{0}31.5 & \hphantom{0}35.0 & \hphantom{00}0.5 \\
\midrule 

\multirow{4}{*}{20M} & Ours & $\bestscore{\hphantom{0}98.0}$ & $\bestscore{100.0}$ & $\bestscore{100.0}$ & $\bestscore{\hphantom{0}98.5}$ & $\bestscore{\hphantom{0}55.5}$ & $\bestscore{100.0}$ & $\bestscore{\hphantom{0}99.5}$ & $\bestscore{\hphantom{0}15.0}$ & \hphantom{0}88.5 & $\bestscore{\hphantom{0}99.5}$ & $\bestscore{\hphantom{0}44.0}$ & $\bestscore{\hphantom{0}29.5}$ \\
  & \vima-Gato & \hphantom{0}46.5 & \hphantom{0}55.0 & \hphantom{0}44.5 & \hphantom{0}57.0 & \hphantom{0}31.5 & \hphantom{0}47.5 & \hphantom{0}51.5 & \hphantom{00}2.5 & \hphantom{0}72.5 & \hphantom{0}30.5 & $\bestscore{\hphantom{0}44.0}$ & \hphantom{00}0.0 \\
  & \vima-Flamingo & \hphantom{0}47.0 & \hphantom{0}54.5 & \hphantom{0}53.0 & \hphantom{0}55.0 & \hphantom{0}36.0 & \hphantom{0}42.5 & \hphantom{0}48.0 & \hphantom{00}6.5 & \hphantom{0}70.0 & \hphantom{0}33.0 & \hphantom{0}41.5 & \hphantom{00}0.0 \\
  & \vima-GPT & \hphantom{0}50.0 & \hphantom{0}60.5 & \hphantom{0}56.5 & \hphantom{0}48.0 & \hphantom{0}33.5 & \hphantom{0}51.0 & \hphantom{0}46.0 & \hphantom{00}6.5 & $\bestscore{\hphantom{0}92.5}$ & \hphantom{0}32.5 & \hphantom{0}43.5 & \hphantom{00}1.5 \\
\midrule 

\multirow{4}{*}{43M} & Ours & $\bestscore{\hphantom{0}99.0}$ & $\bestscore{100.0}$ & $\bestscore{100.0}$ & $\bestscore{\hphantom{0}98.0}$ & $\bestscore{\hphantom{0}47.5}$ & $\bestscore{100.0}$ & $\bestscore{\hphantom{0}99.5}$ & $\bestscore{\hphantom{0}18.5}$ & $\bestscore{\hphantom{0}93.0}$ & $\bestscore{\hphantom{0}98.0}$ & $\bestscore{\hphantom{0}45.0}$ & $\bestscore{\hphantom{0}84.0}$ \\
  & \vima-Gato & \hphantom{0}44.0 & \hphantom{0}55.0 & \hphantom{0}59.5 & \hphantom{0}58.0 & \hphantom{0}34.0 & \hphantom{0}49.0 & \hphantom{0}54.0 & \hphantom{00}7.0 & \hphantom{0}74.0 & \hphantom{0}40.0 & \hphantom{0}35.0 & \hphantom{00}0.5 \\
  & \vima-Flamingo & \hphantom{0}47.0 & \hphantom{0}54.0 & \hphantom{0}56.5 & \hphantom{0}52.5 & \hphantom{0}37.0 & \hphantom{0}46.5 & \hphantom{0}44.5 & \hphantom{00}6.5 & \hphantom{0}69.5 & \hphantom{0}27.0 & \hphantom{0}43.0 & \hphantom{00}0.0 \\
  & \vima-GPT & \hphantom{0}47.5 & \hphantom{0}57.0 & \hphantom{0}61.0 & \hphantom{0}50.0 & \hphantom{0}34.5 & \hphantom{0}48.0 & \hphantom{0}53.5 & \hphantom{00}8.0 & \hphantom{0}74.0 & \hphantom{0}40.5 & \hphantom{0}41.5 & \hphantom{00}0.5 \\
\midrule 

\multirow{4}{*}{92M} & Ours & $\bestscore{\hphantom{0}99.0}$ & $\bestscore{\hphantom{0}99.5}$ & $\bestscore{\hphantom{0}99.5}$ & $\bestscore{\hphantom{0}97.0}$ & $\bestscore{\hphantom{0}58.0}$ & $\bestscore{100.0}$ & $\bestscore{\hphantom{0}99.0}$ & \hphantom{0}13.0 & $\bestscore{\hphantom{0}94.5}$ & $\bestscore{\hphantom{0}99.0}$ & \hphantom{0}42.0 & $\bestscore{\hphantom{0}82.5}$ \\
  & \vima-Gato & \hphantom{0}61.5 & \hphantom{0}54.0 & \hphantom{0}73.0 & \hphantom{0}56.0 & \hphantom{0}36.0 & \hphantom{0}50.0 & \hphantom{0}48.0 & $\bestscore{\hphantom{0}17.0}$ & \hphantom{0}66.5 & \hphantom{0}44.0 & \hphantom{0}41.5 & \hphantom{00}0.0 \\
  & \vima-Flamingo & \hphantom{0}51.0 & \hphantom{0}51.5 & \hphantom{0}68.0 & \hphantom{0}51.5 & \hphantom{0}36.5 & \hphantom{0}50.5 & \hphantom{0}47.0 & \hphantom{00}6.0 & \hphantom{0}69.5 & \hphantom{0}28.0 & $\bestscore{\hphantom{0}45.5}$ & \hphantom{00}0.5 \\
  & \vima-GPT & \hphantom{0}50.0 & \hphantom{0}56.5 & \hphantom{0}63.0 & \hphantom{0}52.5 & \hphantom{0}32.0 & \hphantom{0}49.5 & \hphantom{0}53.0 & \hphantom{00}5.0 & \hphantom{0}78.0 & \hphantom{0}34.5 & \hphantom{0}37.5 & \hphantom{00}0.0 \\
\midrule 

\multirow{4}{*}{200M} & Ours & $\bestscore{\hphantom{0}99.0}$ & $\bestscore{100.0}$ & $\bestscore{100.0}$ & $\bestscore{\hphantom{0}97.0}$ & $\bestscore{\hphantom{0}54.5}$ & $\bestscore{100.0}$ & $\bestscore{\hphantom{0}99.0}$ & $\bestscore{\hphantom{0}17.5}$ & $\bestscore{\hphantom{0}90.5}$ & $\bestscore{\hphantom{0}97.5}$ & $\bestscore{\hphantom{0}46.0}$ & $\bestscore{\hphantom{0}43.5}$ \\
  & \vima-Gato & \hphantom{0}51.0 & \hphantom{0}58.0 & \hphantom{0}84.5 & \hphantom{0}56.5 & \hphantom{0}35.5 & \hphantom{0}53.5 & \hphantom{0}49.0 & \hphantom{0}15.0 & \hphantom{0}65.0 & \hphantom{0}52.0 & \hphantom{0}33.0 & \hphantom{00}0.0 \\
  & \vima-Flamingo & \hphantom{0}49.0 & \hphantom{0}50.0 & \hphantom{0}66.5 & \hphantom{0}47.0 & \hphantom{0}35.0 & \hphantom{0}47.5 & \hphantom{0}50.0 & \hphantom{00}4.0 & \hphantom{0}66.0 & \hphantom{0}30.5 & \hphantom{0}43.5 & \hphantom{00}0.5 \\
  & \vima-GPT & \hphantom{0}52.0 & \hphantom{0}51.0 & \hphantom{0}55.0 & \hphantom{0}49.5 & \hphantom{0}40.0 & \hphantom{0}46.0 & \hphantom{0}50.5 & \hphantom{00}5.0 & \hphantom{0}82.0 & \hphantom{0}37.0 & \hphantom{0}38.0 & \hphantom{00}1.5 \\

\bottomrule 
\end{tabular}
}

\end{table}

\begin{table}
\caption{L4 level generalization results. Model indicates robot controller parameter count. Integers in the first row refer to indices of tasks described in Appendix, Sec.~\ref{supp:sec:task_suite}.}
\label{supp:table:l4}
\centering

\begin{tabular}{c|r|cccccccccccc}
\toprule 

\textbf{Model} & \textbf{Method} & \textbf{08} & \textbf{10} & \textbf{13} & \textbf{14} \\ \midrule 

\multirow{4}{*}{2M} & Ours & \hphantom{00}6.5 & \hphantom{00}0.0 & $\bestscore{\hphantom{00}0.0}$ & $\bestscore{\hphantom{0}96.5}$ \\
  & \vima-Gato & \hphantom{0}21.0 & $\bestscore{\hphantom{00}0.5}$ & $\bestscore{\hphantom{00}0.0}$ & \hphantom{0}32.0 \\
  & \vima-Flamingo & \hphantom{0}22.0 & \hphantom{00}0.0 & $\bestscore{\hphantom{00}0.0}$ & \hphantom{0}27.5 \\
  & \vima-GPT & $\bestscore{\hphantom{0}22.5}$ & \hphantom{00}0.0 & $\bestscore{\hphantom{00}0.0}$ & \hphantom{0}22.0 \\
\midrule 

\multirow{4}{*}{4M} & Ours & $\bestscore{\hphantom{0}97.0}$ & \hphantom{00}0.0 & $\bestscore{\hphantom{00}0.0}$ & $\bestscore{\hphantom{0}99.0}$ \\
  & \vima-Gato & \hphantom{0}17.0 & $\bestscore{\hphantom{00}2.0}$ & $\bestscore{\hphantom{00}0.0}$ & \hphantom{0}34.0 \\
  & \vima-Flamingo & \hphantom{0}17.0 & \hphantom{00}0.5 & $\bestscore{\hphantom{00}0.0}$ & \hphantom{0}29.0 \\
  & \vima-GPT & \hphantom{0}19.0 & \hphantom{00}0.0 & $\bestscore{\hphantom{00}0.0}$ & \hphantom{0}22.5 \\
\midrule 

\multirow{4}{*}{9M} & Ours & $\bestscore{\hphantom{0}92.0}$ & $\bestscore{\hphantom{00}0.0}$ & $\bestscore{\hphantom{00}0.0}$ & $\bestscore{\hphantom{0}96.5}$ \\
  & \vima-Gato & \hphantom{0}18.0 & $\bestscore{\hphantom{00}0.0}$ & $\bestscore{\hphantom{00}0.0}$ & \hphantom{0}31.0 \\
  & \vima-Flamingo & \hphantom{0}21.5 & $\bestscore{\hphantom{00}0.0}$ & $\bestscore{\hphantom{00}0.0}$ & \hphantom{0}21.5 \\
  & \vima-GPT & \hphantom{0}20.5 & $\bestscore{\hphantom{00}0.0}$ & $\bestscore{\hphantom{00}0.0}$ & \hphantom{0}30.5 \\
\midrule 

\multirow{4}{*}{20M} & Ours & $\bestscore{100.0}$ & \hphantom{00}0.0 & $\bestscore{\hphantom{00}0.0}$ & $\bestscore{\hphantom{0}95.5}$ \\
  & \vima-Gato & \hphantom{0}20.5 & \hphantom{00}0.0 & $\bestscore{\hphantom{00}0.0}$ & \hphantom{0}29.0 \\
  & \vima-Flamingo & \hphantom{0}21.0 & \hphantom{00}0.0 & $\bestscore{\hphantom{00}0.0}$ & \hphantom{0}27.5 \\
  & \vima-GPT & \hphantom{0}20.5 & $\bestscore{\hphantom{00}0.5}$ & $\bestscore{\hphantom{00}0.0}$ & \hphantom{0}36.0 \\
\midrule 

\multirow{4}{*}{43M} & Ours & $\bestscore{\hphantom{0}99.0}$ & $\bestscore{\hphantom{00}0.0}$ & $\bestscore{\hphantom{00}0.0}$ & $\bestscore{\hphantom{0}97.0}$ \\
  & \vima-Gato & \hphantom{0}21.0 & $\bestscore{\hphantom{00}0.0}$ & $\bestscore{\hphantom{00}0.0}$ & \hphantom{0}30.5 \\
  & \vima-Flamingo & \hphantom{0}18.5 & $\bestscore{\hphantom{00}0.0}$ & $\bestscore{\hphantom{00}0.0}$ & \hphantom{0}24.5 \\
  & \vima-GPT & \hphantom{0}17.5 & $\bestscore{\hphantom{00}0.0}$ & $\bestscore{\hphantom{00}0.0}$ & \hphantom{0}30.0 \\
\midrule 

\multirow{4}{*}{92M} & Ours & $\bestscore{100.0}$ & \hphantom{00}0.0 & $\bestscore{\hphantom{00}0.0}$ & $\bestscore{\hphantom{0}98.5}$ \\
  & \vima-Gato & \hphantom{0}22.0 & \hphantom{00}0.0 & $\bestscore{\hphantom{00}0.0}$ & \hphantom{0}32.0 \\
  & \vima-Flamingo & \hphantom{0}19.5 & \hphantom{00}0.0 & $\bestscore{\hphantom{00}0.0}$ & \hphantom{0}25.0 \\
  & \vima-GPT & \hphantom{0}18.5 & $\bestscore{\hphantom{00}0.5}$ & $\bestscore{\hphantom{00}0.0}$ & \hphantom{0}29.5 \\
\midrule 

\multirow{4}{*}{200M} & Ours & $\bestscore{100.0}$ & $\bestscore{\hphantom{00}0.0}$ & $\bestscore{\hphantom{00}0.0}$ & $\bestscore{\hphantom{0}94.5}$ \\
  & \vima-Gato & \hphantom{0}30.5 & $\bestscore{\hphantom{00}0.0}$ & $\bestscore{\hphantom{00}0.0}$ & \hphantom{0}37.0 \\
  & \vima-Flamingo & \hphantom{0}24.5 & $\bestscore{\hphantom{00}0.0}$ & $\bestscore{\hphantom{00}0.0}$ & \hphantom{0}24.0 \\
  & \vima-GPT & \hphantom{0}20.0 & $\bestscore{\hphantom{00}0.0}$ & $\bestscore{\hphantom{00}0.0}$ & \hphantom{0}28.5 \\

\bottomrule 
\end{tabular}

\end{table}

\end{document}